\let\mypdfximage\pdfximage\def\pdfximage{\immediate\mypdfximage}%
\documentclass[acmtog]{acmart}

\acmSubmissionID{1415}

\usepackage{booktabs} %

\usepackage[ruled]{algorithm2e} %

\SetAlFnt{\small}
\SetAlCapFnt{\small}
\SetAlCapNameFnt{\small}
\SetAlCapHSkip{0pt}

\usepackage[normalem]{ulem}

\definecolor{twred}{rgb}{0.85,0,0.2}
\definecolor{twgreen}{rgb}{0,0.65,0.2}

\def\clap#1{\hbox to 0pt{\hss #1\hss}}%
\def\initials#1{\protect\clap{\smash{\raisebox{1.4ex}{\tiny{\textsf{\textit{~~#1}}}}}}}%
\makeatletter
\newcommand{\NOTE}[3]{\protect\@ifundefined{hidecomments}{%
  \strut{\color{#2}{\hspace{0pt}\initials{#1}\protect{{\small$\lfloor$}#3{\small]}}}}%
  }{}}
\newcommand{\EDITbyauthor}[4][]{\protect\@ifundefined{hidecomments}{%
  \strut{\color{#3}{\hspace{0pt}\initials{#2}\protect\sout{#1}{#4}}}%
  }{}}
\newcommand{\EDITredandgreen}[4][]{\protect\@ifundefined{hidecomments}{%
  \strut{\color{twred}{\hspace{0pt}\protect\sout{#1}{\color{twgreen}{#4}}}}%
  }{}}
\newcommand{\EDITgreenonly}[4][]{\protect\@ifundefined{hidecomments}{%
  \strut{\color{twgreen}{#4}}%
  }{}}
\newcommand{\EDITfinal}[4][]{\protect%
  \strut{#4}%
  {}}

\newcommand{\EDIT}[4][]{\EDITfinal[#1]{#2}{#3}{#4}}

\newcommand{\NOTEboxed}[3]{\protect\@ifundefined{hidecomments}{%
  {\centering\fbox{\parbox{0.97\linewidth}{\protect\EDIT{#1}{#2}{#3}}}}%
  }{}}

\def\myhrulefill{\leavevmode\leaders\hrule height 2pt\hfill\kern\z@}

\makeatother

\newcommand{\TAedit}[2][]{\protect\EDIT[#1]{TA}{orange}{#2}}

\def\ignore#1{}

\usepackage{bm}

\newcommand{\vect}[1]{{\ensuremath{\boldsymbol{#1}}}}    %

\usepackage{xspace}
\usepackage{pifont}%
\newcommand{\GG}[1]{}

\makeatletter
\DeclareRobustCommand\onedot{\futurelet\@let@token\@onedot}
\def\@onedot{\ifx\@let@token.\else.\null\fi\xspace}

\def\eg{\emph{e.g}\onedot} 
\def\ie{\emph{i.e}\onedot}

\def\iid{i.i.d\onedot}

\makeatother

\newcommand{\myparagraph}[2][\hspace{0.5em}]{\subsection{#2}}

\newcommand{\kmeans}{$k$-means\xspace}

\newcommand{\bfx}{\vect{x}}
\newcommand{\bfeps}{\vect{\epsilon}}

\usepackage{bbold}
\usepackage{amsfonts}
\usepackage{tikz}
\usepackage{nicefrac}
\usetikzlibrary{shadings,spy,calc,positioning}

\acmJournal{TOG}
\acmVolume{44}
\acmNumber{6}
\acmArticle{183}
\acmYear{2025}
\acmMonth{12}

\setcopyright{acmlicensed}

\acmDOI{10.1145/3763301}

\begin{document}

\title{Example-Based Feature Painting on Textures}

\author{Andrei-Timotei Ardelean}
\affiliation{%
\institution{Friedrich-Alexander-Universität Erlangen-Nürnberg}
\city{Erlangen}
\country{Germany}}
\email{timotei.ardelean@fau.de}
\author{Tim Weyrich}
\affiliation{%
 \institution{Friedrich-Alexander-Universität Erlangen-Nürnberg}
 \city{Erlangen}
 \country{Germany}
}
\email{tim.weyrich@fau.de}

\renewcommand\shortauthors{Timotei Ardelean \& Tim Weyrich}

\begin{abstract}
   In this work, we propose a system that covers the complete workflow for achieving controlled authoring and editing of textures that present distinctive local characteristics. These include various effects that change the surface appearance of materials, such as stains, tears, holes, abrasions, discoloration, and more. Such alterations are ubiquitous in nature, and including them in the synthesis process is crucial for generating realistic textures.
   We introduce a novel approach for creating textures with such blemishes, adopting a learning-based approach that leverages unlabeled examples. Our approach does not require manual annotations by the user; instead, it detects the appearance-altering features through unsupervised anomaly detection. The various textural features are then automatically clustered into semantically coherent groups, which are used to guide the conditional generation of images.
   Our pipeline as a whole goes from a small image collection to a versatile generative model that enables the user to interactively create and paint features on textures of arbitrary size. 
   Notably, the algorithms we introduce for diffusion-based editing and infinite stationary texture generation are generic and should prove useful in other contexts as well.\\
   Project page:
   {\footnotesize\texttt{\href{https://reality.tf.fau.de/pub/ardelean2025examplebased.html}{reality.tf.fau.de/pub/ardelean2025examplebased.html}}}
\end{abstract}



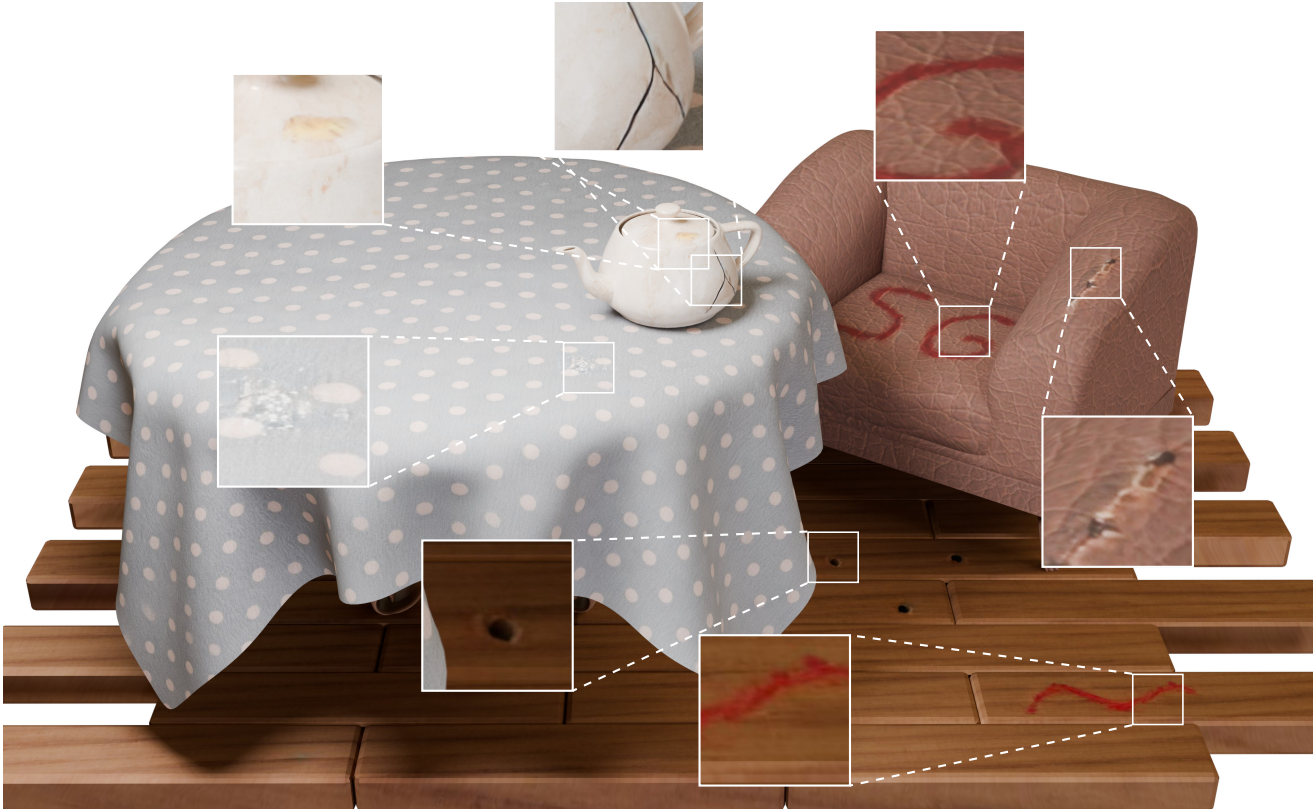
\begin{teaserfigure}
    \centering
  \begin{tikzpicture}[scale=0.97, spy using outlines={rectangle,size=10pt}]
    \begin{scope}[
        x=1mm,y=1mm,      %
        inner sep = 0pt,spy using outlines={rectangle, red, magnification=3,
          every spy on node/.append style={semithick},
        }
      ]
      \node (n0)  {\includegraphics[width=0.97\linewidth,keepaspectratio]{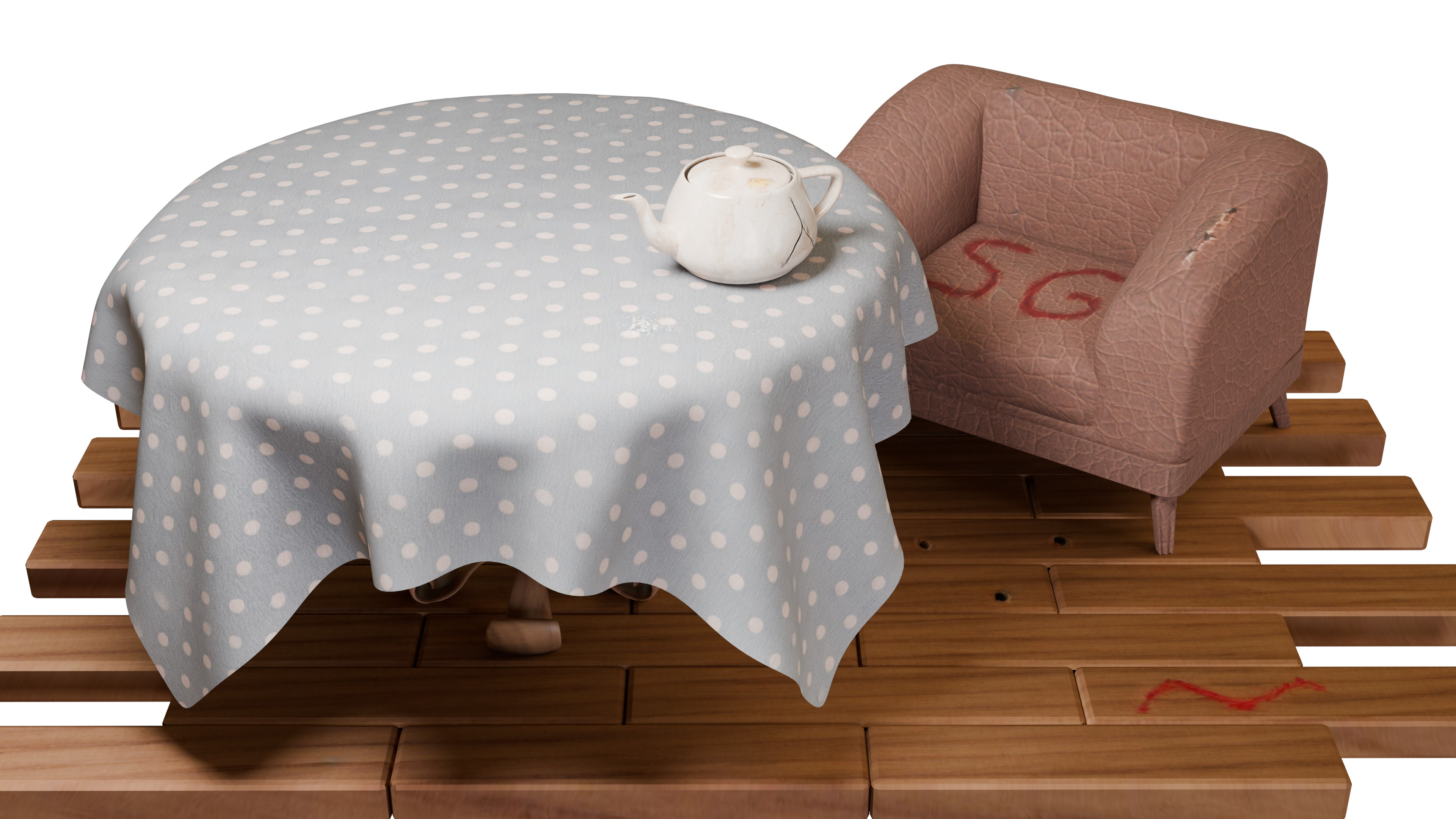}};
      \node [size=6.7mm] (fromNd4) at (-9.5,10) {};
      \spy [white,ultra thick,size=20mm] on (fromNd4.center) in node [name=toNd4] at (-50, 4);
      \node [size=6.7mm] (fromNd5) at (3.5,27) {};
      \spy [white,ultra thick,size=20mm] on (fromNd5.center) in node [name=toNd5] at (-48,40);
      \node [size=6.7mm] (fromNd6) at (8,22) {};
      \spy [white,ultra thick,size=20mm] on (fromNd6.center) in node [name=toNd6] at (-4,50);
      \node [size=6.7mm] (fromNd7) at (42,15) {};
      \spy [white,ultra thick,size=20mm] on (fromNd7.center) in node [name=toNd7] at (40,46);
      \node [size=6.7mm] (fromNd8) at (60,23) {};
      \spy [white,ultra thick,size=20mm] on (fromNd8.center) in node [name=toNd8] at (63,-7);
      \node [size=6.7mm] (fromNd9) at (68.5,-35.5) {};
      \spy [white,ultra thick,size=20mm] on (fromNd9.center) in node [name=toNd9] at (16,-37);
      \node [size=6.7mm] (fromNd10) at (24,-16) {};
      \spy [white,ultra thick,size=20mm] on (fromNd10.center) in node [name=toNd10] at (-22,-24);

    \end{scope}
    \draw[dashed,thick,white] (fromNd4.north west) -- (toNd4.north east);
    \draw[dashed,thick,white] (fromNd4.south west) -- (toNd4.south east);
    \draw[dashed,thick,white] (fromNd5.north west) -- (toNd5.north east);
    \draw[dashed,thick,white] (fromNd5.south west) -- (toNd5.south east);
    \draw[dashed,thick,white] (fromNd6.north east) -- (toNd6.north east);
    \draw[dashed,thick,white] (fromNd6.south west) -- (toNd6.south west);
    \draw[dashed,thick,white] (fromNd7.north west) -- (toNd7.south west);
    \draw[dashed,thick,white] (fromNd7.north east) -- (toNd7.south east);
    \draw[dashed,thick,white] (fromNd8.south west) -- (toNd8.north west);
    \draw[dashed,thick,white] (fromNd8.south east) -- (toNd8.north east);
    \draw[dashed,thick,white] (fromNd9.north west) -- (toNd9.north east);
    \draw[dashed,thick,white] (fromNd9.south west) -- (toNd9.south east);
    \draw[dashed,thick,white] (fromNd10.north west) -- (toNd10.north east);
    \draw[dashed,thick,white] (fromNd10.south west) -- (toNd10.south east);
\end{tikzpicture}

  \caption{\label{fig:teaser}%
    Our method generates large textures with non-stationary features learned from a small number of images. The type and shape of the painted features can be freely controlled: the figure is the result of an interactive editing session (see supplementary video). The scene uses free 3D models from \href{cgtrader.com}{cgtrader.com}.}%
  \Description{This is the teaser figure for the article. A 3D scene with different objects with textures generated by our method. The scene uses free 3D models from \href{cgtrader.com}{cgtrader.com}.}
\end{teaserfigure}

\maketitle

\newlength{\cw}

\newcommand{\figlabel}[1]{\small{\textsf{#1\vphantom{Tly}}}}
\newcommand{\figvlabel}[1]{\raisebox{0.5\cw}{\rotatebox{90}{\clap{\figlabel{#1}}}}~}
\newcommand{\figvlabelX}[1]{\raisebox{-1.25pt}{\rotatebox{90}{{\clap{\figlabel{#1}}}}}~}
\newcommand{\fighlabel}[1]{\figlabel{#1}}

\newcommand{\tablefontfamily}{\sffamily}

\section{Introduction}

In
visual content creation, automatic image generation and texturing methods seek to assist artists, reducing the time and effort required to develop photo-realistic texture assets.
Many of the existing tools, however, are biased toward an idealized, pristine appearance, so that texture artists spend significant time to augment their textures with the type of blemishes and imperfections characteristic to real-world surfaces.
Our work offers an automated framework to analyze texture samples, separating their characteristic (stationary) statistics from sporadic irregularities, that are equally characteristic, feeding an interactive system to paint such prominent features on top of the underlying pristine texture.

Our system takes as input a small number of (unannotated) images representative of a certain material, which include both normal appearance and irregular features (stains, cracks, holes, and abrasions, etc.).
The user then specifies the locations of irregularities, and our method synthesizes arbitrarily large textures that resemble the texture in the input images, with the user-specified features naturally blending into the surrounding texture.
We present the first framework that simultaneously holds the following capabilities:
\begin{enumerate}
    \item \textbf{Automatically} extracts the normal and irregular texture appearance from a small number of images.
    \item Generates textures with \textbf{spatial} and \textbf{semantic} control, that is also \textbf{interactive}.
    \item Facilitates \textbf{painting features} on both synthesized textures and real images through \textbf{feature transfer}.
    \item Creates \textbf{arbitrarily large} textures without distribution drift.
\end{enumerate}

\section{Related Work}

Painting features on textures is a task traditionally undertaken by artists that design digital assets. 
The toolbox (\TAedit{e.g., Substance 3D painter~\cite{adobe_substance}}) generally includes a library of carefully crafted materials that covers frequently occurring effects, such as dirt accumulation, or cracks.
Alongside the materials, there usually are pre-made alpha-masks that represent a realistic distribution of the desired features.
An artist would then
select an appropriate combination to overlay on the canvas as needed.
This process can be time-consuming and it is generally limited to the available pool of materials and patterns, which may not be easy to customize for a specific application.
In order to extend their capabilities, digital creation software (\eg, Photoshop~\cite{photoshop_clone}, Gimp~\cite{gimp_clone}, Substance 3D painter~\cite{adobe_substance}) include clone stamp brushes; however, the features are simply copied over the canvas, making it difficult to ensure realistic effects and transitions.

\myparagraph{Example-based feature synthesis}

\begin{figure*}
  \centering%
  \setlength{\cw}{0.157\linewidth}
  \setlength{\tabcolsep}{+0.003\linewidth}
  \begin{tabular}{cccccc}
    \figvlabel{Input labels}%
    \includegraphics[width=\cw]{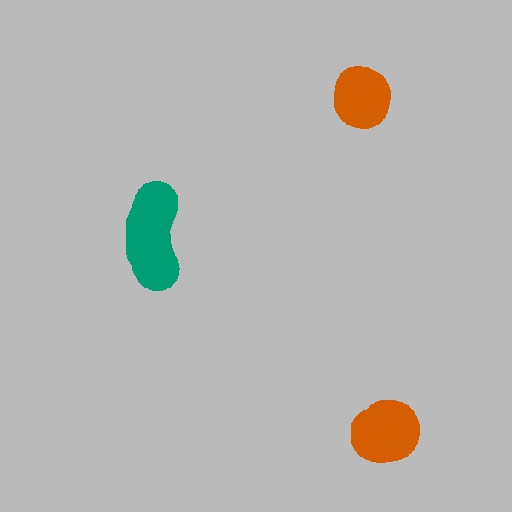} &
    \includegraphics[width=\cw]{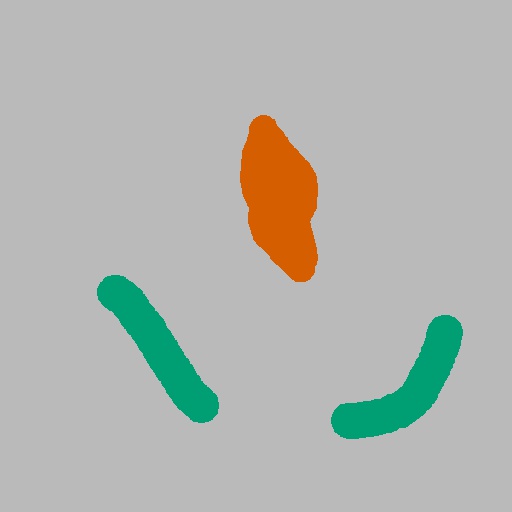} &
    \includegraphics[width=\cw]{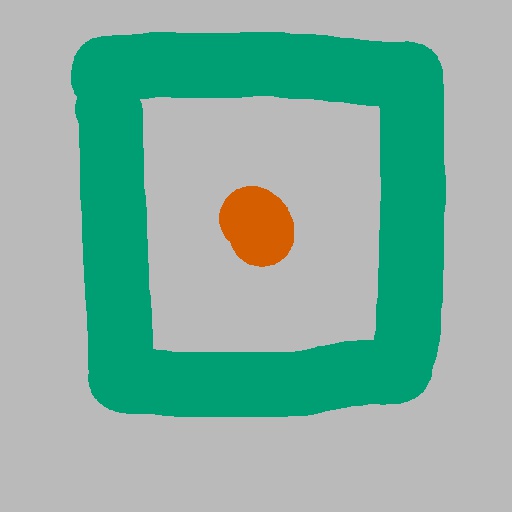} &
    \includegraphics[width=\cw]{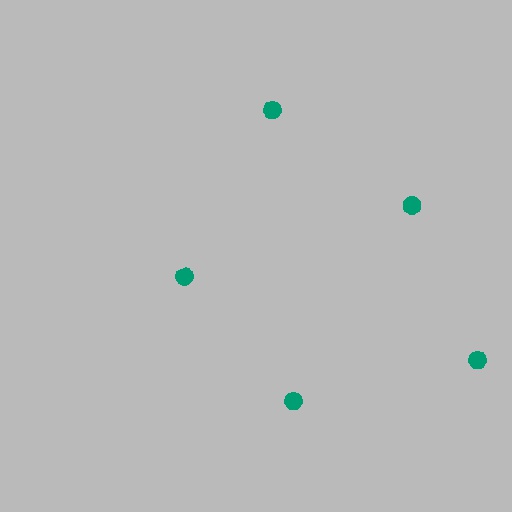} &
    \includegraphics[width=\cw]{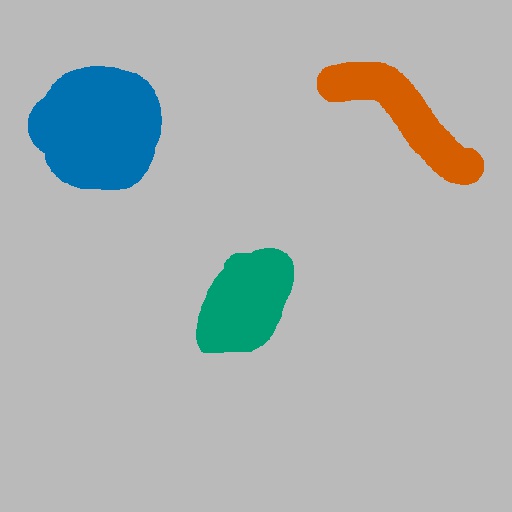} &
    \includegraphics[width=\cw]{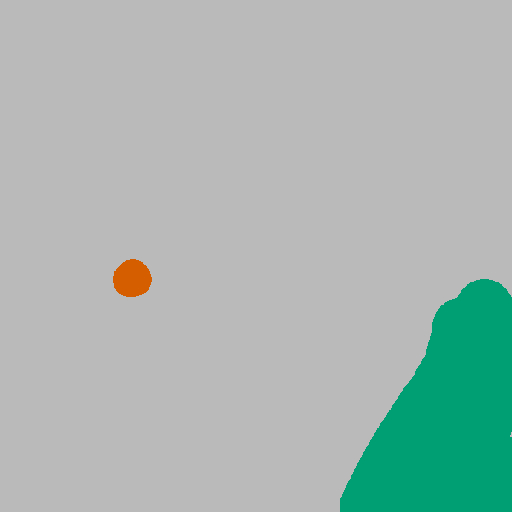} \\
    
    \figvlabel{Synthesis}%
    \includegraphics[width=\cw]{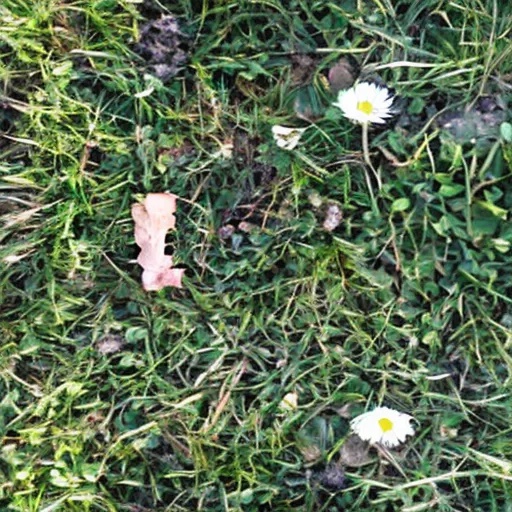} &
    \includegraphics[width=\cw]{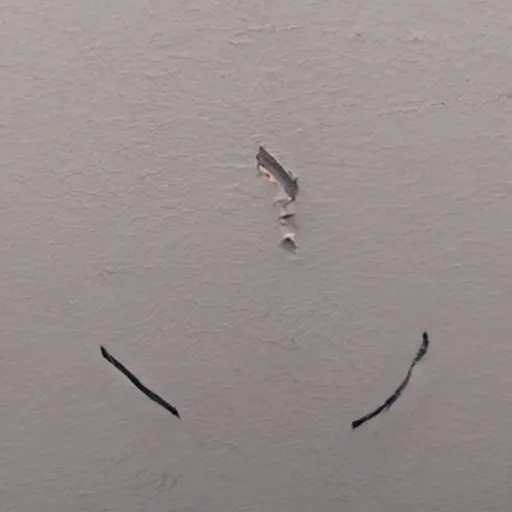} &
    \includegraphics[width=\cw]{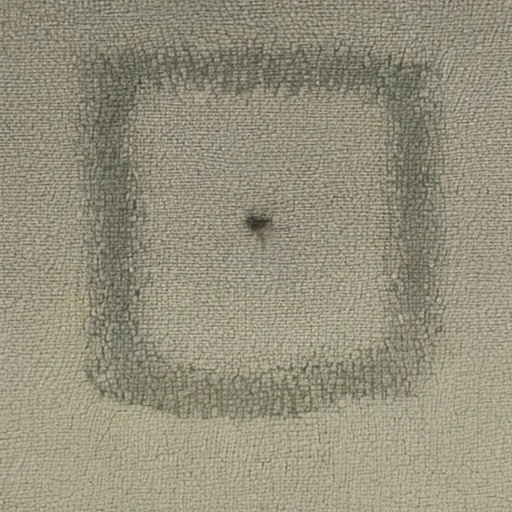} &
    \includegraphics[width=\cw]{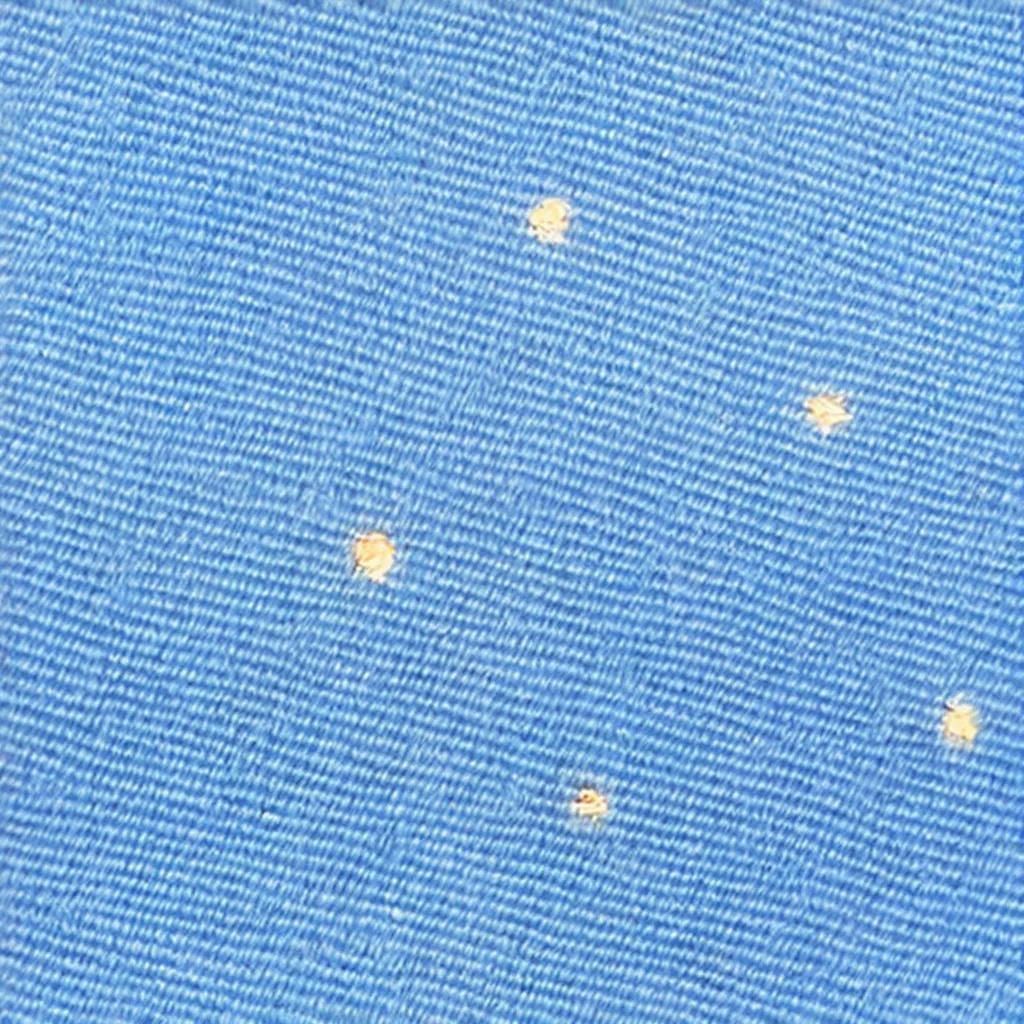} &
    \includegraphics[width=\cw]{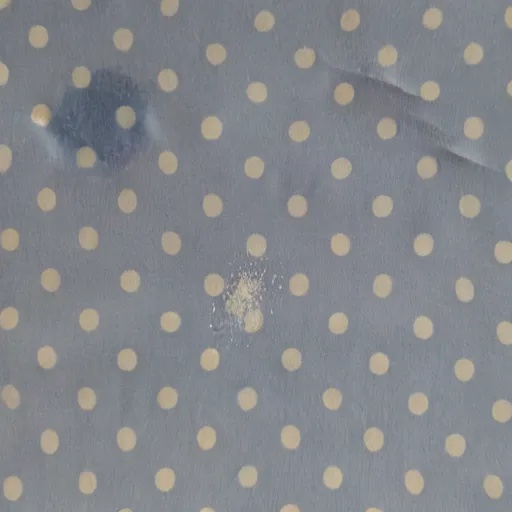} &
    \includegraphics[width=\cw]{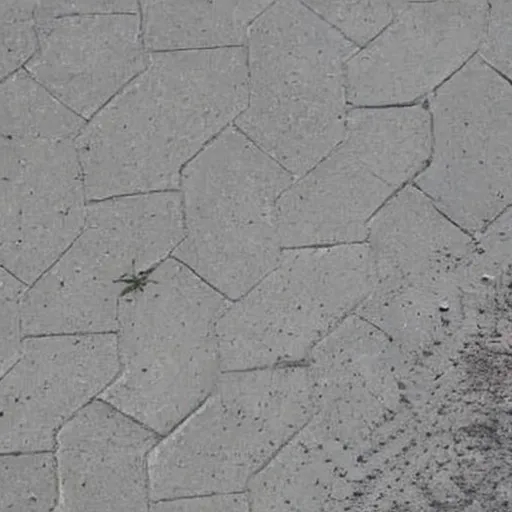} \\
    
  \end{tabular}
  \caption{\label{fig:ml_synthesis}%
    Feature-conditioned results generated by our model on various textures. The label maps are sketched by the user/artist. Please note the graceful, realistic transitions between the normal class and the painted features; digital zoom recommended.%
  }
  \Description{The Figure shows textures generated by our method.}
\end{figure*}

Several research projects directly or indirectly target the replication of image features into a new context.
In the seminal work of image analogies~\cite{hertzmann2001image}, the texture-by-numbers method is used to create a new instance of a texture and paint the prominent features as guided by a layout map.
The same can be achieved using single-image generative models and image reshuffling techniques.
SinGAN~\cite{shaham2019singan} and SinDiffusion~\cite{wang2022sindiffusion} train a generative adversarial network (GAN) and a diffusion model~\cite{dhariwal2021diffusion} respectively from a single image. The models can
then
be used to generate similar, realistic patches in a new configuration.
GPNN~\cite{granot2022drop}, 
Image style transfer~\cite{gatys2016image}, Sliced Wasserstein~\cite{heitz2021sliced}, and GPDN~\cite{elnekave2022generating} are non-parametric generative methods that can be used to paint features by mimicking the patches and statistics from a given source image.
Neural Texture Synthesis with Guided Correspondence~\cite{zhou2023neural}, Non-Stationary Textures using Self-Rectification~\cite{zhou2024generating}, and Texture Reformer~\cite{wang2022texture} focus specifically on textures with non-stationary regions. 
These approaches showcase different ways to condition the generation in order to improve the authoring process: orientation and progression control, content image, and a collage of crops.
Painting With Texture~\cite{ritter2006painting}, Painting by Feature~\cite{lukavc2013painting}, Brushables~\cite{lukavc2015brushables}, and Neural Brushstroke~\cite{shugrina2022neube} propose different ways to create brushes from images, enabling painting of the extracted textural features on a canvas.
Recently,
Diffusion Painting~\cite{hu2024diffusion} displayed remarkable capabilities in terms of the texture complexity
controllable
by a brush.
Using a diffusion model pretrained on a large dataset, the method hallucinates realistic variations and transitions from a single texture image, albeit with limited fidelity to the input mask.

While the above-mentioned approaches share the advantage of being able to work from a single image,
it is desirable to incorporate the information from multiple source images of the same texture class if available,
as a single image often does not cover the appearance of a feature type in its fullness, or does not contain all transition types that are characteristic for the given material.
That being said, in most cases it is not straight forward to extend the methods to accept multiple images in a way that actually improves the generation process.
Our method is fine-tuned on multiple images from a single texture class in order to learn from several instantiations of a certain prominent feature type.

A limitation of previous methods is the requirement to manually select and indicate the relevant features from the input images.
As we want to capture the entire distribution of a texture and/or the possible prominent features (stains, cuts, etc), several dozen images might be needed. In this case, manual segmentation would put an unreasonable burden on the user.
Therefore, in this work we also address a preprocessing part \TAedit{in the asset-creation} process, by automatically finding and grouping the relevant features.

\myparagraph{Anomaly Localization and Classification}
In order to automatically detect the prominent features, which are non-stationary regions in the texture, we pose the problem as an anomaly detection task.
Since the input consists of a mix of (unlabeled) normal and anomalous features, we find ourselves in a fully unsupervised setting.
This is more challenging than the one-class classification task employed by most anomaly detection methods, where the normal instances are labeled~\cite{roth2022towards, deng2022anomaly, batzner2024efficientad}.

Fully unsupervised anomaly localization is framed as normality-supervised detection with contamination~\cite{yoon2022selfsupervise, liu2022unsupervised, zhang2024s, patel2023self} or zero-shot anomaly detection with test-time adaptation~\cite{li2023zeroshot, li2024musc}. 
BlindLCA~\cite{ardelean2024blind} is the current state-of-the-art method designed specifically for textures.
We build on this method and adapt it to pixel-level anomaly segmentation.
The unsupervised classification of anomalies into semantic categories has been approached by prior work~\cite{sohn2023anomaly, ardelean2024blind} at the image level. Differently, we perform the semantic segmentation at the level of pixels rather than images and lift the limitation of a single anomaly type per texture instance.

More often than not, the features that are painted on a certain texture are trying to replicate naturally occurring defects that can be attributed to weathering. Therefore, weathering synthesis is targeted by methods related to altering the appearance of textures.

\myparagraph{Weathering synthesis}
A traditional approach to weathering is to develop material-specific simulations based on the physical and chemical processes that occur through time and update the appearance accordingly~\cite{dorsey2006modeling, chen2005visual, liu2012physically, bajo2021physically}.
While there are certain advantages granted by a physically-based model, this class of methods does not generalize across different materials, and it often requires an explicit model of the external factors that influence the weathering.

\begin{figure*}[tbp]
  \centering%
  \includegraphics[width=.97\linewidth]{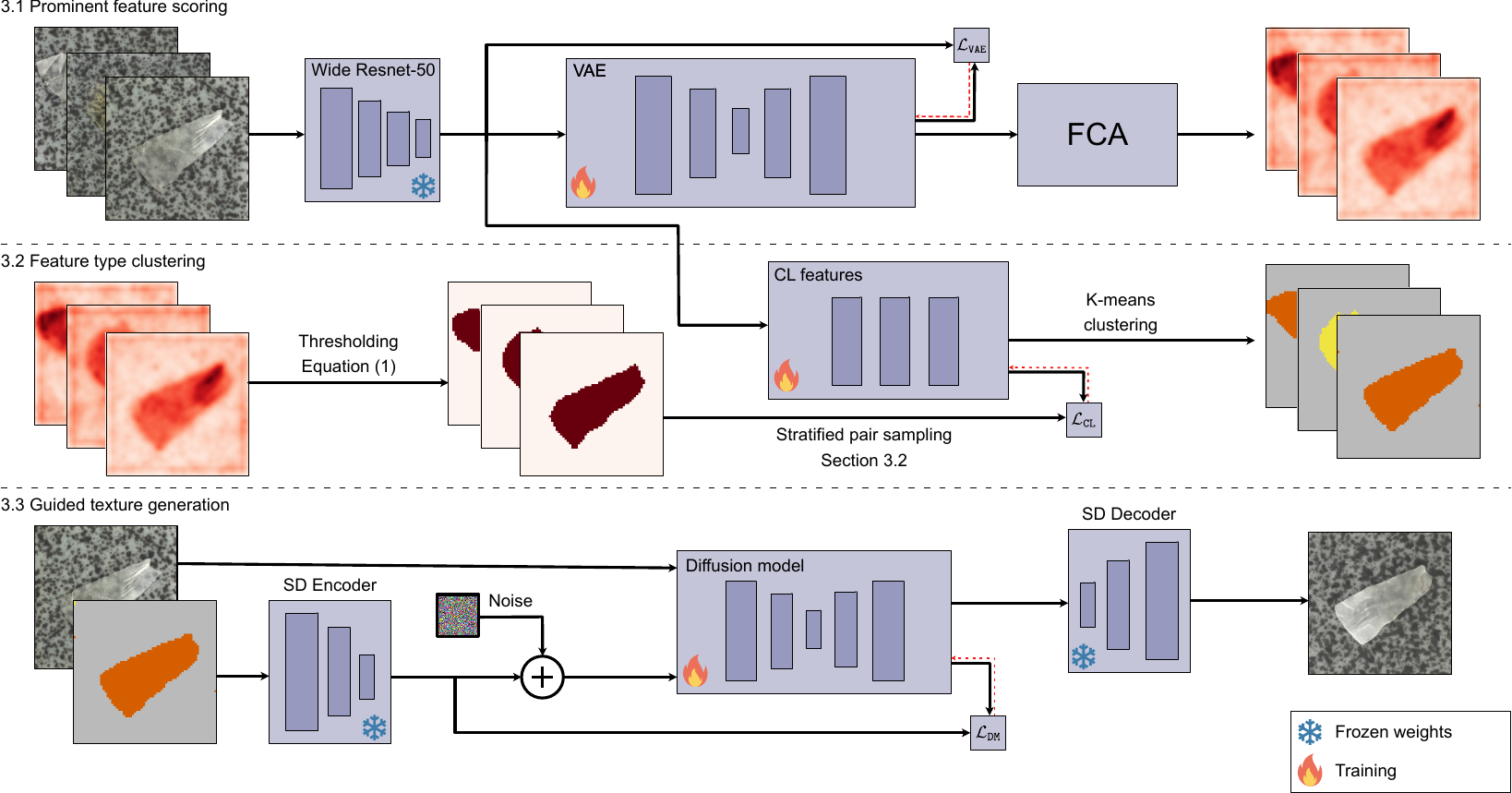}\\[-1ex]  %
  \caption{\label{fig:pipeline}%
    Our 3-stage pipeline for obtaining a generative model for textures with prominent features.\hspace*{1in}}
  \Description{The figure presents the 3 stages of our pipeline, from top to bottom: feature detection, semantic feature segmentation, and feature painting/synthesis.}
\end{figure*}

A different line of work builds time-varying appearance models based on a single image and a few annotations provided by the user. 
\citet{wang_appearance_2006} are the first to develop an appearance manifold in
BRDF space and use it to model the time-dependent change from normal to weathered. 
The manifold is constructed with the help of the user, who selects the least- and most-weathered parts of the image. 
A similar strategy was
successfully applied directly in color space~\cite{xuey_image-based_2008, bandeira_synthesis_2009}, separating
illuminance and reflectance.
\citet{iizuka_single_2016} advance from simple chromatic transitions and produce weathering with more complex appearance.
Synthesis is performed using image quilting techniques~\cite{efros_image_2001} based on a
weathering
exemplar.
To improve the coverage of weathered appearance distribution, \citet{du_multi-exemplar-guided_2023} additionally identify discrete degrees of weathering and create multiple
weathering
exemplars accordingly.
\citet{Bellini2016TimevaryingWI} propose a time-varying weathering framework that does not require user-annotations. 
The method predicts anomaly-maps which are used to remove the anomalies or to increase their amount by repeating the existing flaws.
While these methods yield impressive results from just one image, they are restricted to a single weathering trajectory (type) and
can mostly model relatively simple effects,
\eg,
decoloration, peeling, moss growth, and oxidation.

Learning-based methods can use multiple images and model various weathering types and effects.
\citet{Chen2021GuidedIW} pose the weathering task as an image-to-image translation problem using a pix2pix~\citep{isola2017image} GAN. The model is trained on a single image, with automatic annotations~\cite{Bellini2016TimevaryingWI}. While their approach could support multiple images for training, it is limited to a single weathering type.
Recently, \citet{Hao2023NaturalID} introduced an approach for weathered texture synthesis that supports multiple defect types; however, the method is trained on several thousand weathered images prefiltered and grouped by type.
The guided texture transition method of \citet{guerrero2024texsliders} allows users to holistically modify an image with fine control over the degree of weathering, \TAedit{and the concurrent work of \citet{hadadan2025generative} aims to add realistic details, such as signs of wear, using text prompts. In contrast, we focus on spatially controllable edits, while preserving the appearance of unweathered regions.}

Our approach allows the user to create weathered textures of arbitrary size, and therefore relates to the creation of infinite textures using generative models~\cite{bergmann2017learning, lin2021infinitygan, wang2024infinite}. Similarly to \citet{wang2024infinite}, we use a diffusion model and averaged denoising scores~\cite{omer2023multidiffusion} to progressively generate a large texture. 
Additionally, we propose a way to reduce distribution drift, preserving global consistency.

\section{Method}

Our pipeline
consists of three
stages (Fig.~\ref{fig:pipeline}).
The first stage
is
our approach for automatically identifying
regions with prominent
features
by posing that task as a fully-unsupervised anomaly localization problem.
This framing is feasible since
regions with irregular features break the stationarity assumption of textures, making these regions deviate from the overall statistics.
After identifying
such regions,
the second stage addresses the separation of prominent features in different groups. 
Leveraging the anomaly maps, we devise an approach for sampling positive and negative pairs of pixels, which are used to optimize a contrastive learning objective. 
This produces a disentangled feature space that we cluster using \kmeans to obtain pixel-level semantic maps where the labels indicate the feature type or the absence of a prominent feature accordingly.
The result of the second stage enables us to train a generative model that follows the desired spatial and semantic conditions.
Our diffusion model can generate new textures interactively (1 sec for a 512\texttimes{}512 image).
Moreover, we demonstrate synthesis of arbitrary size using constant GPU memory, and design a way to avoid distribution drift across the generated texture.
Finally, we advance a noise-mixing technique to support texture editing while remaining faithful to the original input. 
Our editing method even supports transferring blemishes from one texture class to an image of another class.

\subsection{Prominent feature scoring}
\label{sec:semantic}

\begin{figure*}
  \centering%
  \setlength{\cw}{0.102\linewidth}%
  \setlength{\tabcolsep}{+0.003\linewidth}%
  \begin{tabular}{ccccccccc}
    \figvlabel{Input}%
    \includegraphics[width=\cw]{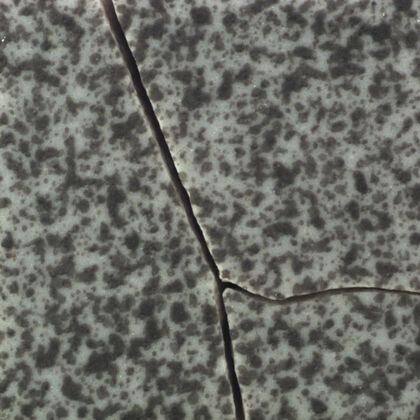} &
    \includegraphics[width=\cw]{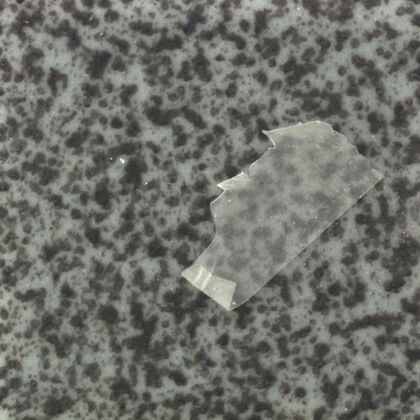} &
    \includegraphics[width=\cw]{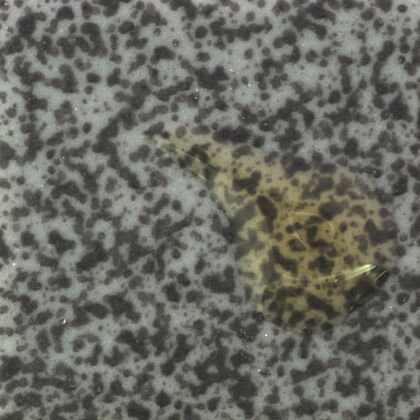} &
    \includegraphics[width=\cw]{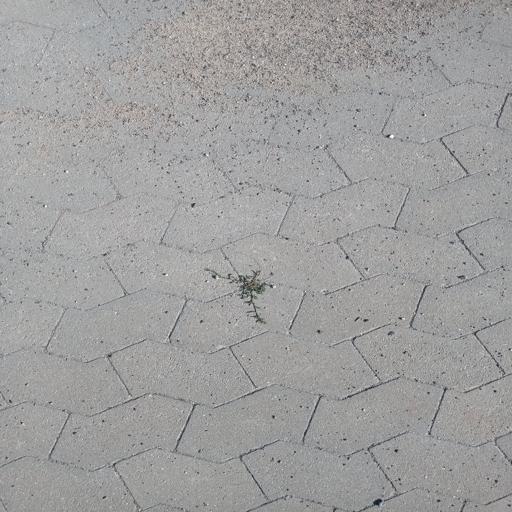} &
    \includegraphics[width=\cw]{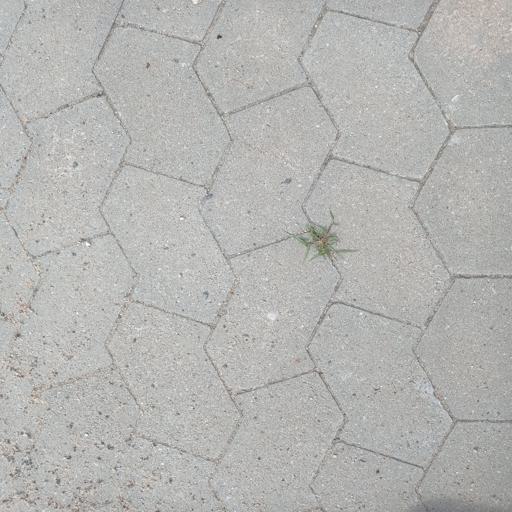} &
    \includegraphics[width=\cw]{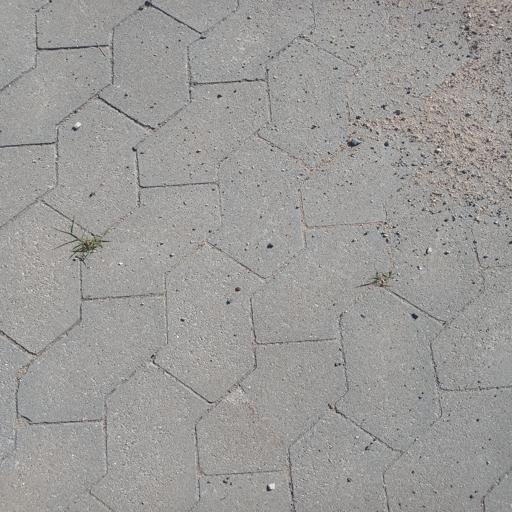} &
    \includegraphics[width=\cw]{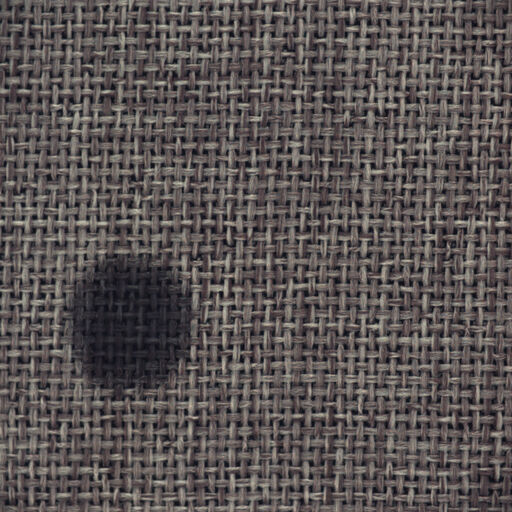} &
    \includegraphics[width=\cw]{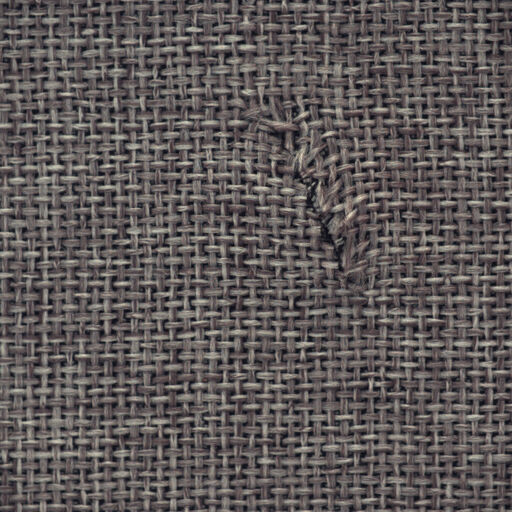} &
    \includegraphics[width=\cw]{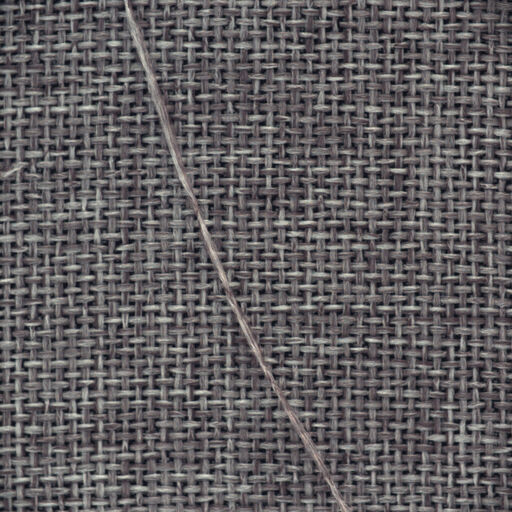} \\
    
    \figvlabel{Ours}%
    \includegraphics[width=\cw]{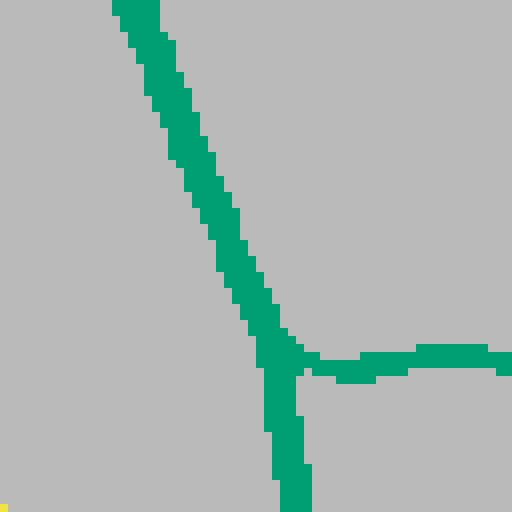} &
    \includegraphics[width=\cw]{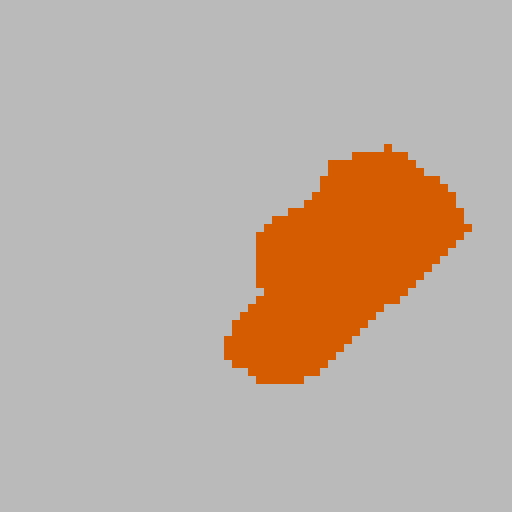} &
    \includegraphics[width=\cw]{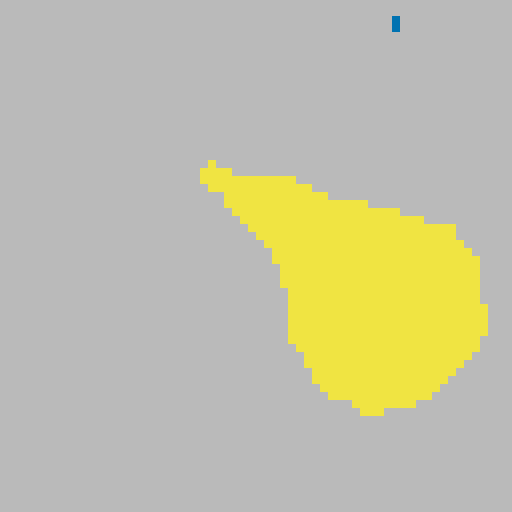} &
    \includegraphics[width=\cw]{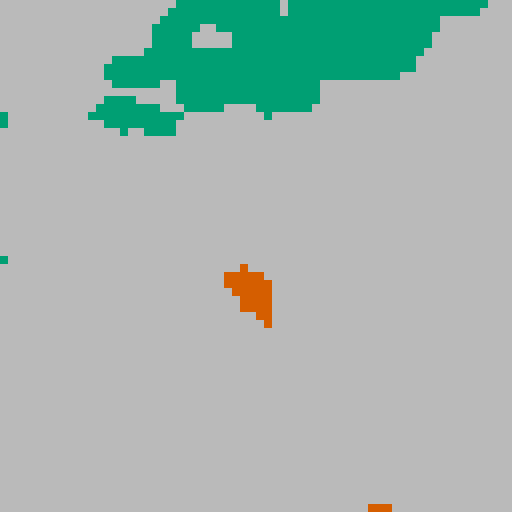} &
    \includegraphics[width=\cw]{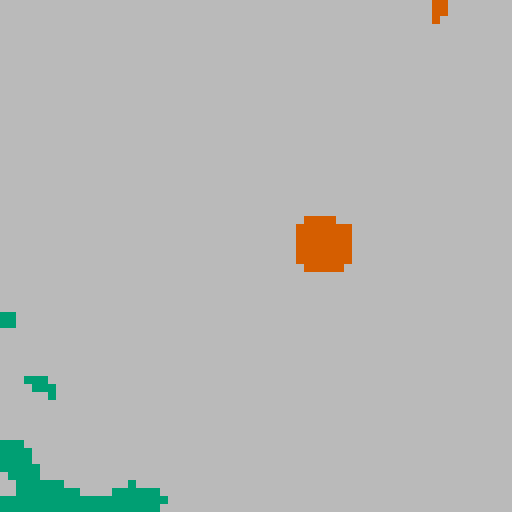} &
    \includegraphics[width=\cw]{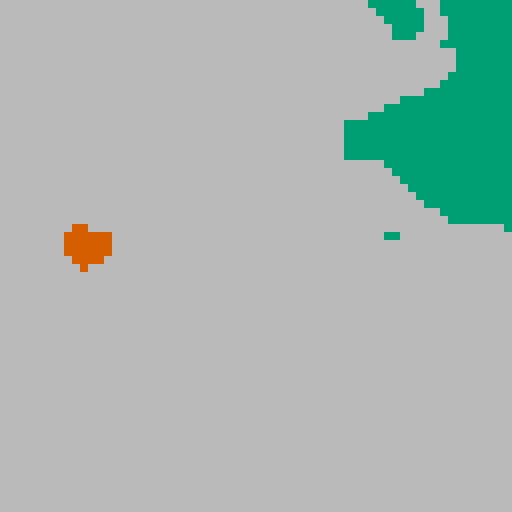} &
    \includegraphics[width=\cw]{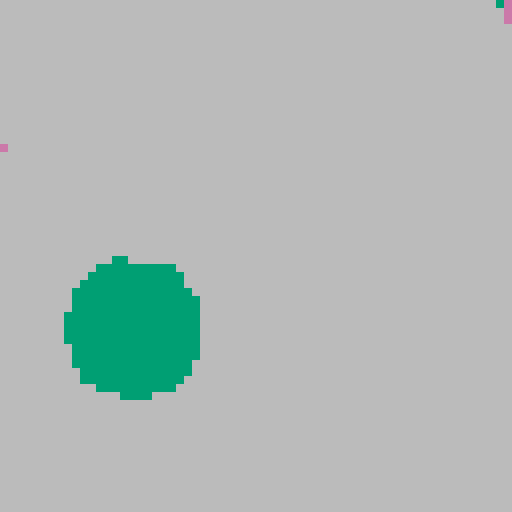} &
    \includegraphics[width=\cw]{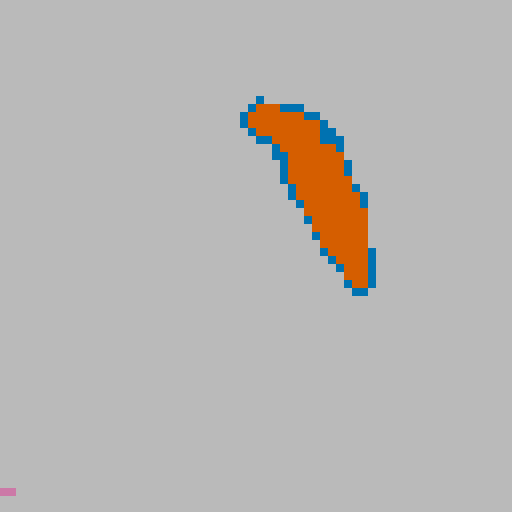} &
    \includegraphics[width=\cw]{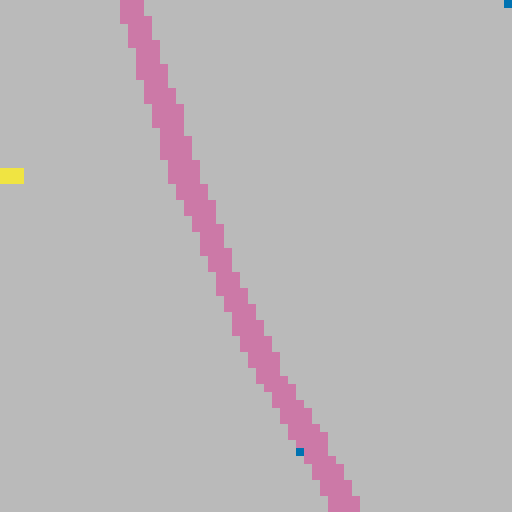} \\
    
    \figvlabel{BlindLCA}%
    \includegraphics[width=\cw]{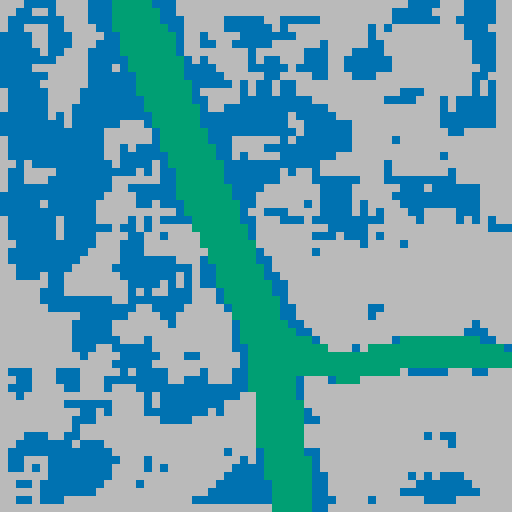} &
    \includegraphics[width=\cw]{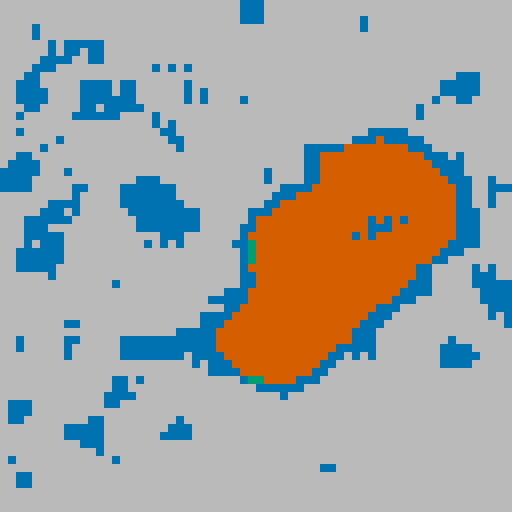} &
    \includegraphics[width=\cw]{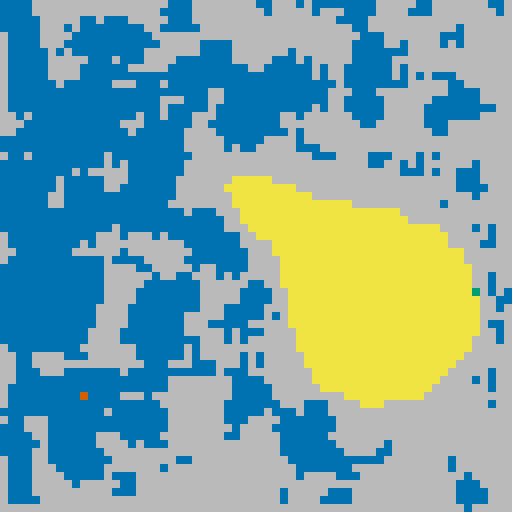} &
    \includegraphics[width=\cw]{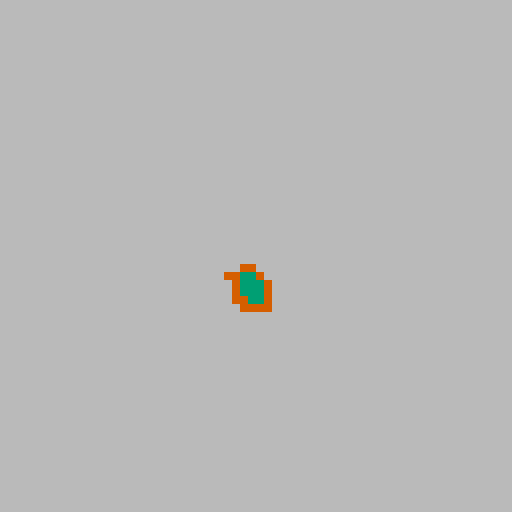} &
    \includegraphics[width=\cw]{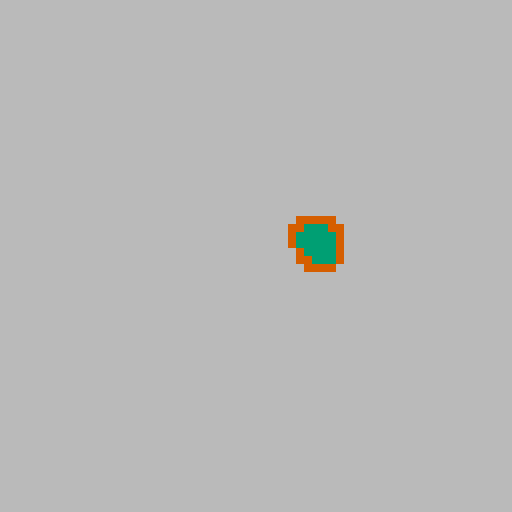} &
    \includegraphics[width=\cw]{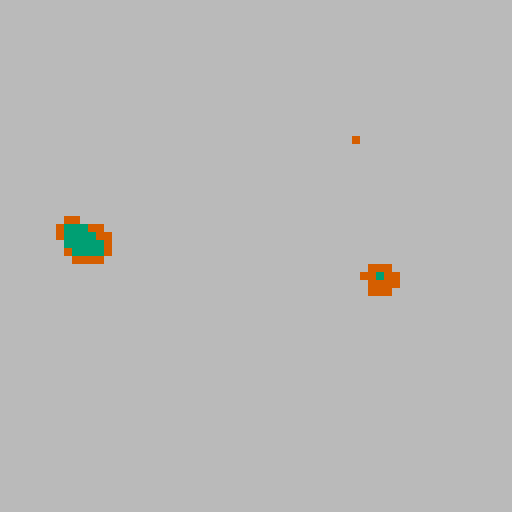} &
    \includegraphics[width=\cw]{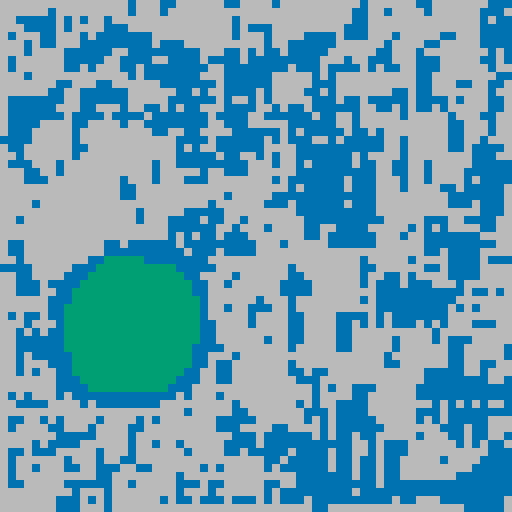} &
    \includegraphics[width=\cw]{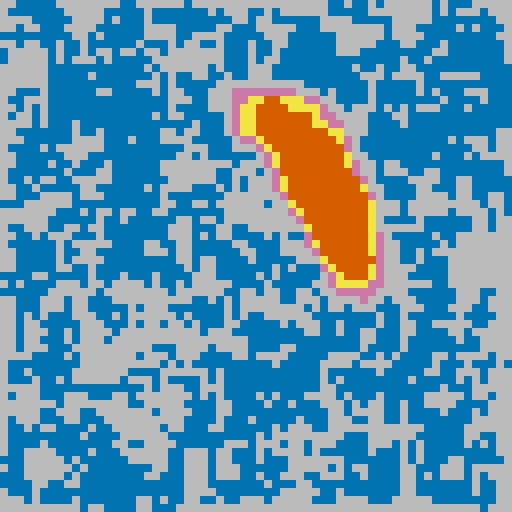} &
    \includegraphics[width=\cw]{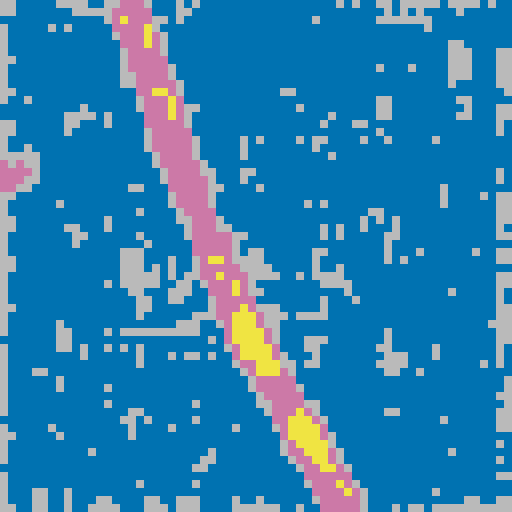} \\

    \figvlabel{STEGO}%
    \includegraphics[width=\cw]{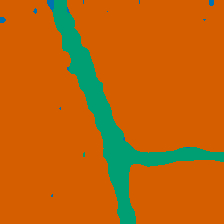} &
    \includegraphics[width=\cw]{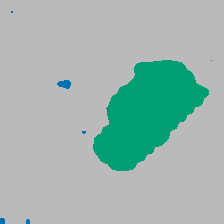} &
    \includegraphics[width=\cw]{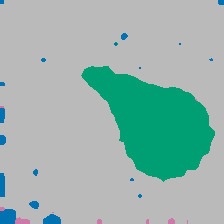} &
    \includegraphics[width=\cw]{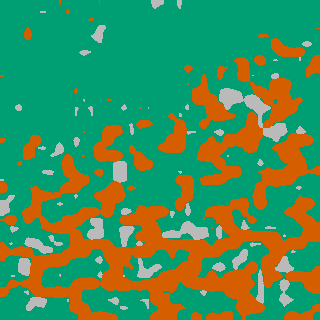} &
    \includegraphics[width=\cw]{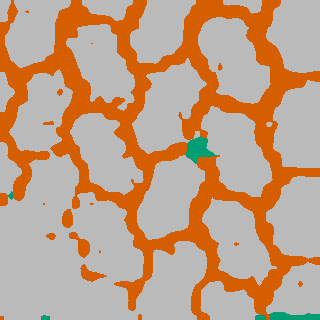} &
    \includegraphics[width=\cw]{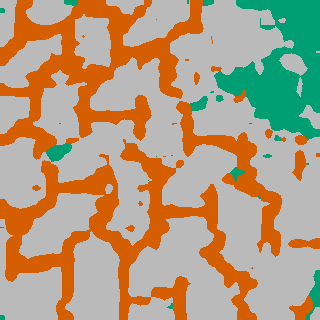} &
    \includegraphics[width=\cw]{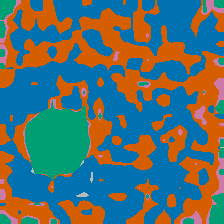} &
    \includegraphics[width=\cw]{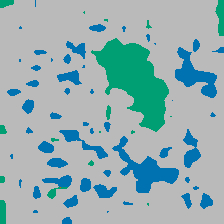} &
    \includegraphics[width=\cw]{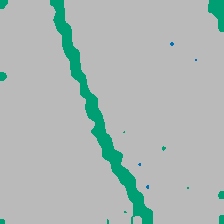} \\
  \end{tabular}
  \caption{\label{fig:segmentation}%
    Results of our pixel-level anomaly segmentation. We seek to have the background pixels mapped to a single class and the different features to distinct labels.
    The figure compares the results of our approach, BlindLCA~\cite{ardelean2024blind}, and STEGO~\cite{hamilton2022unsupervised}.%
    \\[-0.25\baselineskip]}
  \Description{
    The figure compares the results of our approach, BlindLCA and STEGO. Our method correctly identifies the normal class and distinguishes between different anomaly types.
  }
\end{figure*}

The input of our pipeline consists of a handful of photographs of textures that are largely stationary, yet contain prominent features that we want to disentangle. These irregularities can be considered anomalies in the context of the global textures, which can be detected using unsupervised anomaly localization.

\TAedit{The initial stage of our pipeline leverages an unsupervised anomaly detection approach, as a first step towards removing the need for manual user annotations, required by previous methods~\cite{iizuka_single_2016, du_multi-exemplar-guided_2023}}.
We employ FCA, the zero-shot anomaly detection method of \citet{ardelean2024high}, and apply it to the residual obtained from a VAE-based reconstruction, similarly to Blind LCA~\cite{ardelean2024blind}.
The output of this step is a set of anomaly-score maps $\{A_i\}_{i=1}^N$, one for each input image $\{I_i\}_{i=1}^N$ \TAedit{, as shown in the first section of Fig.~\ref{fig:pipeline}.
These methods, however, do not provide a way to threshold the anomaly scores, which have an arbitrary range.

\subsection{Feature type clustering}

Our pixel-level semantic segmentation procedure requires well-defined (binary) anomaly regions.}
Therefore, we propose an adaptive thresholding function to obtain a binary segmentation $B_i$ of the areas of interest. Let
$B_i^{xy} := {\scriptsize\Big\{\!\!\begin{array}{l}1\quad A_i^{xy} > \mathcal{T}(A_i)\!\!\\0\quad\text{otherwise}\end{array}}$\!;
where $\mathcal{T}$ is a function that adaptively computes the threshold based on the anomaly scores.
We observed that global thresholding and simple quantiles do not generalize well across different datasets.
Otsu's method~\cite{otsu1975threshold} yields reasonable results, yet it tends to produce segmentations that are too permissive. To reduce the number of false positives, we skew the distribution using an exponential factor: $\hat{\mathcal{T}}(A_i) = \smash{\texttt{Otsu}(A_i^\beta)^{\nicefrac{1}{\beta}}}$, where $\beta=1.5$ in our experiments.

This function is able to find adequate thresholds across different datasets; however, since it always segments a part of the image as anomalous it will inevitably yield false positives on textures that are completely stationary.
To overcome this issue, we compute a global threshold and segment each anomaly map based on the maximum between the local and global thresholds: 
\begin{equation}
\label{eq:threshold}
\mathcal{T}(A_i) := \max(\texttt{Otsu}(\{\hat{\mathcal{T}}(A_j)\}_{j=1}^N), \hat{\mathcal{T}}(A_i)) \;. 
\end{equation}
We empirically validate our thresholding function in the supplementary material (Sec~\ref{asec:compare_thr}).

To enhance the authoring control of the final generative model, we are interested in grouping the prominent features of the same type into clusters.
Based on the binary detection of anomalous regions, we identify the various types of features through contrastive learning (CL) and clustering.
As opposed to prior work~\cite{ardelean2024blind}, we are interested in pixel-level rather than image-level clustering. Moreover, we do not assume there is only one possible anomaly type per image, making the method more flexible and applicable in practical scenarios.
To this end, we divide each image into regions by computing the connected components in the binary masks $B_i$.
For simplicity, it is reasonable to assume that each such region either belongs to the normal class or contains a single type of feature.
We then compute region-level descriptors by averaging the pixel-level features extracted by a pretrained WideResnet-50 network, and use the descriptors to find positive pairs as nearest neighbors.
Importantly, we observe that naively creating negative pairs for contrastive learning by sampling regions that are far away in feature space yields poor results.
As shown in Fig.~\ref{fig:pair_sampling}, this is due to the oversampling of the normal class, which limits the effectiveness of contrastive learning to separate the various anomaly types.
We alleviate this problem by sampling negative pairs in a stratified manner, \ie, the regions are preclustered using \kmeans based on the computed feature descriptors, and negative pairs are sampled equally across these clusters.
The number of classes used for preclustering is a free hyperparameter of our method; that being said, we observed that using the same number of classes as for the final clustering is a robust choice.

The positive and negative pairs are used to train a 3-layer
CNN,
with a receptive field of 5\texttimes{}5, optimizing for the InfoNCE~\cite{oord2018representation} contrastive objective.
After training, the CL-features produced by the neural net   are easily separable into clusters: to obtain pixel-level segmentations, we pool the CL-features for all pixels in all input images and cluster them using \kmeans.

\subsection{Guided texture generation}

The output of the clustering stage consists of $N$ label maps $\{L\}_{i=1}^N$ with values from $0$ (normal class) to $K$, the number of feature types in the dataset.
We leverage the label maps as conditioning for a diffusion model~\cite{ho2020denoising} (DM) trained to generate the same set of images $\{I_i\}_{i=1}^N$.

Diffusion models are a class of generative models that synthesize outputs through progressive denoising. 
We include here a brief explanation of the concept of diffusion models, and refer the reader to a more extensive analysis of these models~\cite{ho2020denoising, song2021score, song2021denoising, karras2022elucidating}.
Let $p_{\text{data}}(\bfx)$ be the distribution of the data to be modeled; in our case, natural textures.
The forward diffusion process pushes samples away from the true data distribution by adding \iid (independent identically distributed) Gaussian noise with standard deviation $\sigma(t)$.
The noise is increasing with the timestep $t$, so that for a large enough $t_N$ the resulting distribution $p(\bfx; \sigma({t_N}))$ is virtually identical to the normal distribution $\mathcal{N}(\mathbf{0}, \sigma({t_N})^2 \vect{I})$.
Sampling from $p_\text{data}(\bfx)$ can be achieved by reversing the diffusion process. In practice, this is implemented by a progressively applied, learned denoising function, starting from pure noise sampled from $\mathcal{N}(\mathbf{0}, \sigma({t_N})^2 \vect{I})$. The denoising is applied at several steps with decreasing standard deviation, essentially modeling the distribution $p(\bfx; \sigma({t_i}))$, with a monotonic timestep schedule $\{t_i\}_{i=0}^N$, where $\sigma(t_0)=0$.
After iterating from $N$ to $0$, the resulting samples should resemble the original distribution $p(\bfx; 0)$. In our implementation, we follow the ODE formulation from \citet{karras2022elucidating} (EDM)
including scheduling, preconditioning, and the Heun solver. The direction to the data distribution is approximated by a neural network $\bfeps(\bfx, \sigma(t))$ that can be trained with a simple MSE loss. 
Similarly to \citet{karras2024analyzing}, we perform the diffusion in latent space, using the Variational Autoencoder (VAE) from Stable Diffusion (SD)~\cite{rombach2022high}. 
Moreover, we add spatial conditioning to enable fine-grained control over the generated images.

Conditioning a generative model through a spatial map has been used for many image-to-image translation tasks, such as layout-based generation, colorization, and inpainting.
Our use-case is similar to the generation of images based on semantic maps~\cite{park2019SPADE, zhang2023adding, ko2024stochastic}, where the semantic labels are represented by
the $K$ feature types.
Our method is designed to work with a very small number of photographs -- as low as one.
To incorporate this additional input, we modify the U-Net-based~\cite{ronneberger2015u} ADM~\cite{dhariwal2021diffusion} architecture used by EDM~\cite{karras2022elucidating}: \TAedit{The label maps are encoded with a small convolutional network and then added to the timestep embeddings to modulate the activations of the U-Net at several intermediate layers. 
In the original architecture, in each U-Net block, the timestep embedding is processed using a linear layer to predict an affine transformation for each feature channel.
We modify the U-Net blocks to accept spatially varying embeddings, which are processed by a convolutional layer to predict a different scale and shift for each spatial position and channel.}
The additional parameters are trained jointly with the rest of the network.
To facilitate training from a few images, we pretrain our diffusion model on the DTD dataset~\cite{cimpoi14describing} (5640 textures).
When training the conditional diffusion model $\bfeps(\bfx, \vect{c}; \sigma(t))$ the label map $\vect{c}$ is dropped with a probability of $7.5\%$, enabling classifier-free guidance~\cite{ho2021classifierfree} during inference.
\TAedit{Since the dataset contains different (47) texture classes, we use the class labels instead of semantic masks as conditioning, that is, the class label is spatially expanded to match the shape required by the model. 
Note that since the number of prominent features (semantic classes) during fine-tuning differs from the number of classes in the DTD dataset used during pretraining, we cannot reuse the first convolutional layer that embeds the class label; therefore, we drop that specific layer and train it from scratch.}

We choose EDM as our backbone as it generates high-quality images with low latency, enabling our interactive synthesis goal. 
Other diffusion models or different generative approaches could potentially be used for this stage (e.g., inpainting \cite{suvorov2022resolution, Lugmayr_2022_CVPR} or Texture Mixer \cite{yu2019texture}).
We provide an extended analysis of different choices in the supplementary (\ref{asec:diffusion}).

\TAedit{Similarly, various mechanisms for fine-tuning diffusion models have been established in recent years, such as: Dreambooth~\cite{ruiz2023dreambooth} for the preservation of visual characteristic of a certain subject, ControlNet~\cite{zhang2023adding} for injecting spatial control into a diffusion model trained without such condition, and LoRA~\cite{hu2022lora} to make fine-tuning more efficient with respect to the number of training samples, time, and memory footprint.
A combination of the techniques above (e.g., ControlLora~\cite{wu2024controllorav3} and CtrLora~\cite{xu2025ctrlora}) may enable the generative model of our pipeline to use larger models with massive pretraining.
In this work, however, we use a lightweight network and very little pretraining, prioritizing interactive inference time and limiting dependence on large-scale data.}

As we show in Fig.~\ref{fig:ml_synthesis}, this setup allows us to generate new, realistic images, with the desired texture, that follow the conditioning of the label maps. 
This functionality covers the first two capabilities of our pipeline. 
In the following, we present how our method can be used for real-image editing and arbitrary-size image generation. 

\subsection{Editing}

\begin{figure}
  \centering%
  \setlength{\cw}{0.32\linewidth}%
  \setlength{\tabcolsep}{0.003\linewidth}%
  \begin{tabular}{ccc}
    \fighlabel{Original} &
    \fighlabel{Examples} &
    \fighlabel{Mask} \\
    
    \includegraphics[width=\cw]{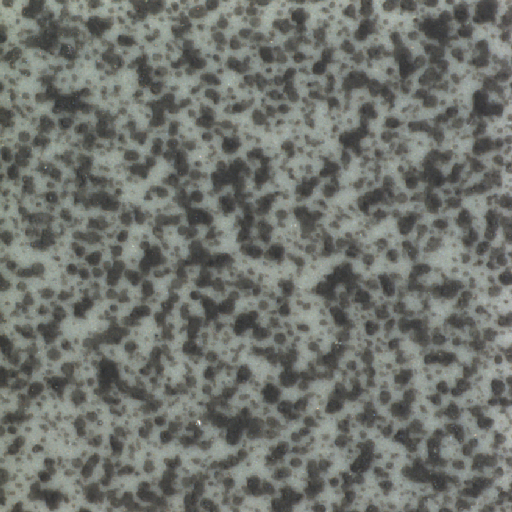} &
    \raisebox{0.45\cw}{\begin{tabular}{cc}
        \includegraphics[width=0.49\cw]{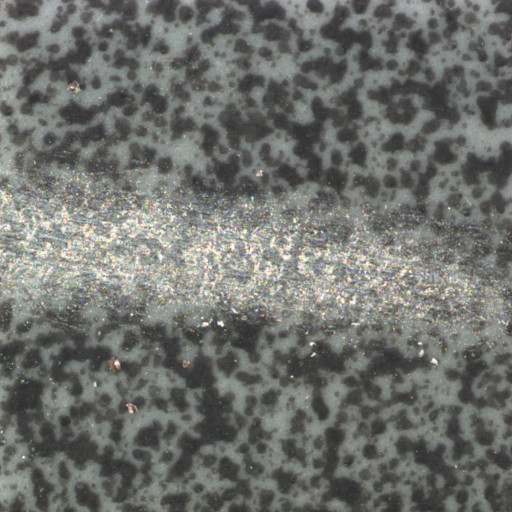} &
        \includegraphics[width=0.49\cw]{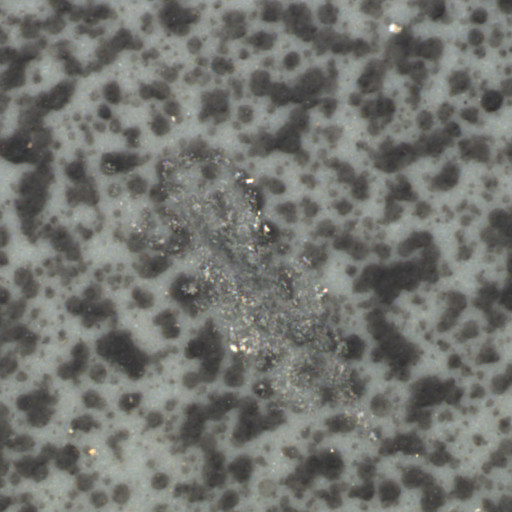}\\[-2pt]
        \includegraphics[width=0.49\cw]{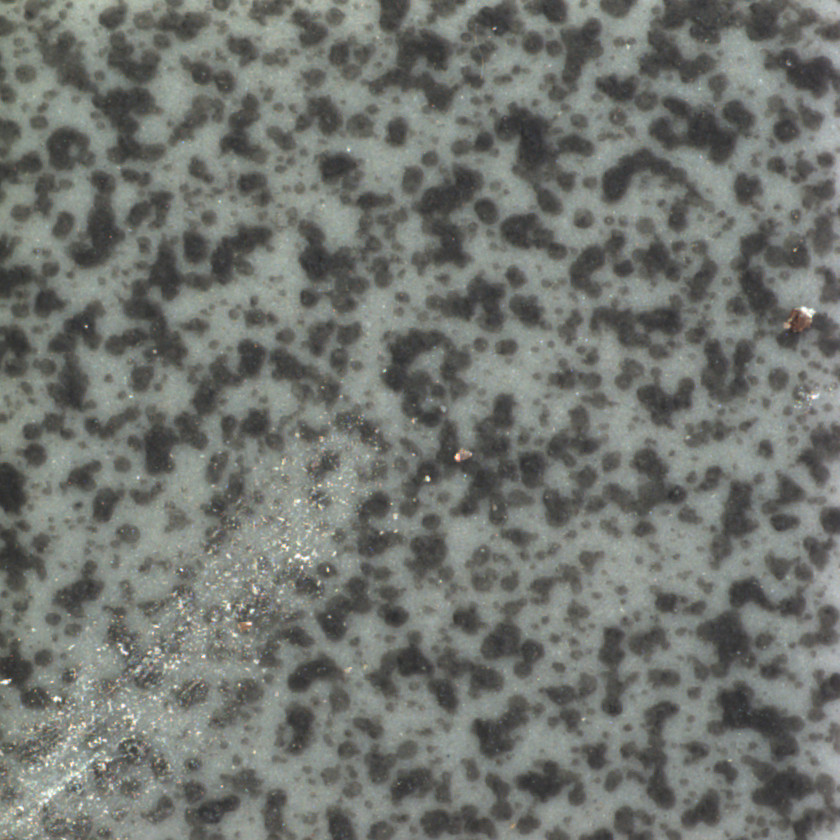} &
        \includegraphics[width=0.49\cw]{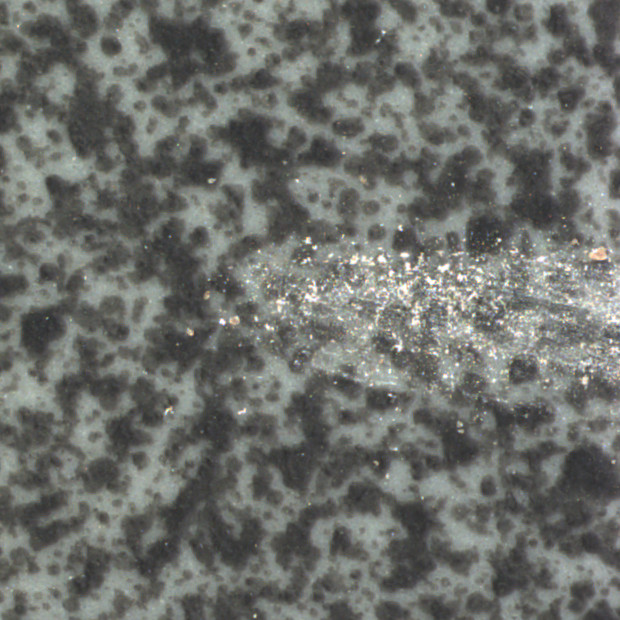}
    \end{tabular}} &
    \includegraphics[width=\cw]{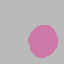}\\[-1.5pt]
    \includegraphics[width=\cw]{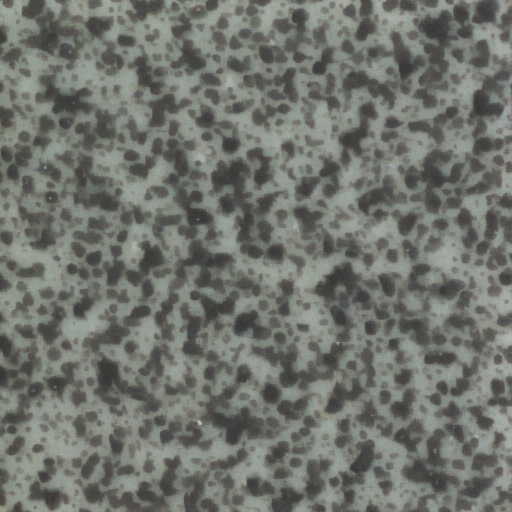} &
    \includegraphics[width=\cw]{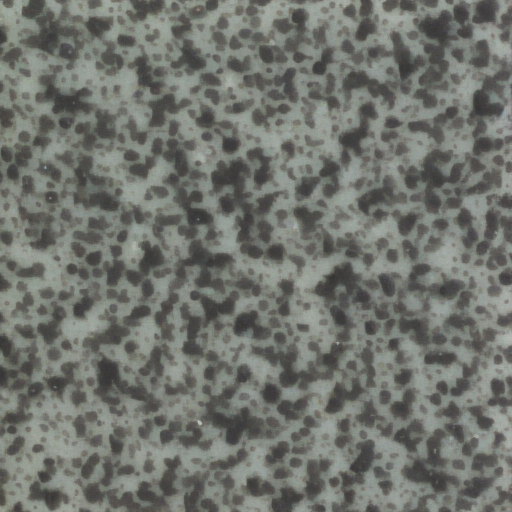} &
    \includegraphics[width=\cw]{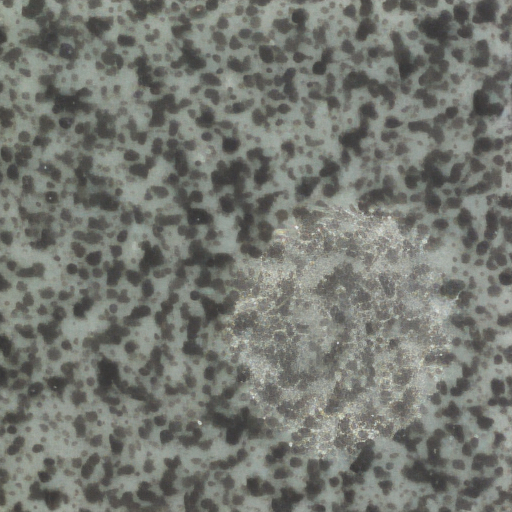}%
    \rlap{\raisebox{0.867\cw}{\smash{\hspace*{-2.64\cw}\includegraphics[origin=c,height=0.2\cw,angle=-90]{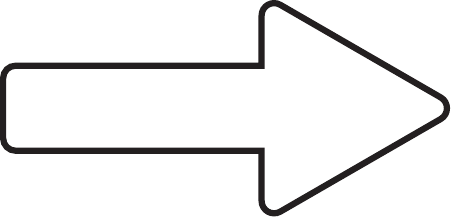}}}}%
    \rlap{\raisebox{0.4\cw}{\smash{\hspace*{-2.23\cw}\includegraphics[origin=c,height=0.2\cw]{images/edit_simple/arrow.pdf}}}}%
    \rlap{\raisebox{0.867\cw}{\smash{\hspace*{-0.62\cw}\includegraphics[origin=c,height=0.2\cw,angle=-90]{images/edit_simple/arrow.pdf}}}}%
    \rlap{\raisebox{0.4\cw}{\smash{\hspace*{-1.19\cw}\includegraphics[origin=c,height=0.2\cw]{images/edit_simple/arrow.pdf}}}}%
    \rlap{\raisebox{0.867\cw}{\smash{\hspace*{-1.23\cw}\includegraphics[origin=c,height=0.2\cw,angle=-45]{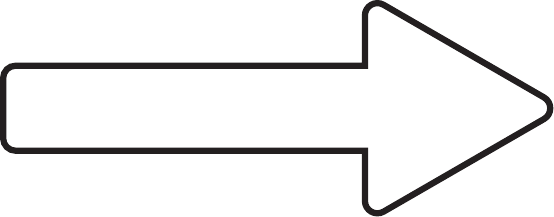}}}}%
    \\[-0.25ex]
    
    \fighlabel{VAE Representation} &
    \fighlabel{Inverted} &
    \fighlabel{Result}
  \end{tabular}
  \caption{\label{fig:edit_simple}%
    Image editing visualization, showing: the original image, irregular feature examples from the dataset, desired edit mask, the image after being encoded and decoded by the VAE of SD, the result of our diffusion model for the inverted noise, and the result of our noise-mixing synthesis.%
  }
  \Description{The Figure shows the results at intermediary steps of our image editing.}
\end{figure}

\begin{figure}
  \centering%
  \setlength{\cw}{0.2333\linewidth}%
  \setlength{\tabcolsep}{+0.003\linewidth}%
  \renewcommand{\arraystretch}{0.3}%
  \begin{tabular}{cccc}
    \figvlabel{Photograph}%
    \includegraphics[width=\cw]{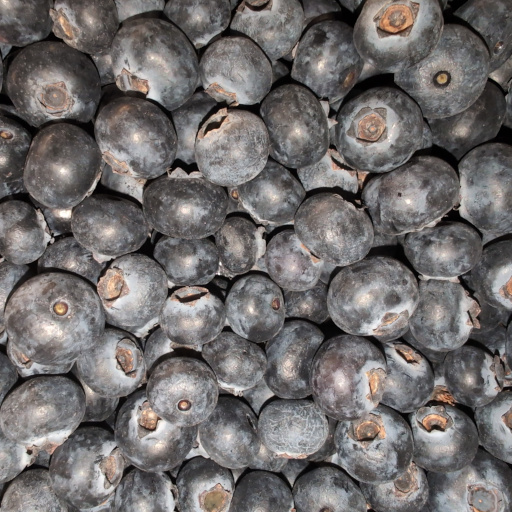} &
    \includegraphics[width=\cw]{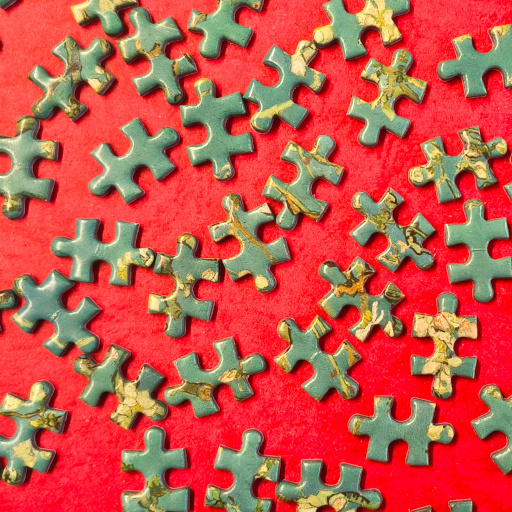} &
    \includegraphics[width=\cw]{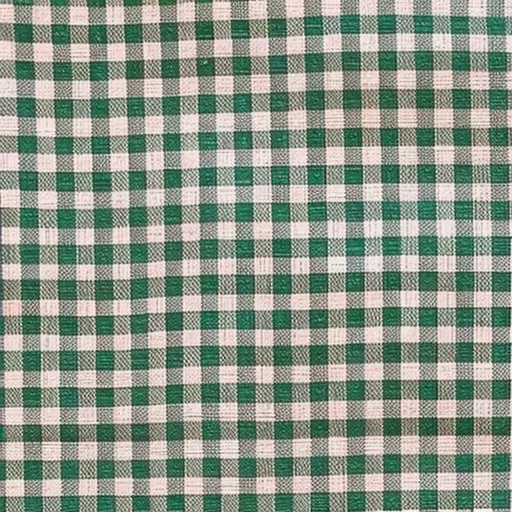} &
    \includegraphics[width=\cw]{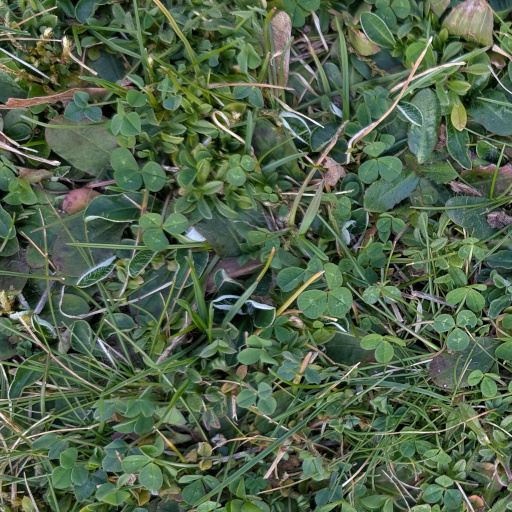} \\
    
    \figvlabel{Labels}%
    \includegraphics[width=\cw]{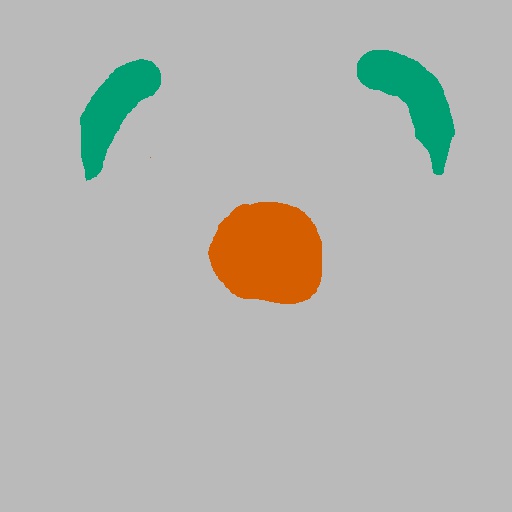} &
    \includegraphics[width=\cw]{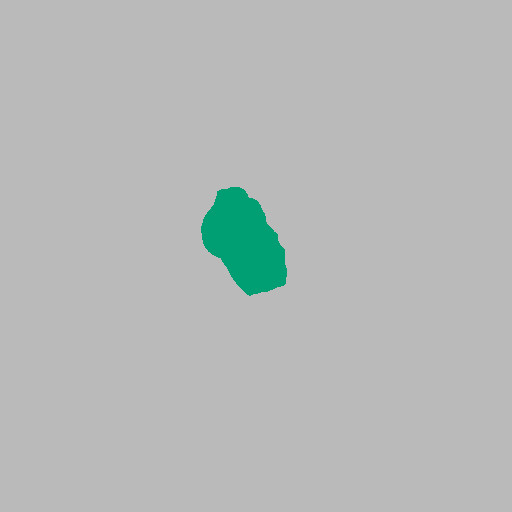} &
    \includegraphics[width=\cw]{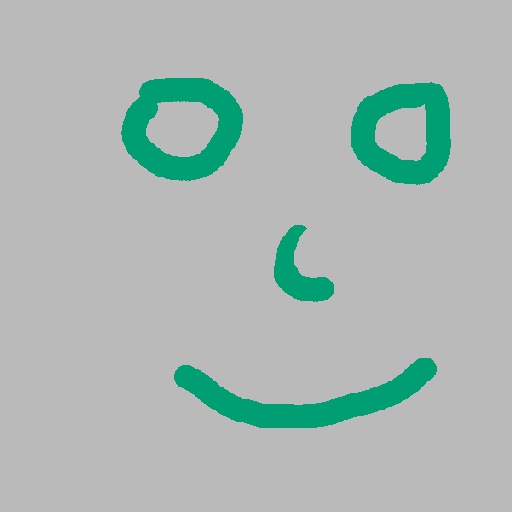} &
    \includegraphics[width=\cw]{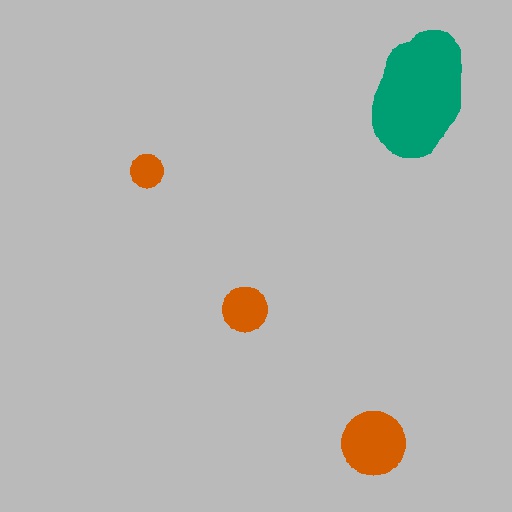} \\

    \figvlabel{\cite{wang2022texture}}%
    \includegraphics[width=\cw]{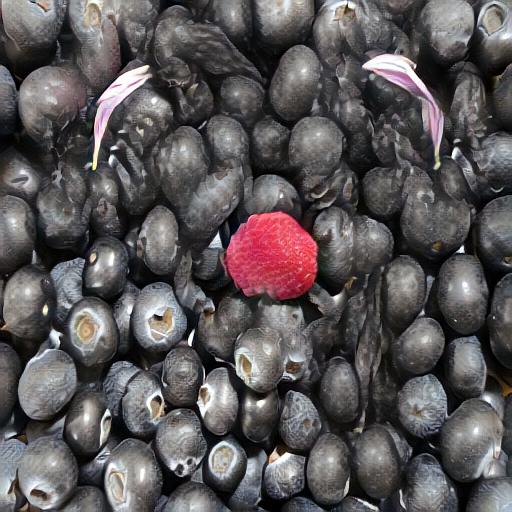} &
    \includegraphics[width=\cw]{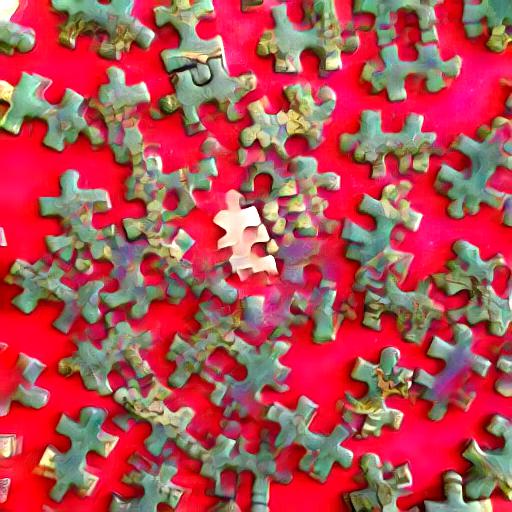} &
    \includegraphics[width=\cw]{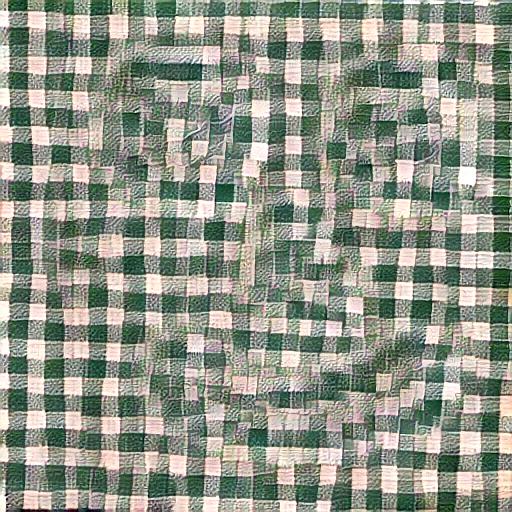} &
    \includegraphics[width=\cw]{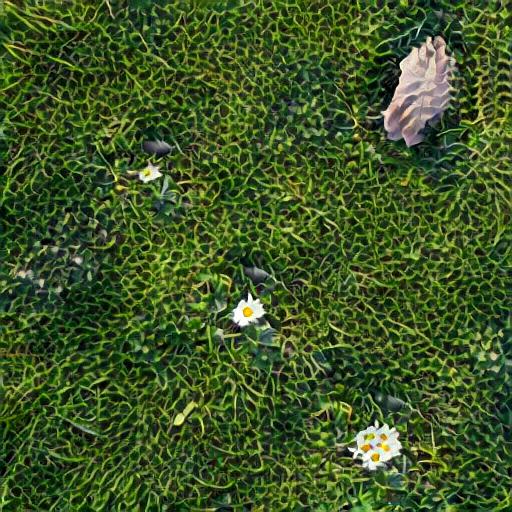} \\
    
    \figvlabel{Ours}%
    \includegraphics[width=\cw]{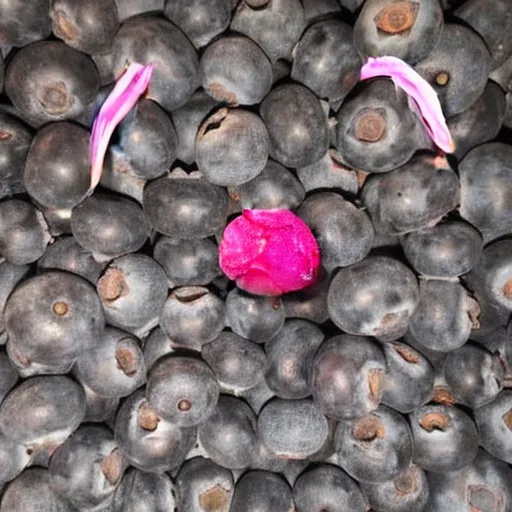} &
    \includegraphics[width=\cw]{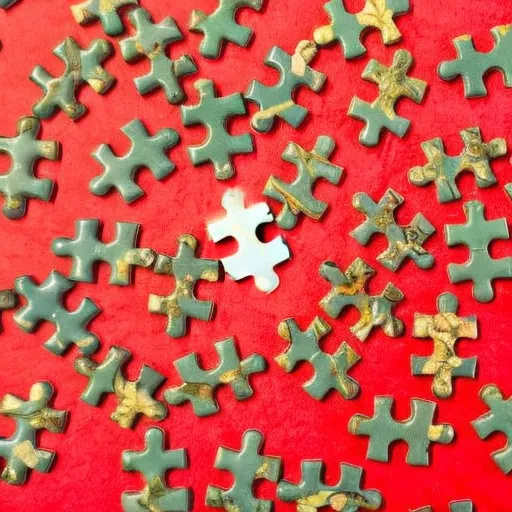} &
    \includegraphics[width=\cw]{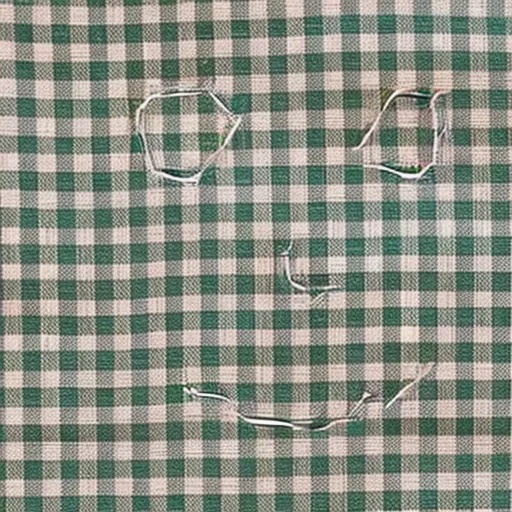} &
    \includegraphics[width=\cw]{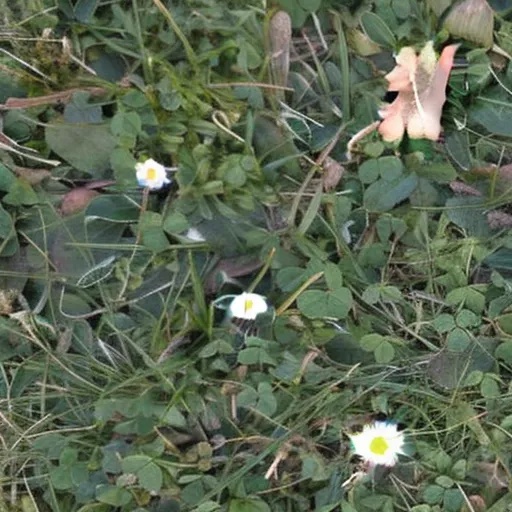} \\
  \end{tabular}
  \caption{\label{fig:edit_results}%
    Editing results with our noise-mixing approach. The rows depict top to bottom: a real photograph, the conditioning label-map, the result obtained with TextureReformer~\cite{wang2022texture}, and our edit.
  }
  \Description{Image showing our result on semantic multi-feature edit.}
\end{figure}

The possibility to edit existing images is a crucial capability in the context of generating textures with non-stationary features.
Moreover, the edited region should not simply \emph{replace} the previous texture, but remain faithful to the original structure, as highlighted in Fig.~\ref{fig:edit_simple} and Fig.~\ref{fig:edit_results}.
Most diffusion-based editing frameworks rely on diffusion inversion~\cite{hertz2023prompt, wallace2023edict, zhang2025exact}, \ie, an input image is reproduced by the diffusion model by finding the noise $\vect{z}_N$ that produces the image when solving the underlying ODE.
This is achieved by running the noise estimation model $\bfeps(\vect{z}_i, \varnothing; \sigma_i)$ with the null conditioning $\varnothing$ for $N$ steps, finding the noise that must be added at each iteration. That is $\vect{z}_{i+1} = \vect{z}_i + (\sigma_{i+1}-\sigma_i)\bfeps(\vect{z}_i, \varnothing, \sigma_i)$, resting on the assumption that $\bfeps(\vect{z}_i, \varnothing, \sigma_i) \sim \bfeps(\vect{z}_{i+1}, \varnothing; \sigma_{i+1})$. Note that we write $\sigma_i := \sigma(t_i)$ for brevity.
\\ \noindent After obtaining $\vect{z}_N$, the edited image can be synthesized by running the generative process from the found noise with the new desired conditioning label map.
We opt for the fixed-point iteration method~\cite{pan2023effective} for diffusion inversion, and implement it into the the EDM formulation. However, we found that the inversion is unstable when the second-order Heun solver is used. This is because each denoising step is harder to invert and errors accumulate; therefore, we use the Euler solver in the context of image editing.

\subsubsection{Localized updates.}\;\;%

Since the painted features are localized, there is an expectation that the image remains as close as possible to the original outside of the edited region; the painted feature should also be compatible with the initial texture (for example, the sugar spill should not alter the texture below, Fig.~\ref{fig:ablation_nm}).
We propose to solve this via \textbf{noise-mixing}: during inversion, we save the intermediate noise estimates $\bfeps(\vect{z}_i, \varnothing; \sigma_i)$, for each level $\sigma_i$.
Then, to generate the new image from the inverted noise $\vect{z}_N$, we mix the noise estimates saved during inversion with the new noise estimates ($\hat{\bfeps})$ that are conditioned by the label map $\vect{c}$: 
\begin{equation}
    \hat{\vect{z}}_{i-1} = \hat{\vect{z}}_{i} + (\sigma_{i-1} - \sigma_i)\texttt{mix}(\hat{\bfeps}(\hat{\vect{z}}_i, \vect{c}; \sigma_i), \bfeps(\vect{z}_i, \varnothing; \sigma_i), i) \;\;.
\end{equation}
Where $\hat{\bfeps}$ denotes the classifier-free guided direction: $\hat{\bfeps}(\vect{z}, \vect{c}, \sigma) = \bfeps(\vect{z}, \varnothing, \sigma) + \gamma(\bfeps(\vect{z}, \vect{c}, \sigma)-\bfeps(\vect{z}, \varnothing, \sigma))$, with guidance $\gamma = 4$.

To strike a balance between preserving the original structure in the edited regions and being faithful to the new condition, we mix the scores adaptively depending on the step in the backward diffusion process:

\begin{align}
    \texttt{mix}(\hat{\bfeps}_i, \bfeps_i, i) &:= \hat{\bfeps}_i + (\bfeps_i - \hat{\bfeps}_i)\left(\frac{i}{N}\right)^\alpha .
\end{align}
The influence of the original (unconditional) noise estimates $\bfeps_i$ decreases from $1$ to almost $0$ over the $N$ steps. $\alpha$ controls how fast the mixing factor decays, and it is set to $0.3$ in our experiments.
\TAedit{In order to minimize the alteration of the texture outside the editing region, we set the mixing factor to $0$ for the background (outside the editing mask).}

\subsubsection{Feature transfer.}\;\;%

Our noise-mixing editing method lends itself to another important capability: feature transfer.
We demonstrate this function in Fig.~\ref{fig:edit_transfer}, where we transfer features from their native class to another textures class from MVTec, as well as to a real image of the author's desk.
To transfer a feature from a texture class to a target image that is out-of-distribution (OOD), we first invert the image using the trained diffusion model.
When the image is very different from the training distribution, the inversion process needs more steps to reconstruct the image with high accuracy; we use 250 in our experiments.
We then perform the steps analogous to image editing, mixing the noise estimates from inversion with the predictions obtained under the desired label map $\vect{c}$.
Thanks to the classifier-free guidance mechanism, the direction of the noise estimates incorporates the difference between the feature and the normal texture, which minimizes the contamination of normal appearance from the training texture to the OOD target image.

\subsection{Stationary infinite texture generation}
\label{sec:infinity}

\begin{figure}
  \centering
  \mbox{} \hfill
  
  \setlength{\cw}{0.18\linewidth}
  \setlength{\tabcolsep}{+0.003\linewidth}
  \begin{tabular}{ccccc}
    \figvlabel{\footnotesize Independent}\!\figvlabel{\footnotesize\strut}%
    \includegraphics[width=\cw]{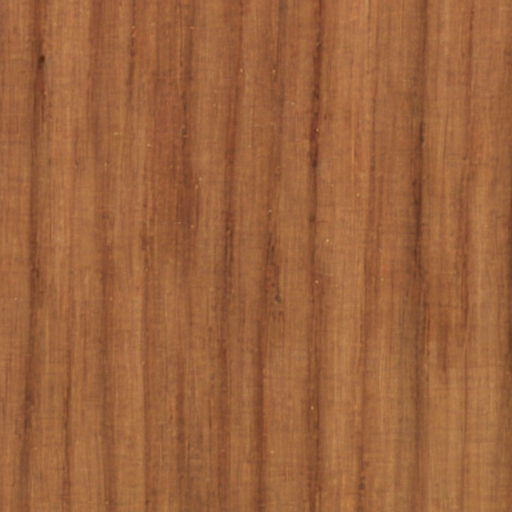} &
    \includegraphics[width=\cw]{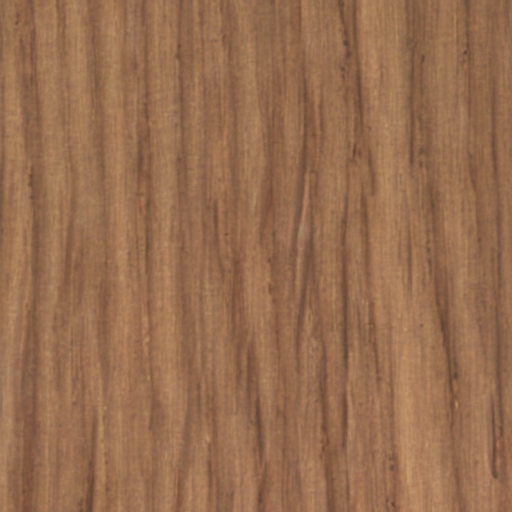} &
    \includegraphics[width=\cw]{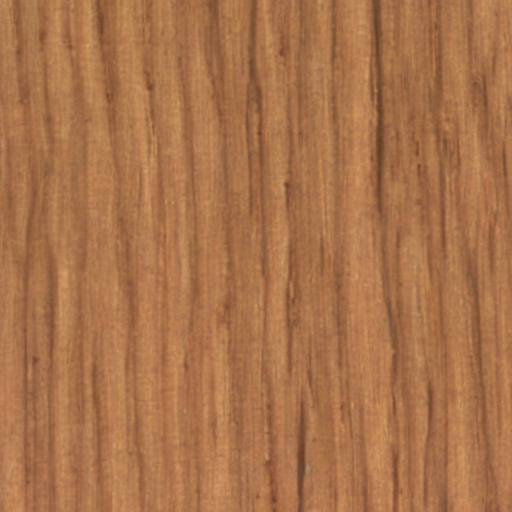} &
    \includegraphics[width=\cw]{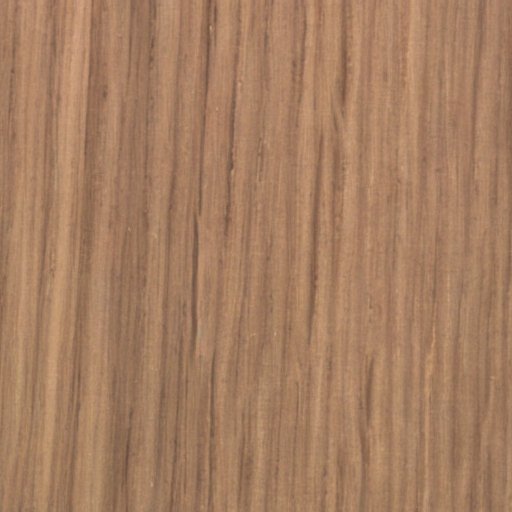} &
    \includegraphics[width=\cw]{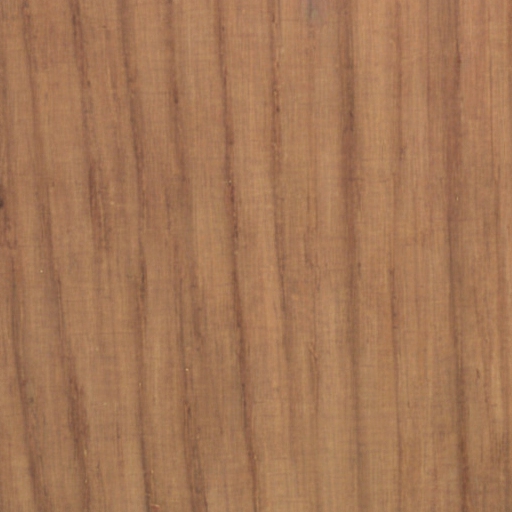} \\[-2pt]

    \figvlabel{\footnotesize Same mean}\!\figvlabel{\footnotesize\strut}%
    \includegraphics[width=\cw]{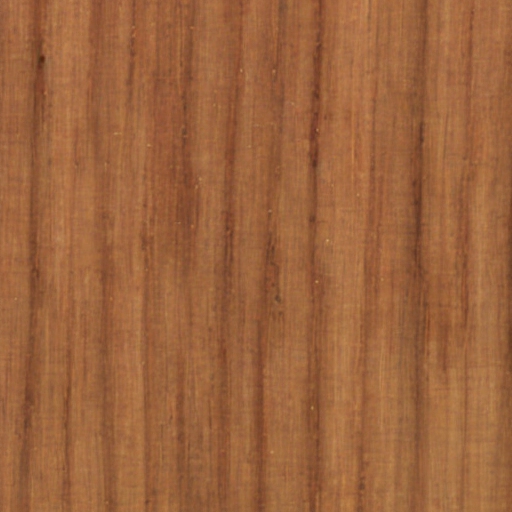} &
    \includegraphics[width=\cw]{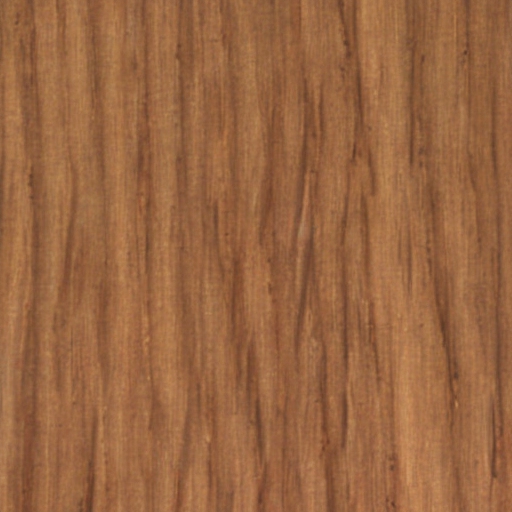} &
    \includegraphics[width=\cw]{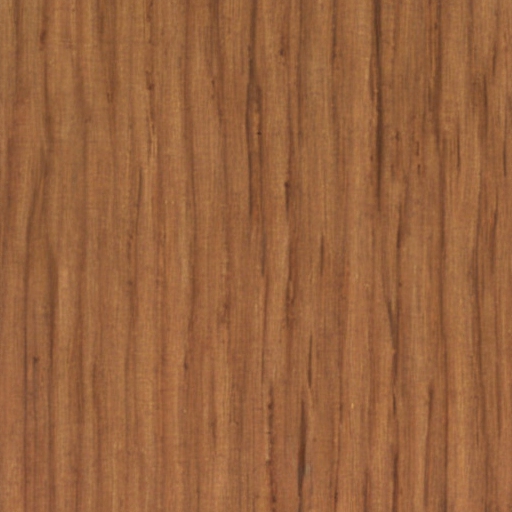} &
    \includegraphics[width=\cw]{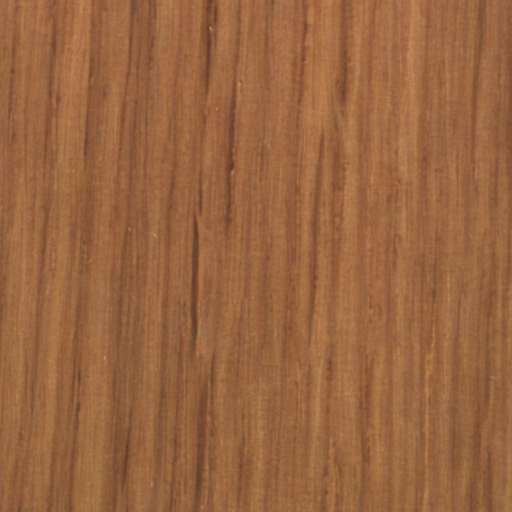} &
    \includegraphics[width=\cw]{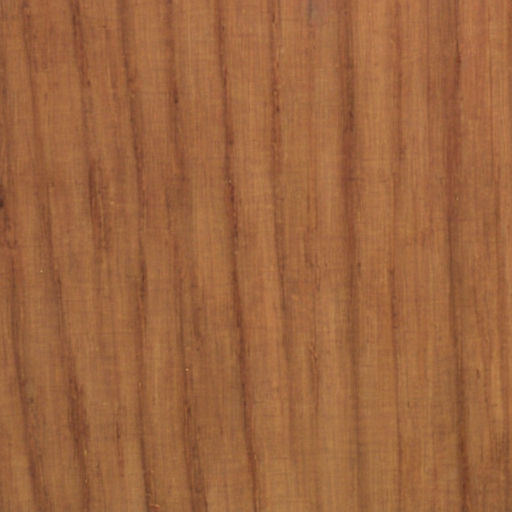} \\[-2pt]

    \figvlabel{\footnotesize Lanczos}\!\figvlabel{\footnotesize\strut$f_c=0.1$}%
    \includegraphics[width=\cw]{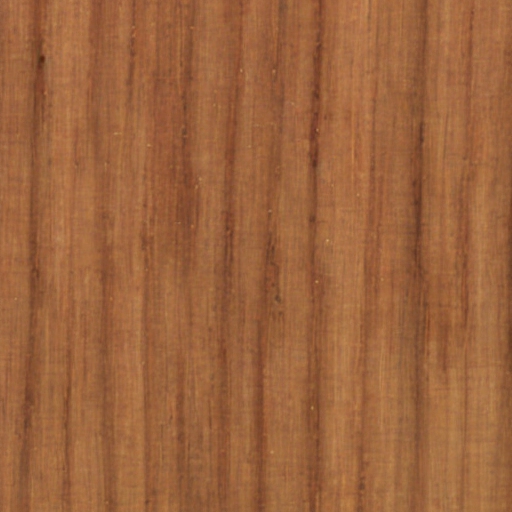} &
    \includegraphics[width=\cw]{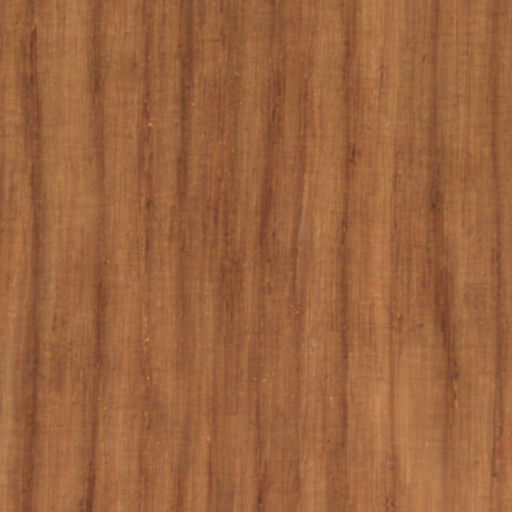} &
    \includegraphics[width=\cw]{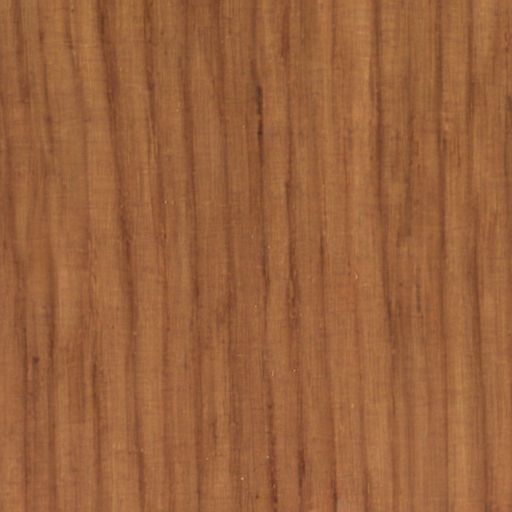} &
    \includegraphics[width=\cw]{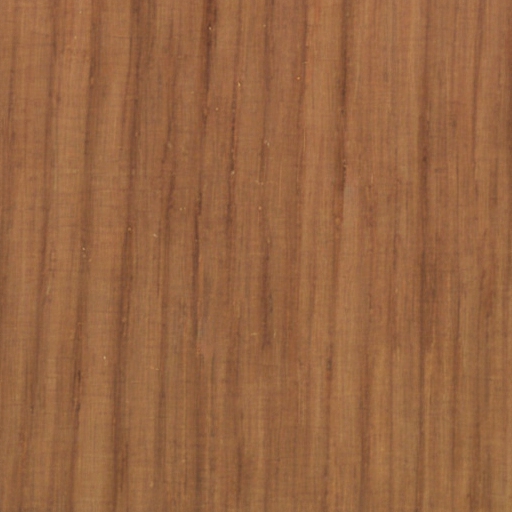} &
    \includegraphics[width=\cw]{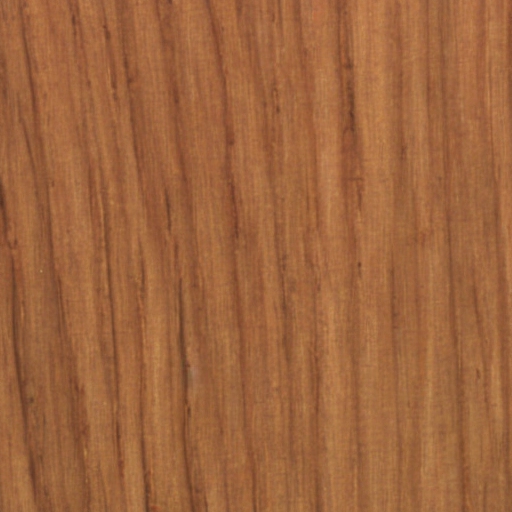} \\[-2pt]

    \figvlabel{\footnotesize Lanczos}\!\figvlabel{\footnotesize\strut$f_c=0.2$}%
    \includegraphics[width=\cw]{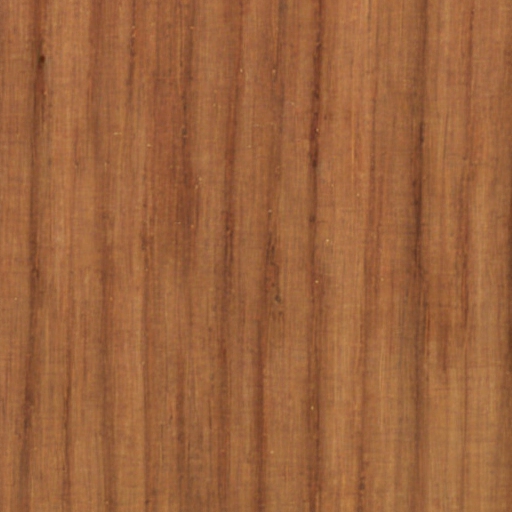} &
    \includegraphics[width=\cw]{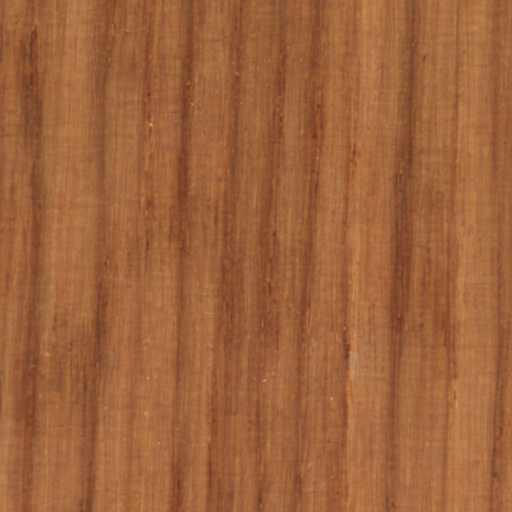} &
    \includegraphics[width=\cw]{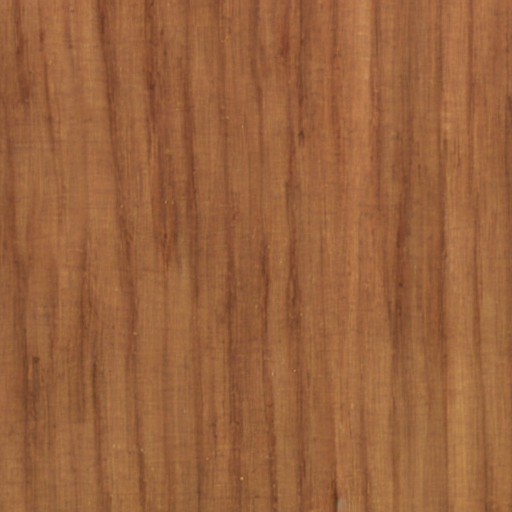} &
    \includegraphics[width=\cw]{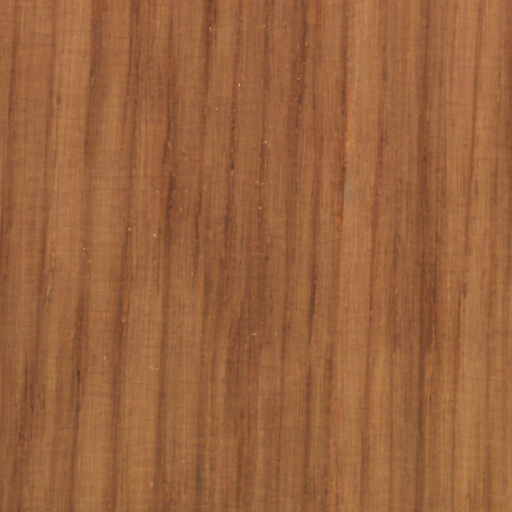} &
    \includegraphics[width=\cw]{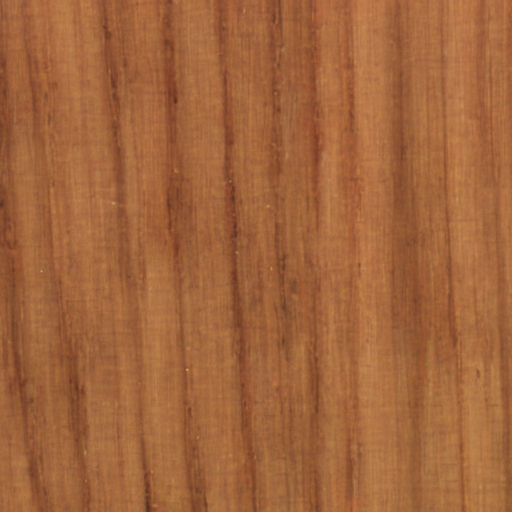} \\

  \end{tabular} 
  \caption{\label{fig:low_freq}%
    Comparison between images sampled independently (\emph{top}) and different uniformization strategies. Rows 2 through 4 attempt to make all images follow the style of the first column, using the methods explained in Sec.~\ref{sec:infinity}.%
  }
  \Description{The figure compares different preprocessing methods on the initial noise and how it influences the consistency of the generated textures.}
\end{figure}

\begin{figure}
  \centering%
  \newlength{\ch}%
  \setlength{\ch}{0.11\linewidth}%
  \setlength{\tabcolsep}{0.002\linewidth}%
  \begin{tabular}{cc}

  \includegraphics[height=\ch]{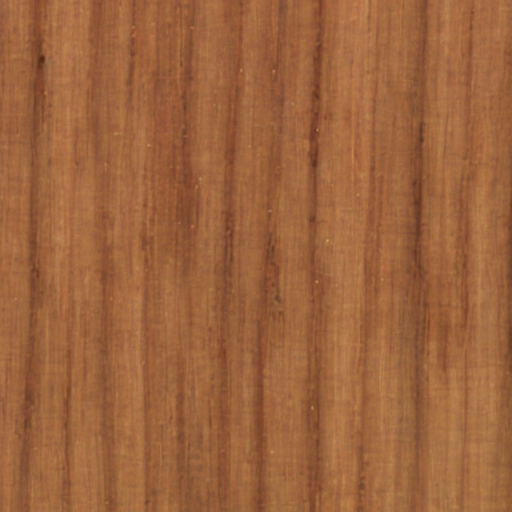} &
  \includegraphics[height=\ch]{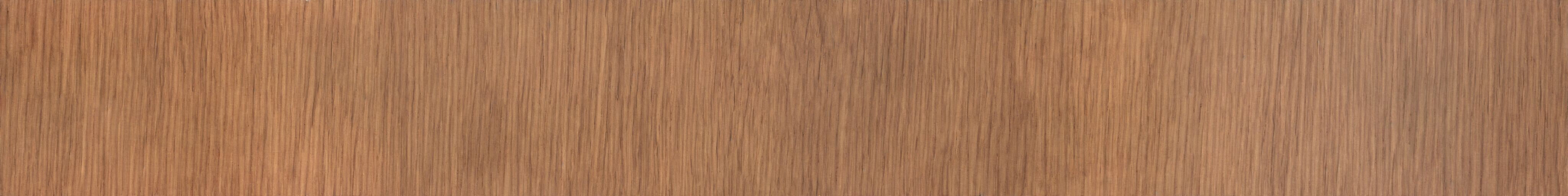} \\
  
  \includegraphics[height=\ch]{images/uniformization/large_prototype.jpg} &
  \includegraphics[height=\ch]{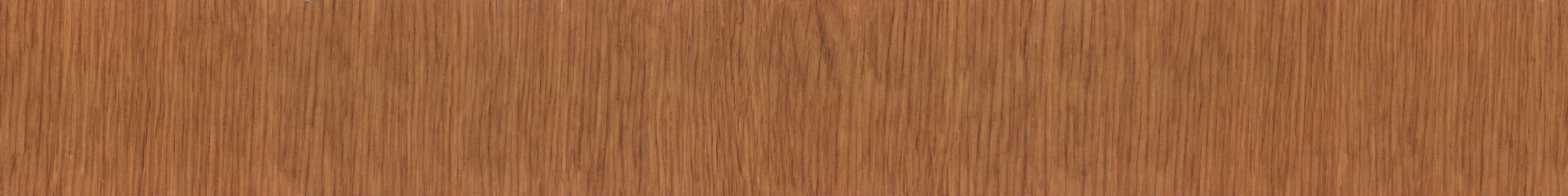} \\
  
  \includegraphics[height=\ch]{images/uniformization/large_prototype.jpg} &
  \includegraphics[height=\ch]{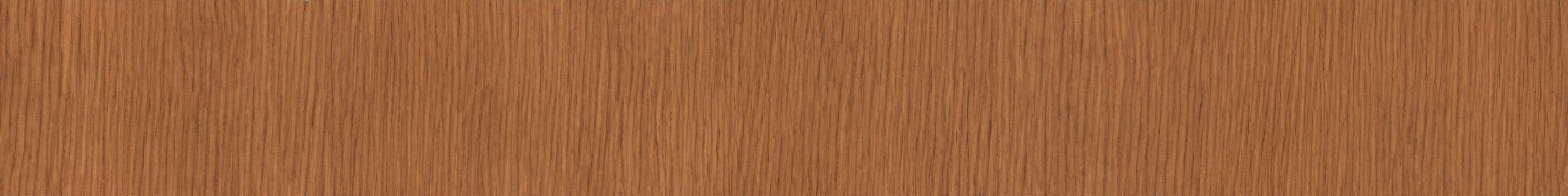} \\

  \fighlabel{Style} & \fighlabel{Large generated texture}

  \end{tabular}
  \caption{\label{fig:full_examples}%
    Synthesized $1024\times 8192$ textures. The top texture uses pure white noise; the middle image is obtained by copying the LF components; the third image is obtained using our noise-uniformization technique. Note that white noise yields a non-stationary texture that deviates from the style-prototype.%
  }
    \Description{Large generated textures using white noise, or preprocessed noise.}
\end{figure}

\begin{figure*}
  \centering%
  \setlength{\cw}{0.1225\linewidth}%
  \setlength{\tabcolsep}{+0.001\linewidth}%
  \begin{tabular}{cccccccc}
            \includegraphics[width=\cw]{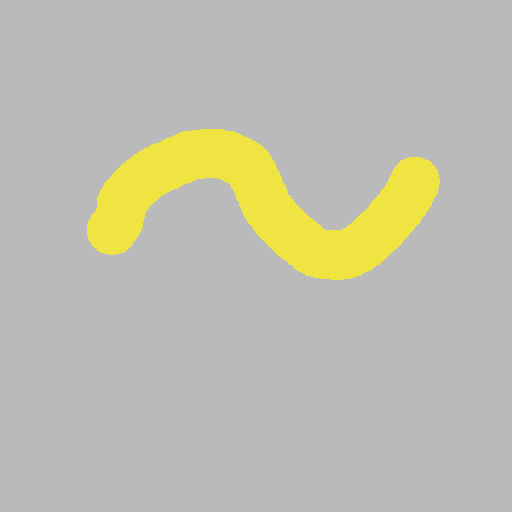 } &
            \includegraphics[width=\cw]{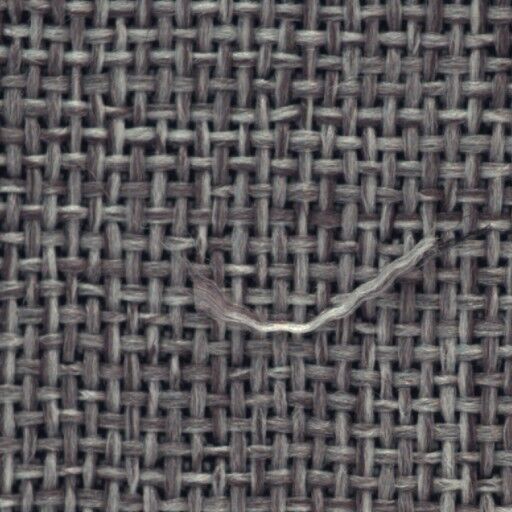} &
            \includegraphics[width=\cw]{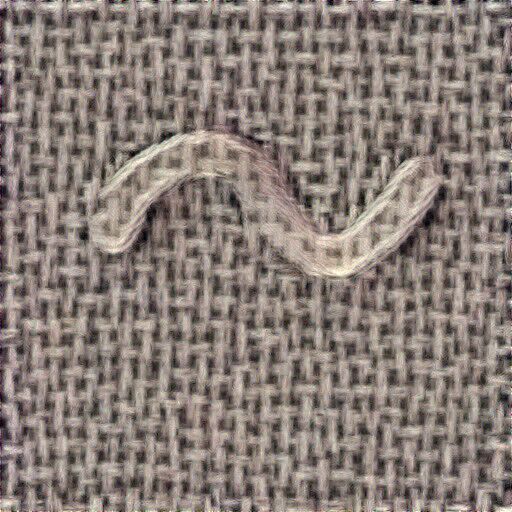} &
            \includegraphics[width=\cw]{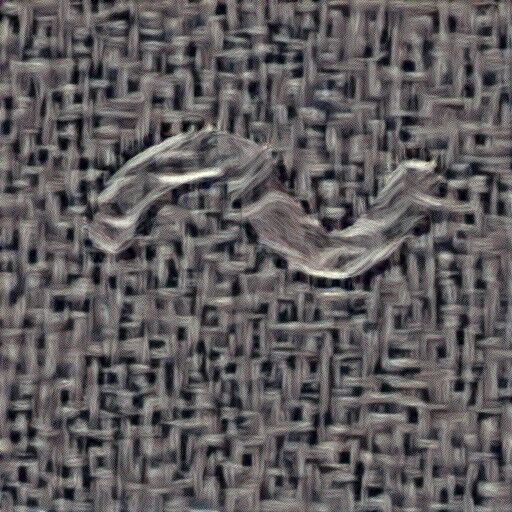} &
            \includegraphics[width=\cw]{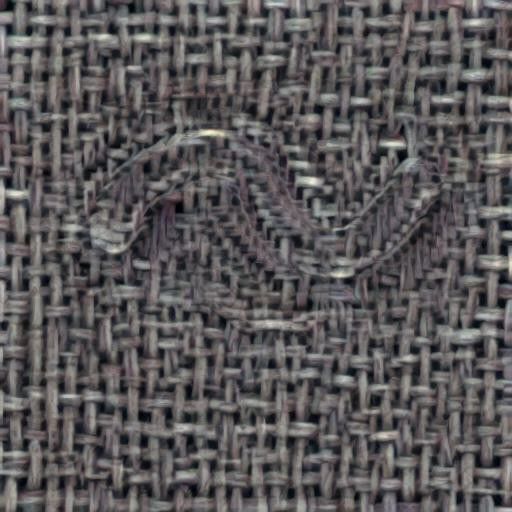} &
            \includegraphics[width=\cw]{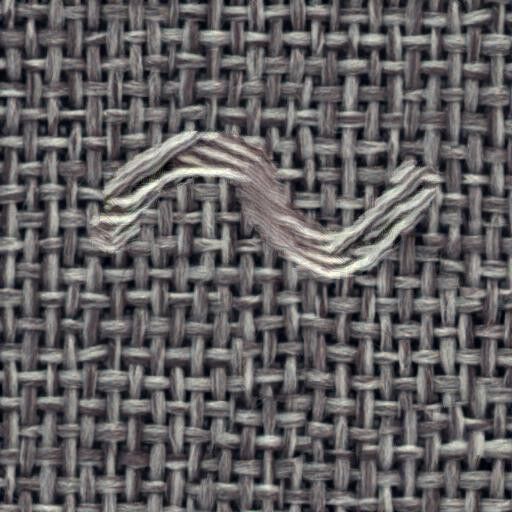} &
            \includegraphics[width=\cw]{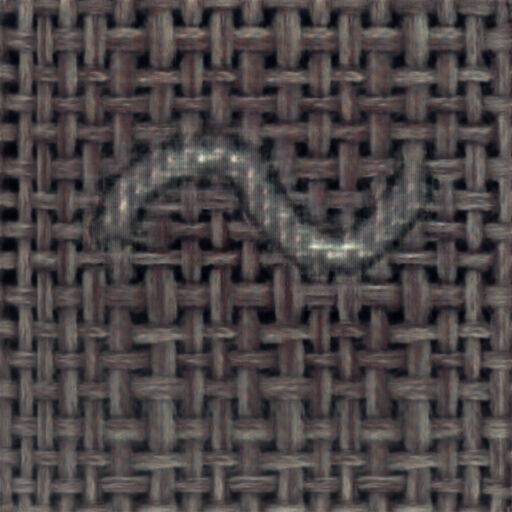} &
            \includegraphics[width=\cw]{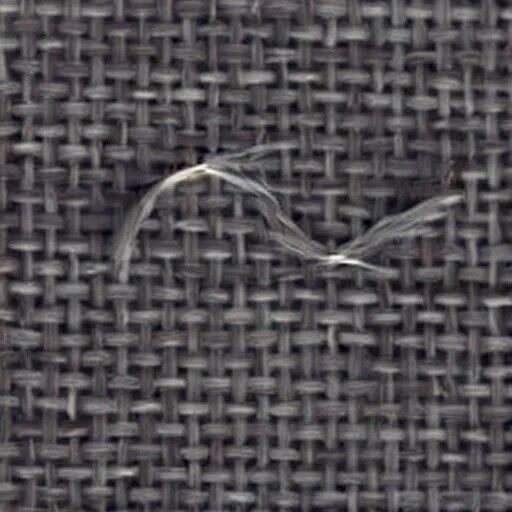} \\
        
            \includegraphics[width=\cw]{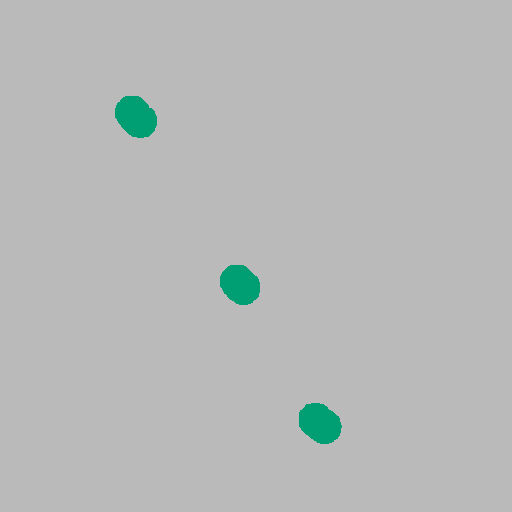} &
            \includegraphics[width=\cw]{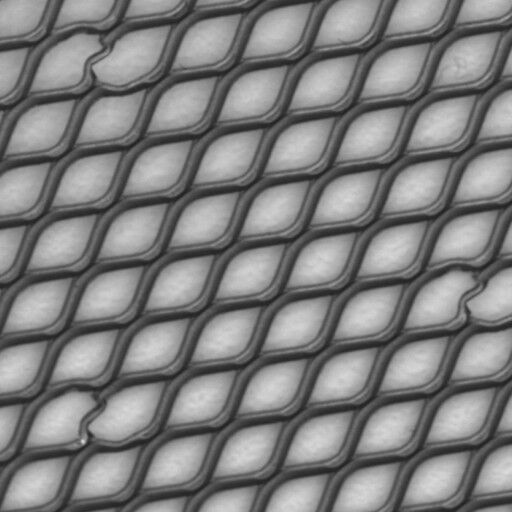} &
            \includegraphics[width=\cw]{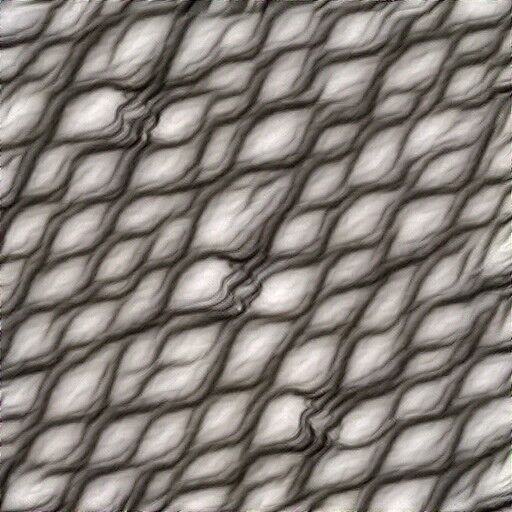} &
            \includegraphics[width=\cw]{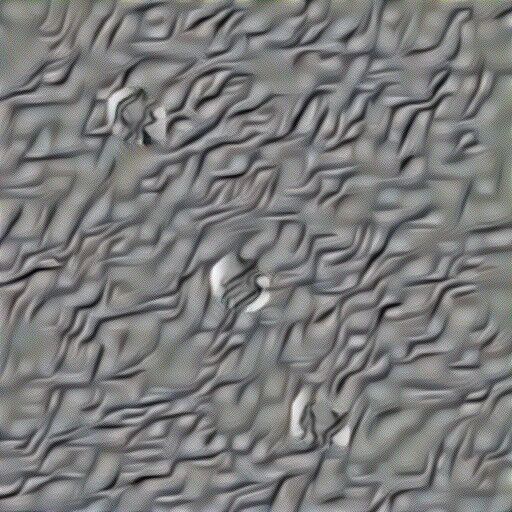} &
            \includegraphics[width=\cw]{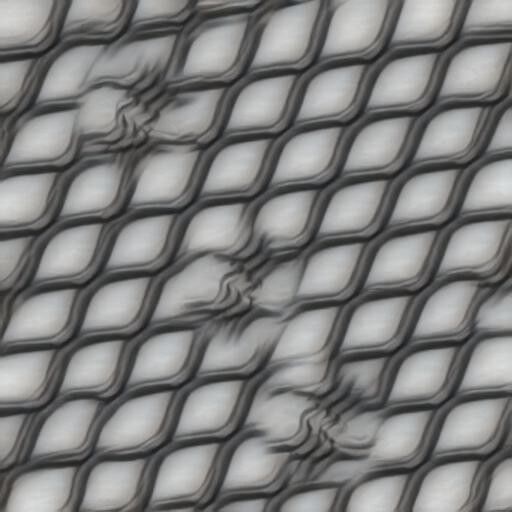} &
            \includegraphics[width=\cw]{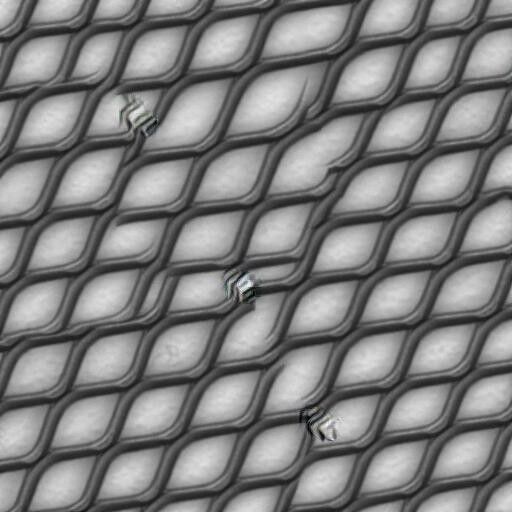} &
            \includegraphics[width=\cw]{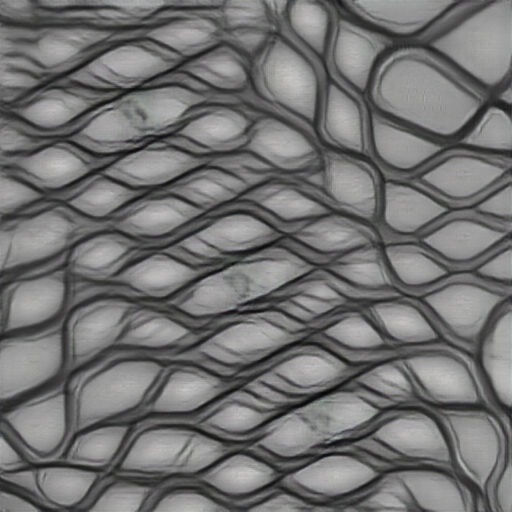} &
            \includegraphics[width=\cw]{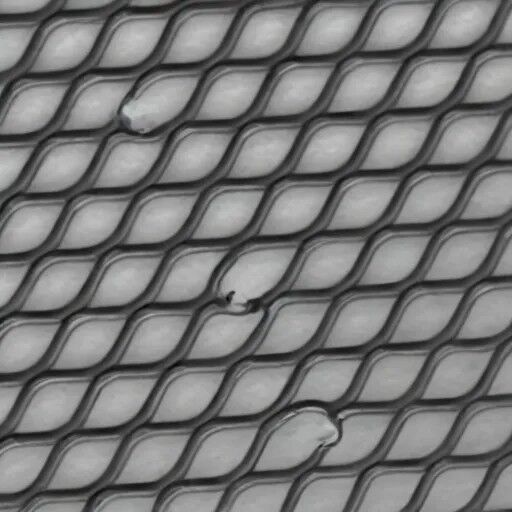} \\
        
            \includegraphics[width=\cw]{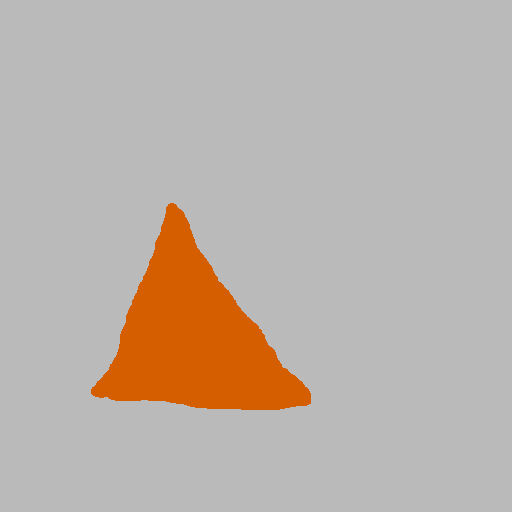} &
            \includegraphics[width=\cw]{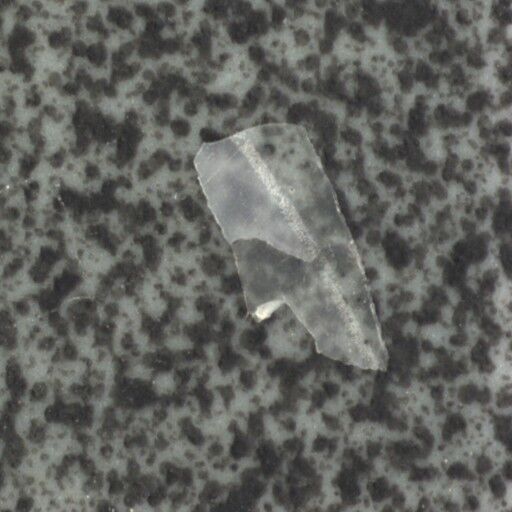} &
            \includegraphics[width=\cw]{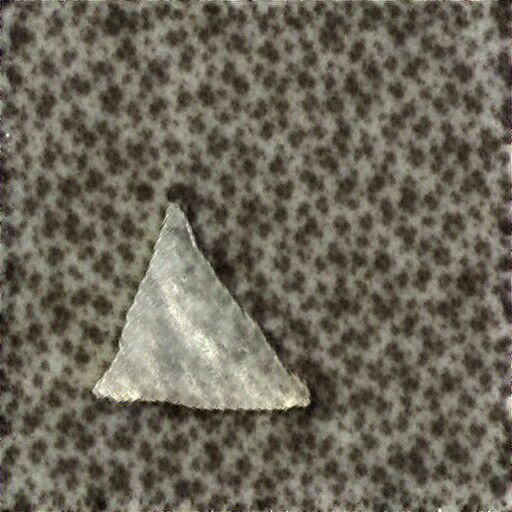} &
            \includegraphics[width=\cw]{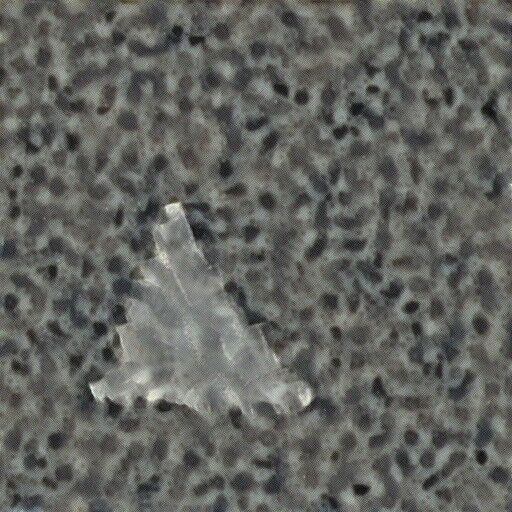} &
            \includegraphics[width=\cw]{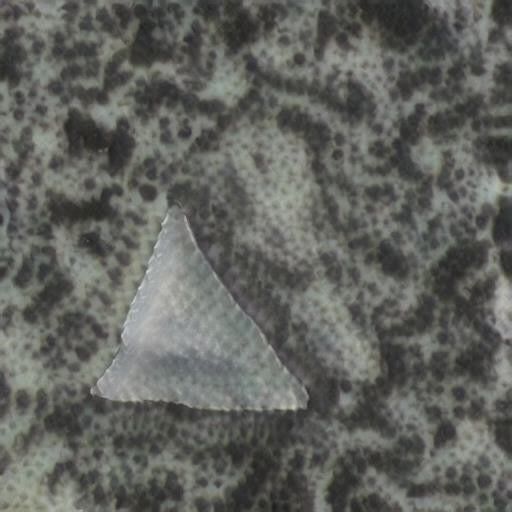} &
            \includegraphics[width=\cw]{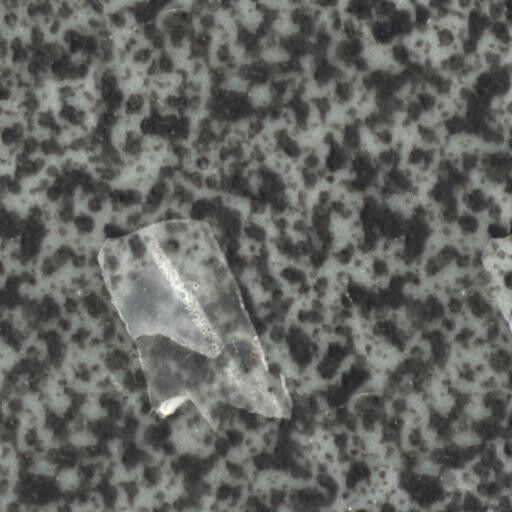} &
            \includegraphics[width=\cw]{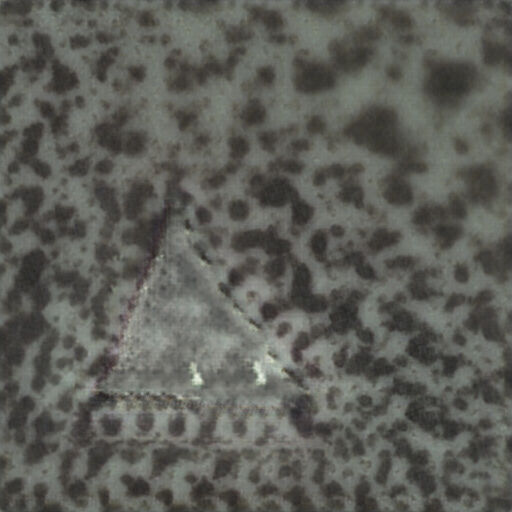} &
            \includegraphics[width=\cw]{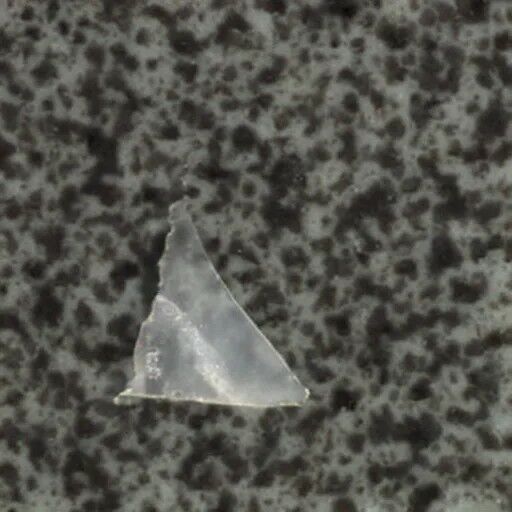} \\
            
            \includegraphics[width=\cw]{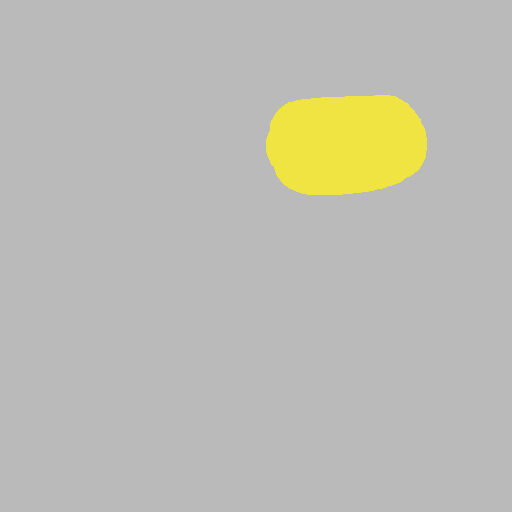} &
            \includegraphics[width=\cw]{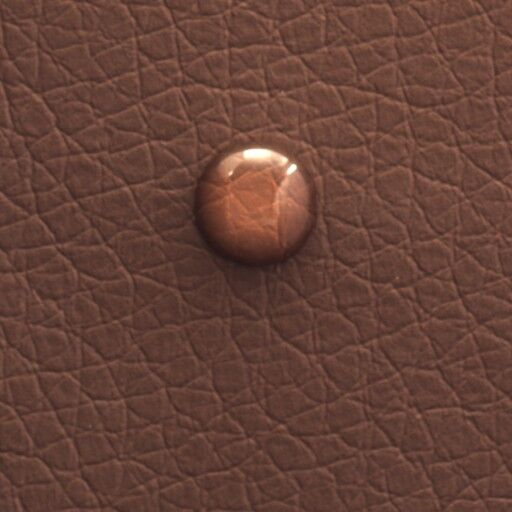} &
            \includegraphics[width=\cw]{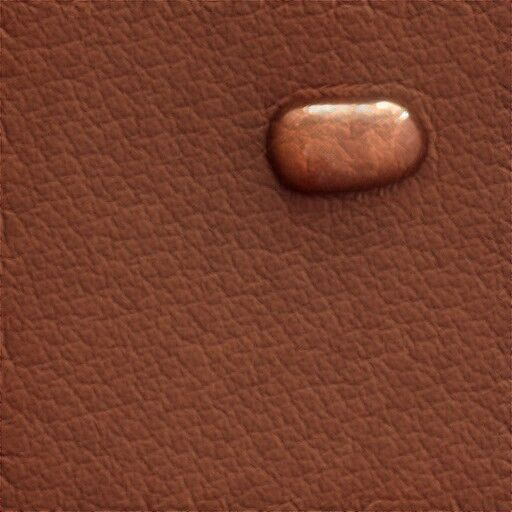} &
            \includegraphics[width=\cw]{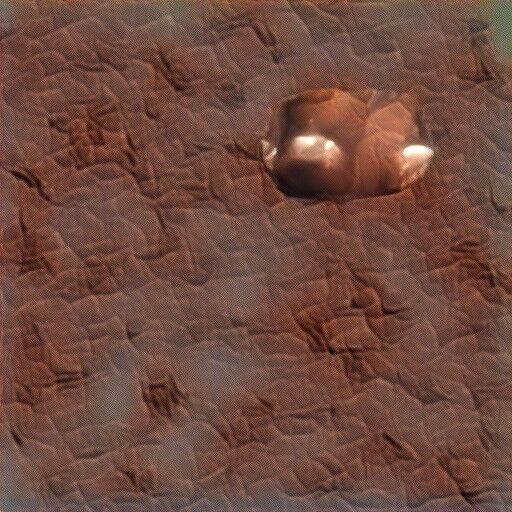} &
            \includegraphics[width=\cw]{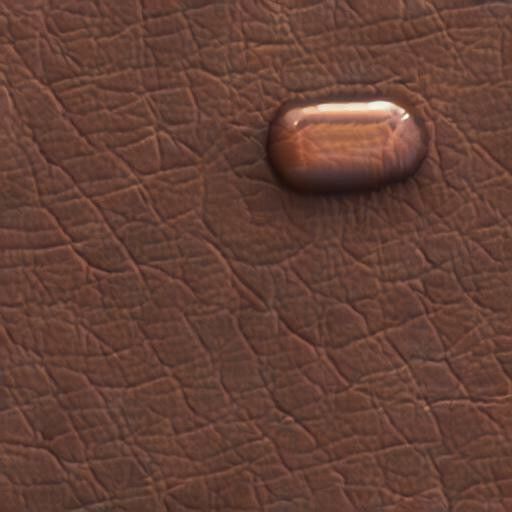} &
            \includegraphics[width=\cw]{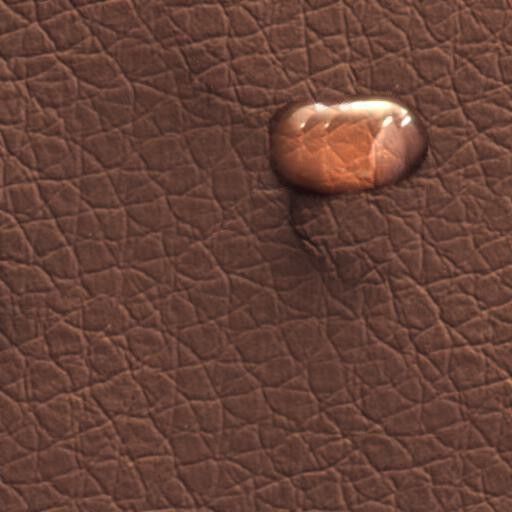} &
            \includegraphics[width=\cw]{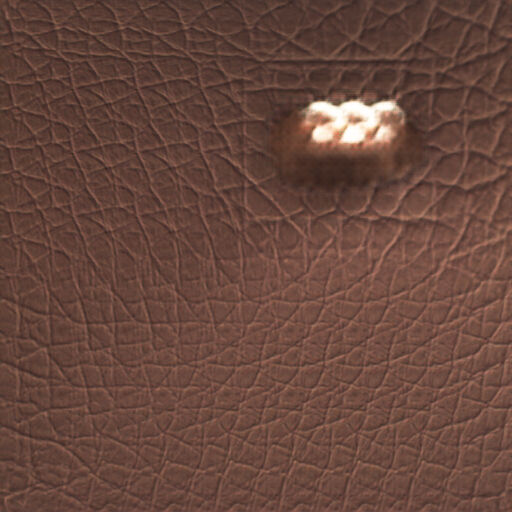} &
            \includegraphics[width=\cw]{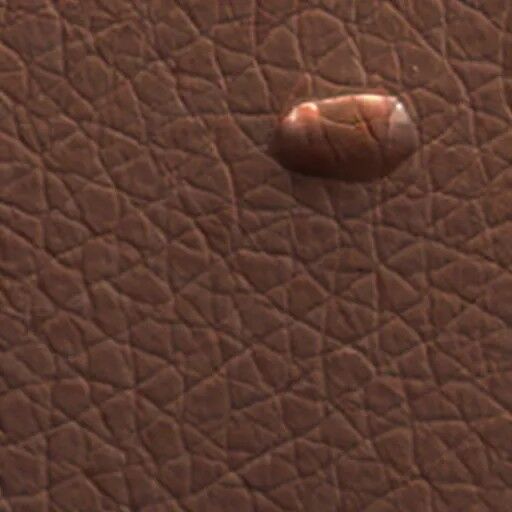} \\
            
            \includegraphics[width=\cw]{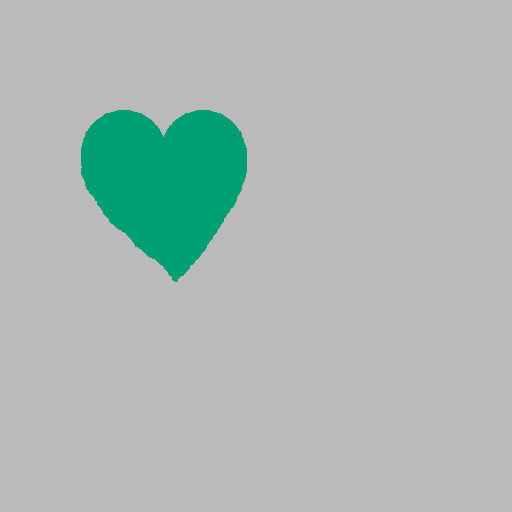} &
            \includegraphics[width=\cw]{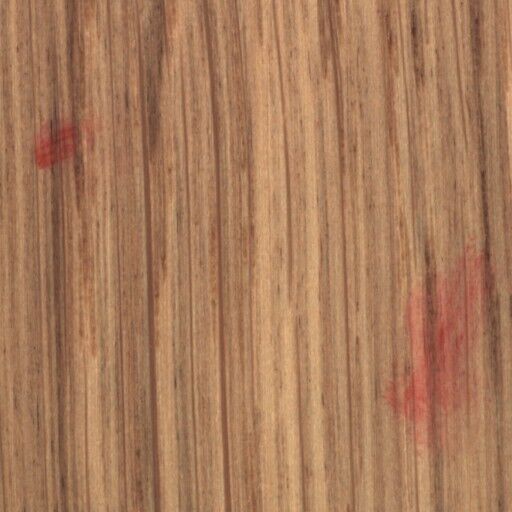} &
            \includegraphics[width=\cw]{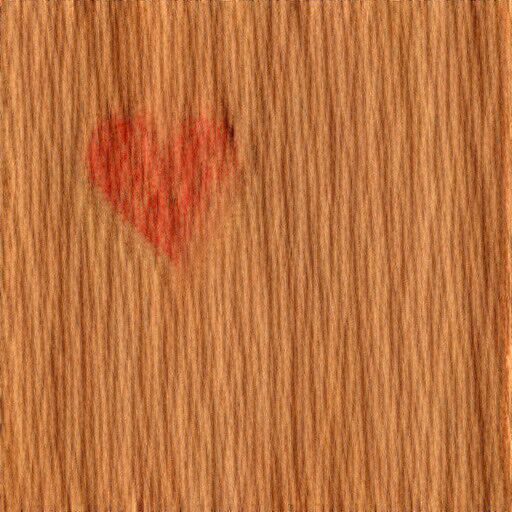} &
            \includegraphics[width=\cw]{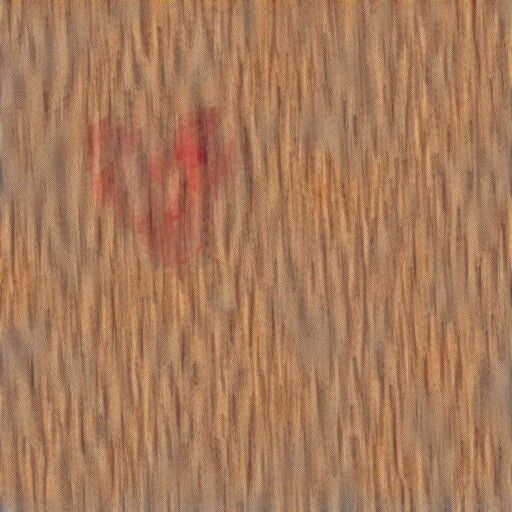} &
            \includegraphics[width=\cw]{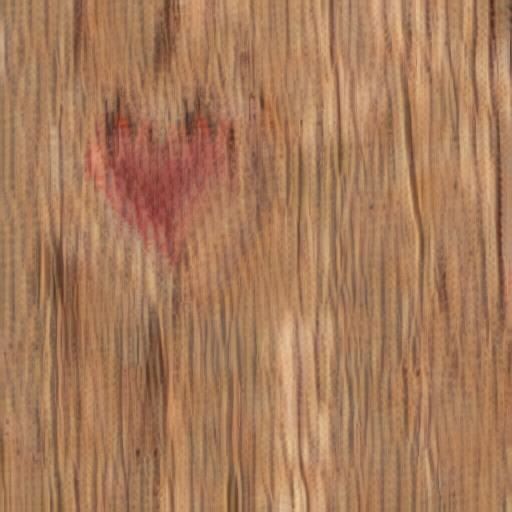} &
            \includegraphics[width=\cw]{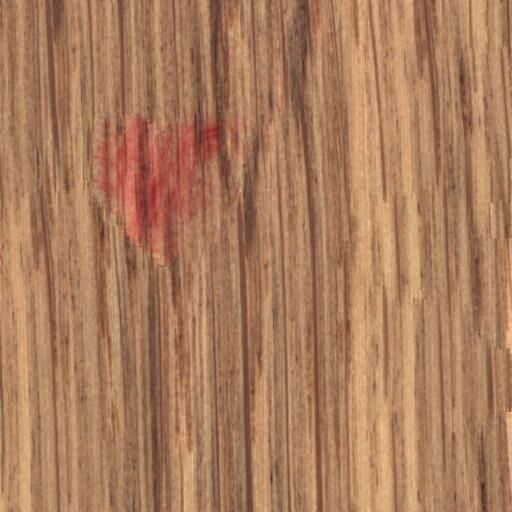} &
            \includegraphics[width=\cw]{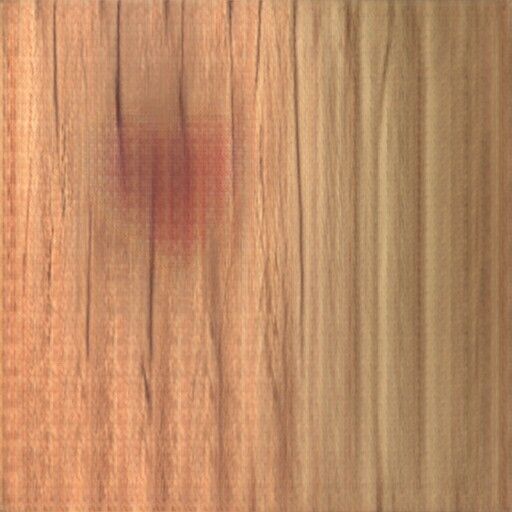} &
            \includegraphics[width=\cw]{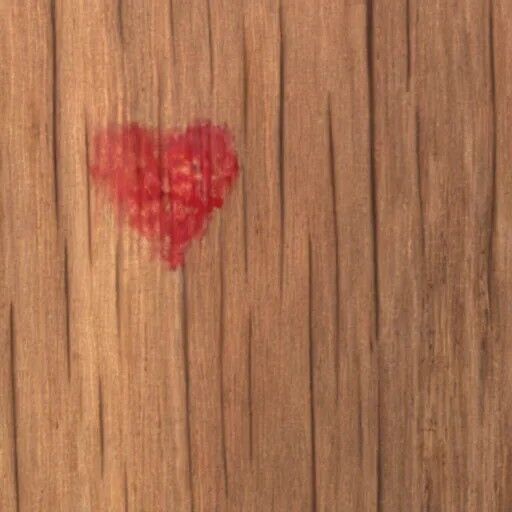} \\

            \fighlabel{Labels} &
            \fighlabel{Feature reference} &
            \fighlabel{Image analogies} &
            \fighlabel{Gatys et al.~\shortcite{gatys2016image}} &
            \fighlabel{Wang et al.~\shortcite{wang2022texture}} &
            \fighlabel{Zhou et al.~\shortcite{zhou2023neural}} &
            \fighlabel{Chen et al.~\shortcite{Chen2021GuidedIW}} &
            \fighlabel{Ours }\\
    \end{tabular}
  \caption{\label{fig:compare_synthesis_transposed}%
           Qualitative comparison for label-conditioned texture generation.  }
    \Description{The image compares the quality of the images generated by several baselines with our approach, with rows transposed to columns.}
\end{figure*}

Our method supports generating images at arbitrary resolutions and aspect ratios.
A naive approach to high-resolution synthesis would be to simply run the denoising model $\hat{\bfeps}(\vect{z}, \vect{c}, \sigma)$ starting with a noise tensor $\vect{z}_N$ significantly larger than the size of the training images. 
This approach does not work on generic image synthesis~\cite{he2023scalecrafter, haji2024elasticdiffusion} because it fails to retain the relationship between elements at a global level; however, textures are defined by local structures, which are well reproduced by the diffusion model even at a resolution well-outside the training range.

This trivial solution works well for image-sizes up to 2048\texttimes{}2048, after which the inference requires more than $24GB$ of VRAM.
In general, the memory requirement of this approach scales poorly with size, making it difficult to apply to larger textures.
Taking inspiration from MultiDiffusion~\cite{omer2023multidiffusion}, we develop a window-based infinite texture generation process that uses constant VRAM.
The idea consists of simultaneously denoising multiple overlapping patches from the larger input $\vect{z}_t$.
The predicted noise for the overlapping regions is averaged across the multiple function calls, effectively harmonizing the different denoising trajectories.
\citet{wang2024infinite} used a similar method for generating an infinite texture of a given exemplar.
A limitation of this approach is that pixels across distant patches are denoised independently, and nothing pushes the texture to be globally consistent.

We note that the high-level directions of variation in a texture class (such as color change or pattern density) are encoded in the low-frequency (LF) components of the initial noise image (also observed by \citet{chung2024style} and \citet{everaert2024exploiting}).
We highlight this relationship in Fig.~\ref{fig:low_freq} by showing the output of the diffusion model for independently-sampled noise latents and by comparing them to different manipulations of the LF components of the latent maps.
By copying the mean of the noise from a prototype texture (first column), it can be observed that the style of the 4 outputs become more consistent.
Copying further LF components (by subtracting a blurred version of itself and adding the blurred prototype) yields increasingly consistent patterns.
More precisely, for rows 3 and 4 we use a Lanczos low-pass filter with a cutoff frequency $f_c$ of $0.1$ and $0.2$ respectively.
Using a large cutoff for the filter results in very consistent textures; however, the LF components of the generated images also become similar, creating repetitive patterns. This becomes obvious when generating large textures (see Fig.~\ref{fig:full_examples}).

To alleviate this issue, we make the observation that obtaining consistent textures does not require identical low frequencies, but only a similar spectral distribution.
This
follows
the intuition that diffusion models perform approximate spectral autoregression~\cite{dieleman2024spectral}.
Therefore, our noise-uniformization method consists of replacing the LF components of the noise with a random permutation of the LF of a prototype which controls the style.
To be exact, the noise tensor $\vect{z}_N$ that follows the style of a different noise prototype $\vect{p}_N$ is computed as
$\vect{z}_N = \vect{w} - \texttt{blur}(\vect{w}) + \texttt{upscale}(\texttt{shuffle}(\texttt{downscale}(\texttt{blur}(\vect{p}_N)))$, where $\vect{w}$ is pure white noise, and \texttt{shuffle} a random permutation of
pixels (cf.\ Fig.~\ref{fig:full_examples}).

To improve the efficiency of the sliding-window denoising, we set upper and lower bounds on the number of overlapping pixels in adjacent patches.
The patches used by the diffusion model randomly shift within these bounds for each denoising step, which avoids the formation of seams at the edges of the patches.
By allowing the shifted windows to wrap around the texture
we obtain an additional quality: the denoised texture can be seamlessly tiled.
Please see Fig.~\ref{fig:tileable} in the supplementary material for tiled texture samples.
We note that our randomly-shifted sliding windows bear similarities with the strategies employed by \citet{wang2024infinite} and \citet{vecchio2024controlmat} (noise rolling). We investigate the synergy between the latter and our noise uniformization in the supplementary (Fig.~\ref{fig:control_uniform}).

\section{Experiments}
\label{sec:experiments}

We firstly evaluate our pipeline on the 5 texture classes from the MVTec dataset \cite{bergmann2021mvtec}, designed for anomaly detection. There are approximately 100 images for each texture (around 20 images per anomaly type).
Additionally, we stress-test our method with smaller image sets; using a handheld phone camera, we collect 9 different textures ranging from 1 to 20 images. Notably, 3 of the textures are single-image.
We include the exact counts and representative samples for each texture in the supplementary (~\ref{asec:data_examples}).

In Fig.~\ref{fig:segmentation}, we compare different anomaly-segmentation approaches.
The labels obtained using our method satisfy the most important criteria for conditional synthesis: semantically similar prominent features are grouped into the same class; the detected regions are compact, with relatively little (spatial) noise; and the background/normal pixels are mapped to a unique class.
On the other hand, the image-level anomaly clustering method BlindLCA produces noisy labels when used for pixel-level segmentation.
Generic unsupervised semantic segmentation methods (such as STEGO~\cite{hamilton2022unsupervised}) fail to disentangle the anomalies due to their rarity.

Fig.~\ref{fig:ml_synthesis} includes images generated by our method.
Please note that the label maps support painting multiple features with various shapes and sizes.
In Fig.~\ref{fig:ml_synthesis_mvtec} \TAedit{(supplementary material)} we show that this holds even for the MVTecAD dataset, where, except for wood, all textures seen during training have a single anomaly-type per image.
That is, the generative model is able to gracefully generalize from single features to multi-feature painting on textures.

We provide a qualitative comparison between our approach and various methods that are adapted for our task in Fig.~\ref{fig:compare_synthesis_transposed}. 
Since our pipeline is the first to offer a complete workflow for painting rare features from an unlabeled collection of textures,
that evaluation includes
methods that are able to paint features, guided by given input masks.
We do not compare to generic prompt-based image editing~\cite{lai2025anomalypainter, sun2024cut, brooks2022instructpix2pix}, as we are interested in painting features as extracted from specific exemplars.
For the baselines designed to use a single image, we select one texture from our training set that contains a feature with a relatively similar shape to the target (see first column) and use the ground-truth mask of the prominent feature.
The significant previous work of \citet{hu2024diffusion} was not included here because the approach proved unsuitable for painting small and subtle features;
a representative failure case is shown in Fig.~\ref{fig:painting_fail} of the supplementary.

\subsection{Editing results}

\begin{figure}
  \centering%
  \setlength{\cw}{0.2333\linewidth}%
  \setlength{\tabcolsep}{+0.003\linewidth}%
  \begin{tabular}{cccc}
    \figvlabel{Native class}%
    \includegraphics[width=\cw]{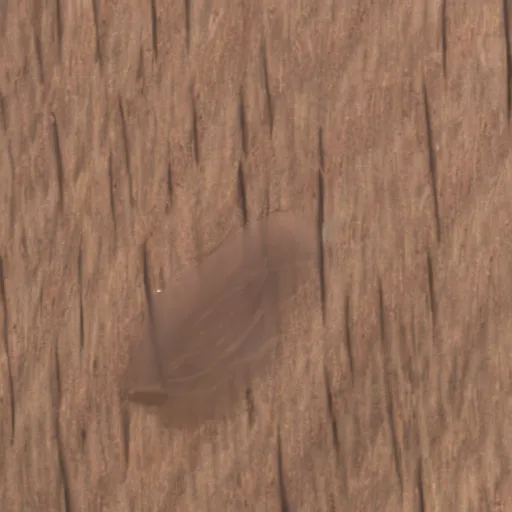} &
    \includegraphics[width=\cw]{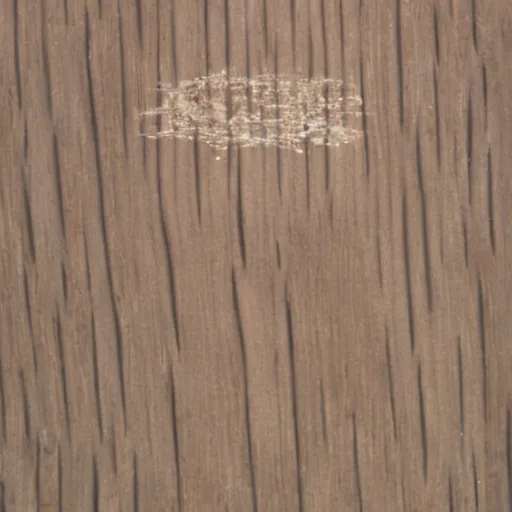} &
    \includegraphics[width=\cw]{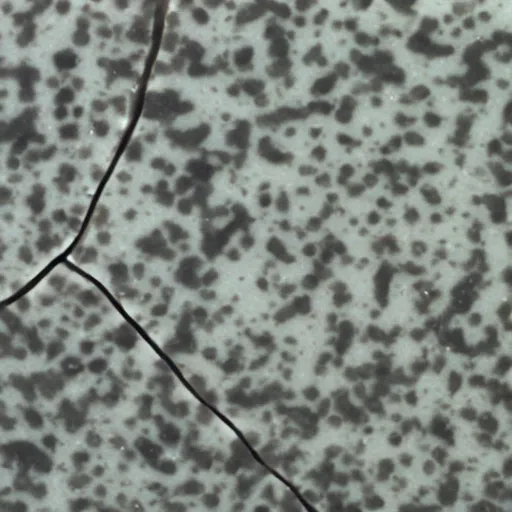} &
    \includegraphics[width=\cw]{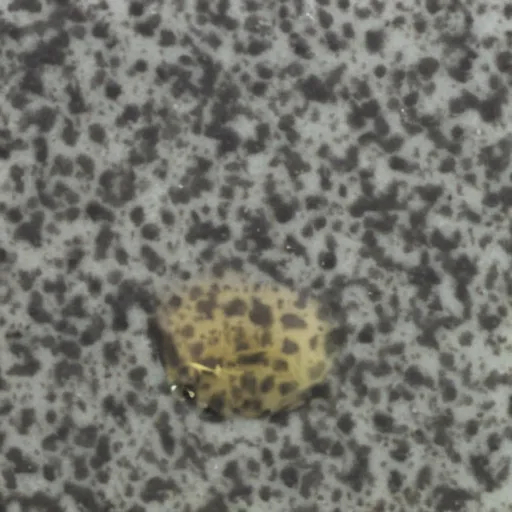} \\
    
    \figvlabelX{Non-native classes}
    \includegraphics[width=\cw]{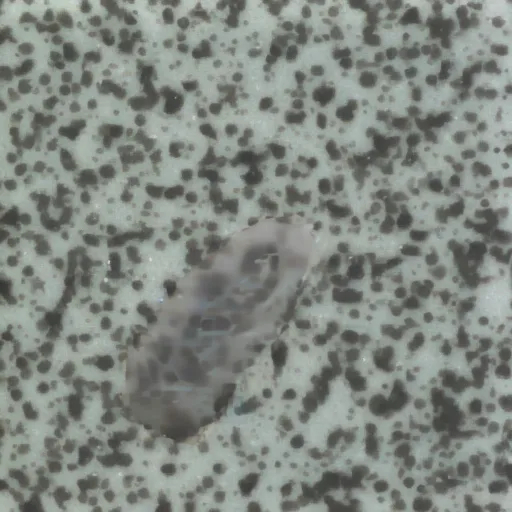} &
    \includegraphics[width=\cw]{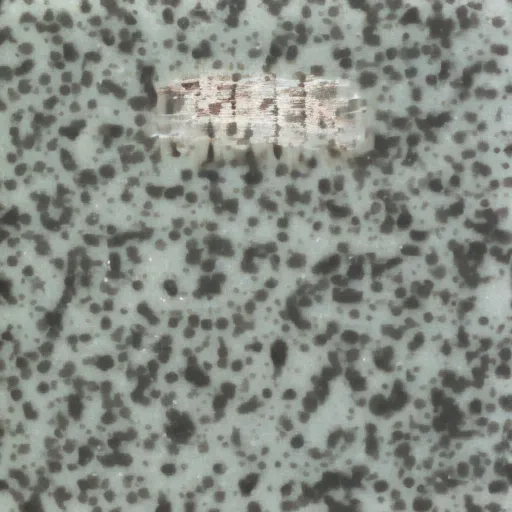} &
    \includegraphics[width=\cw]{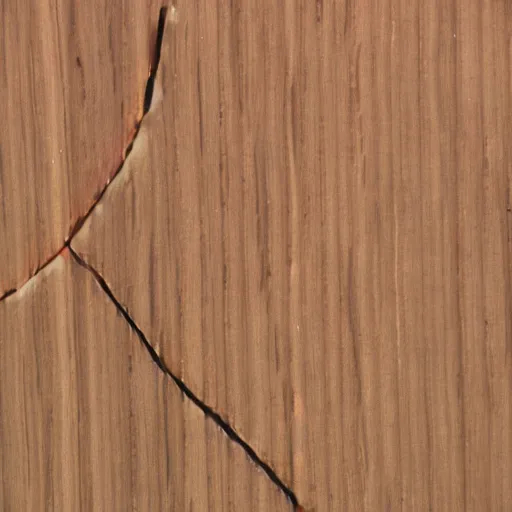} &
    \includegraphics[width=\cw]{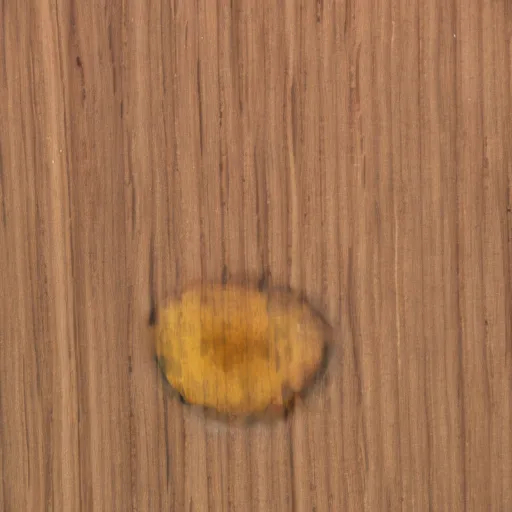} \\[-2.5pt]

    \figvlabel{\phantom{Class}}
    \includegraphics[width=\cw]{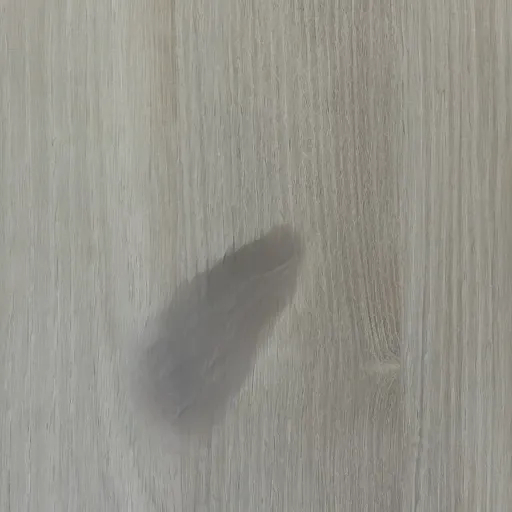} &
    \includegraphics[width=\cw]{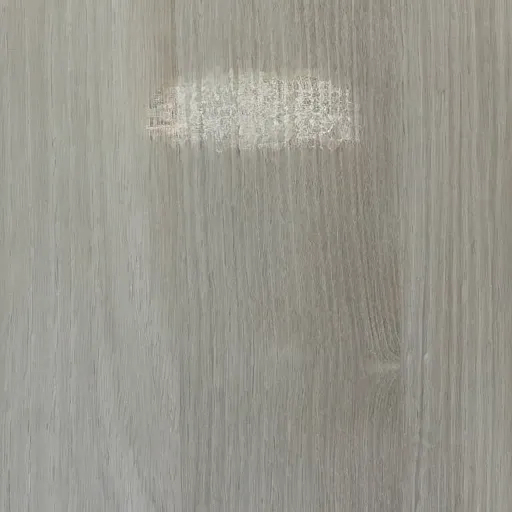} &
    \includegraphics[width=\cw]{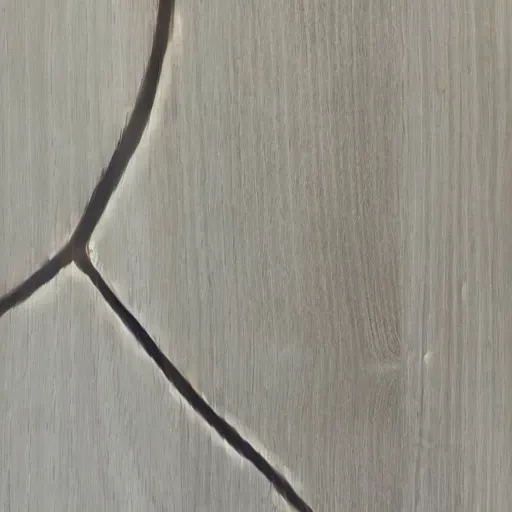} &
    \includegraphics[width=\cw]{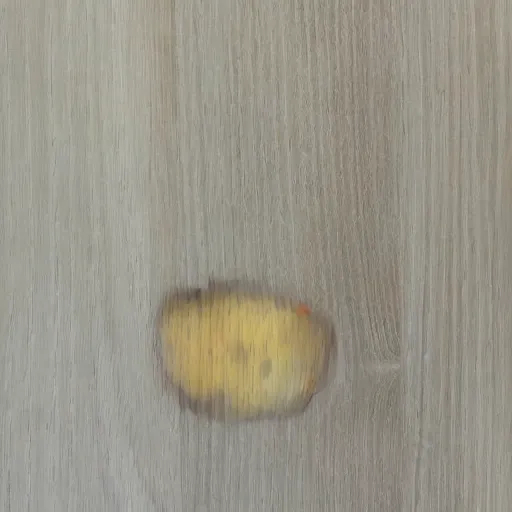} \\
  \end{tabular}
  \caption{\label{fig:edit_transfer}%
    Feature transfer to real images of texture classes non-native to each feature. Due to the novel surrounding, cues on scale are missing, so feature scale may vary across transfers, particularly visible with the crack feature.%
  }
  \Description{The figure contains 4 images showing how features learned from wood textures can be applied to tiles, and the other way around as well.}
\end{figure}

To demonstrate the editing capabilities of our approach, we use anomaly-free images that were not seen during training \-- neither for the contrastive learning, nor for the diffusion model.
We illustrate firstly the intermediate steps in our texture-editing method in Fig.~\ref{fig:edit_simple}.
The resulting image is very close to the original texture outside of the edited region; as shown in the supplementary (Fig.~\ref{fig:inversion_sup}), virtually all existing differences are caused by the limited capacity of the pretrained variational autoencoder from Stable Diffusion~\cite{rombach2022high}.
Notably, thanks to our noise-mixing method, the edited region is integrated seamlessly in the texture, matching the previous appearance underneath the roughened region and at the boundaries.
We include more editing results in Fig.~\ref{fig:edit_results}.

Thanks to the seamless integration of the edited region, our noise-mixing lends itself to high-resolution editing at interactive rates (see video in the supplementary material). This is achieved by running the diffusion only for the texture patch that contains the feature label to be painted. To this end, we save the diffusion trajectory during generation or inversion and only use the slice of the noise estimates that corresponds to the edited patch.
Saving the noise history for the diffusion process makes it possible to forego inversion for each edit, thus reducing the time to about 1.5 seconds per edit.
More details and high-resolution examples are included in the supplementary,~\ref{asec:high_res}.

\subsection{Synthesizing materials}
\begin{figure}
  \newlength{\cc}%
  \setlength{\cc}{0.248\linewidth}
  \begin{minipage}{0.485\linewidth}
    \setlength{\cw}{0.245\linewidth}%
    \setlength{\parindent}{0pt}\centering%
    \fighlabel{Input}

    \includegraphics[width=\cw]{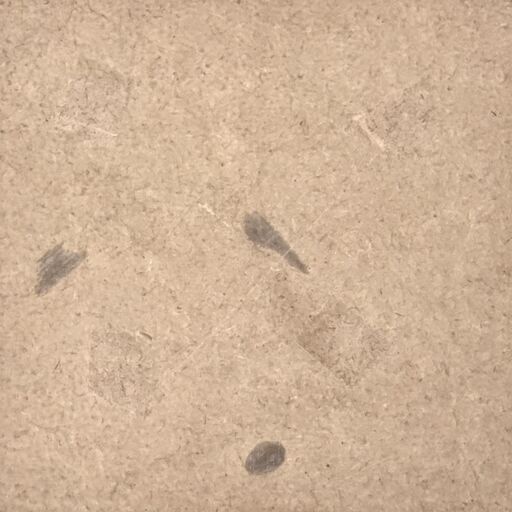}%
    \strut\hfill\strut%
    \includegraphics[width=\cw]{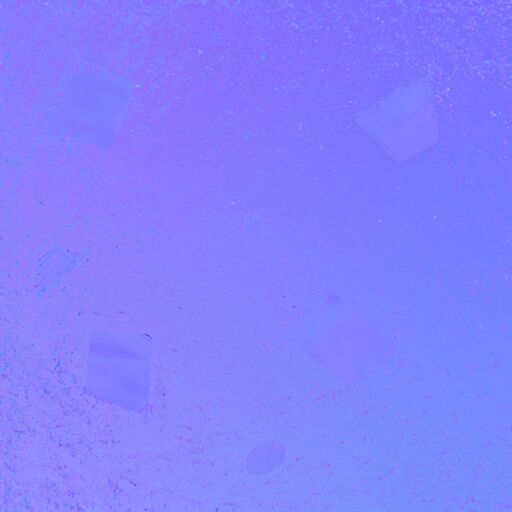}%
    \strut\hfill\strut%
    \includegraphics[width=\cw]{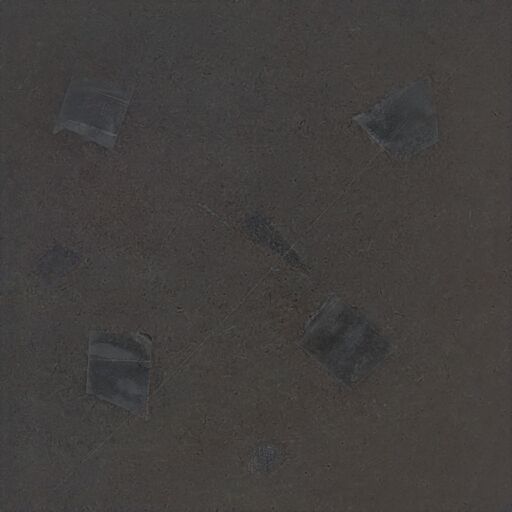}%
    \strut\hfill\strut%
    \includegraphics[width=\cw]{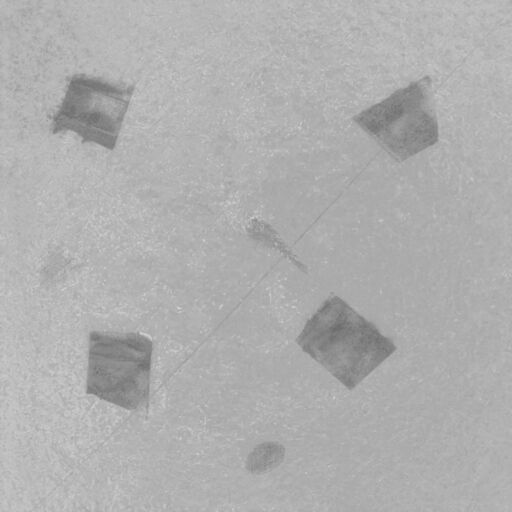}%
  \end{minipage}
  \strut\hfill\strut%
  \begin{minipage}{0.485\linewidth}
    \setlength{\cw}{0.245\linewidth}%
    \setlength{\parindent}{0pt}\centering%
    \fighlabel{Output}
    
    \includegraphics[width=\cw]{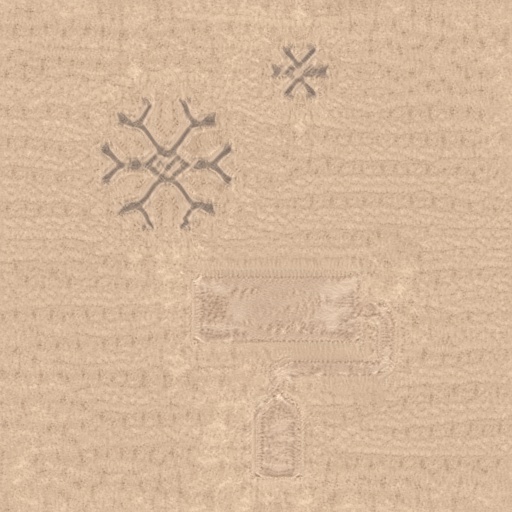}%
    \strut\hfill\strut%
    \includegraphics[width=\cw]{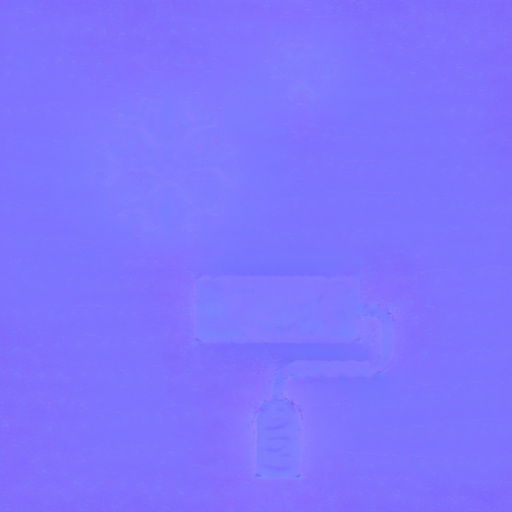}%
    \strut\hfill\strut%
    \includegraphics[width=\cw]{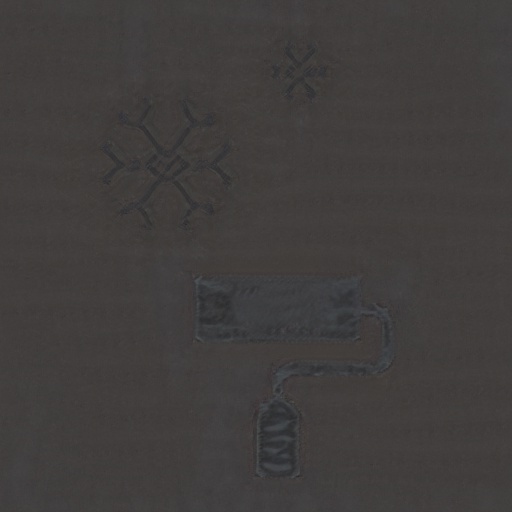}%
    \strut\hfill\strut%
    \includegraphics[width=\cw]{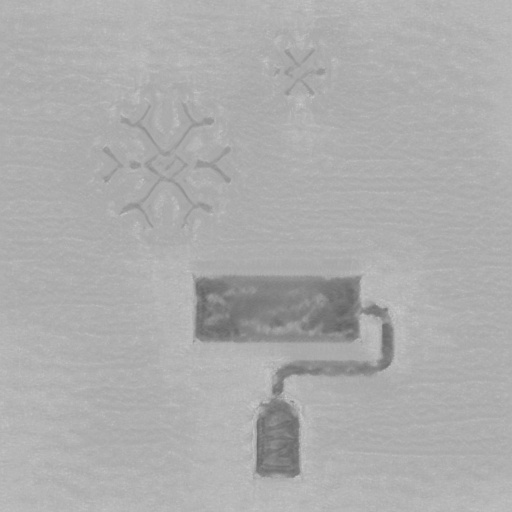}
  \end{minipage}
  
  \includegraphics[width=\cc]{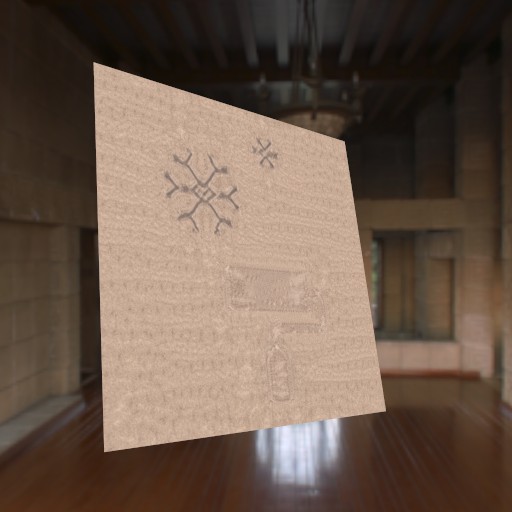}%
  \strut\hfill\strut%
  \includegraphics[width=\cc]{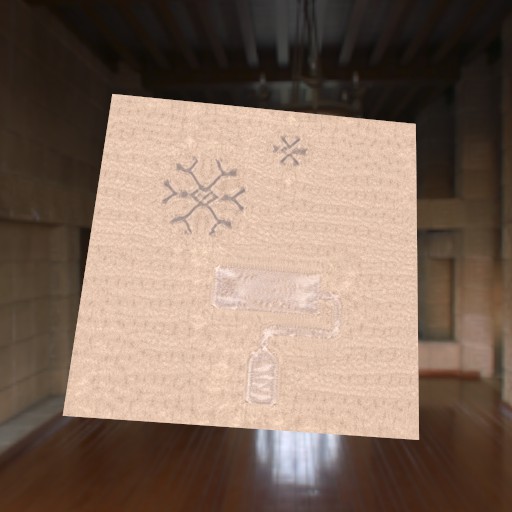}%
  \strut\hfill\strut%
  \includegraphics[width=\cc]{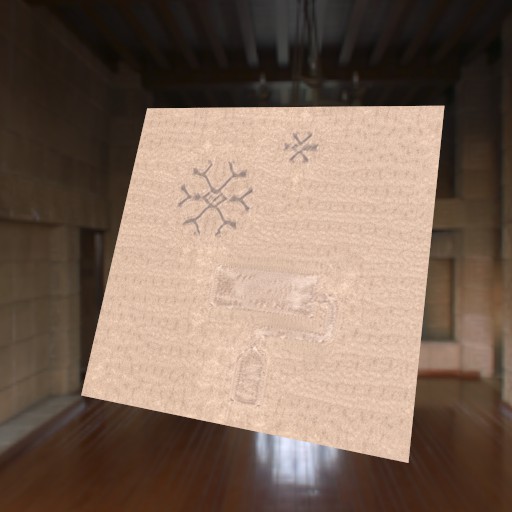}%
  \strut\hfill\strut%
  \includegraphics[width=\cc]{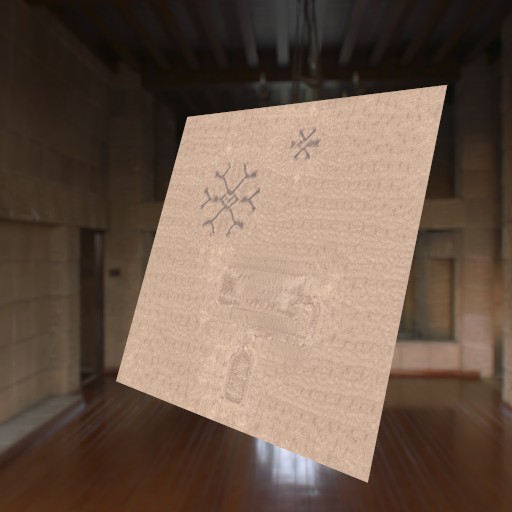}\\[-1ex]
  \caption{\label{fig:svbrdf}%
    Synthesizing new material maps using our pipeline from a single SVBRDF. 
    Please see the supplementary GIF for a high-resolution rendering.}
  \Description{The figure shows the input and generated material maps.}
\end{figure}

So far we have described our approach and demonstrated its capabilities in terms of plain RGB textures.
All our components, however, easily extend to other modalities, such as material maps for synthesizing SVBRDFs.
We include in Fig.~\ref{fig:svbrdf} a simple example, where the input data consists of a single SVBRDF, captured with a smartphone using MaterialGAN~\cite{guo2020MaterialGAN}.

\section{Ablations}

\begin{figure}
  \centering%
  \setlength{\cw}{0.196\linewidth}%
  \setlength{\tabcolsep}{0.001\linewidth}%
  \renewcommand{\arraystretch}{0.3}%
  \begin{tabular}{ccccc}
    \includegraphics[width=\cw]{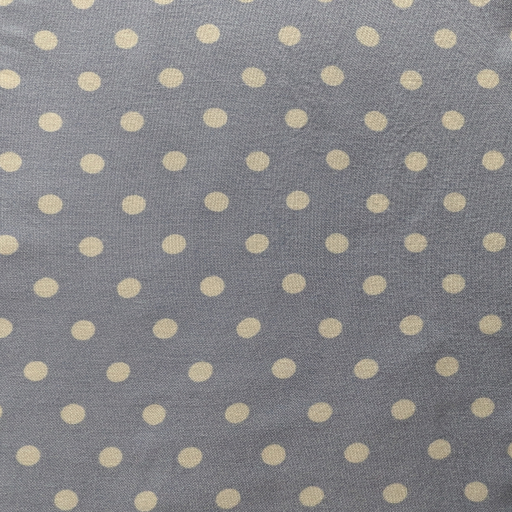} &

    \includegraphics[width=\cw]{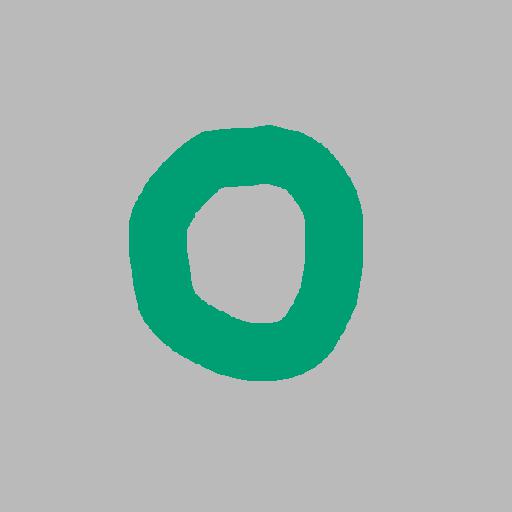} &

    \includegraphics[width=\cw]{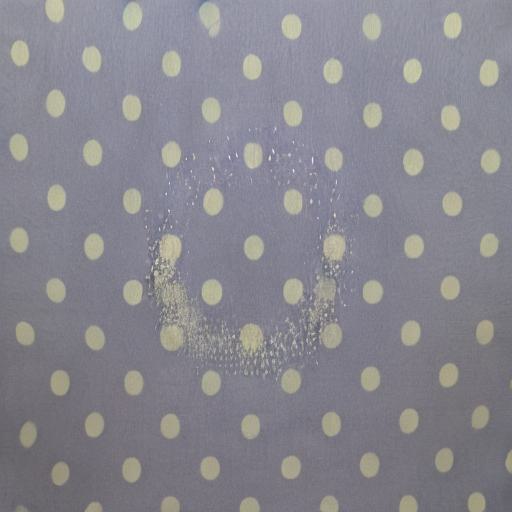} &

    \includegraphics[width=\cw]{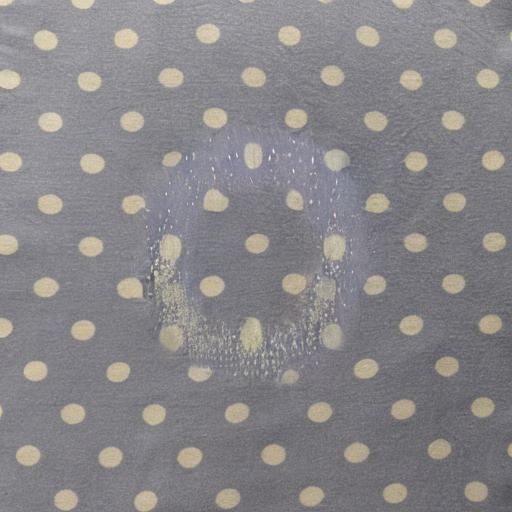} &

    \includegraphics[width=\cw]{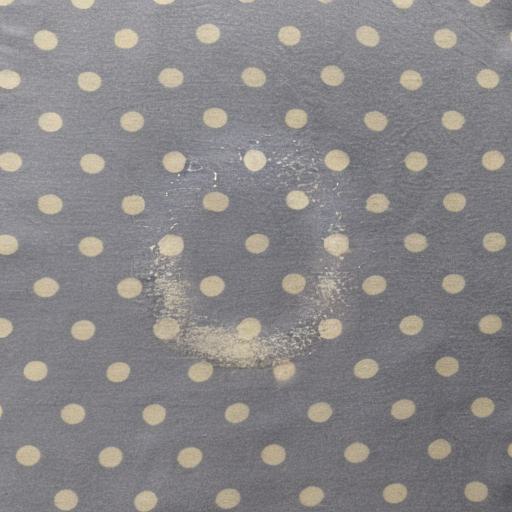} \\

    \includegraphics[trim=320 218 290 580, clip, width=\cw]{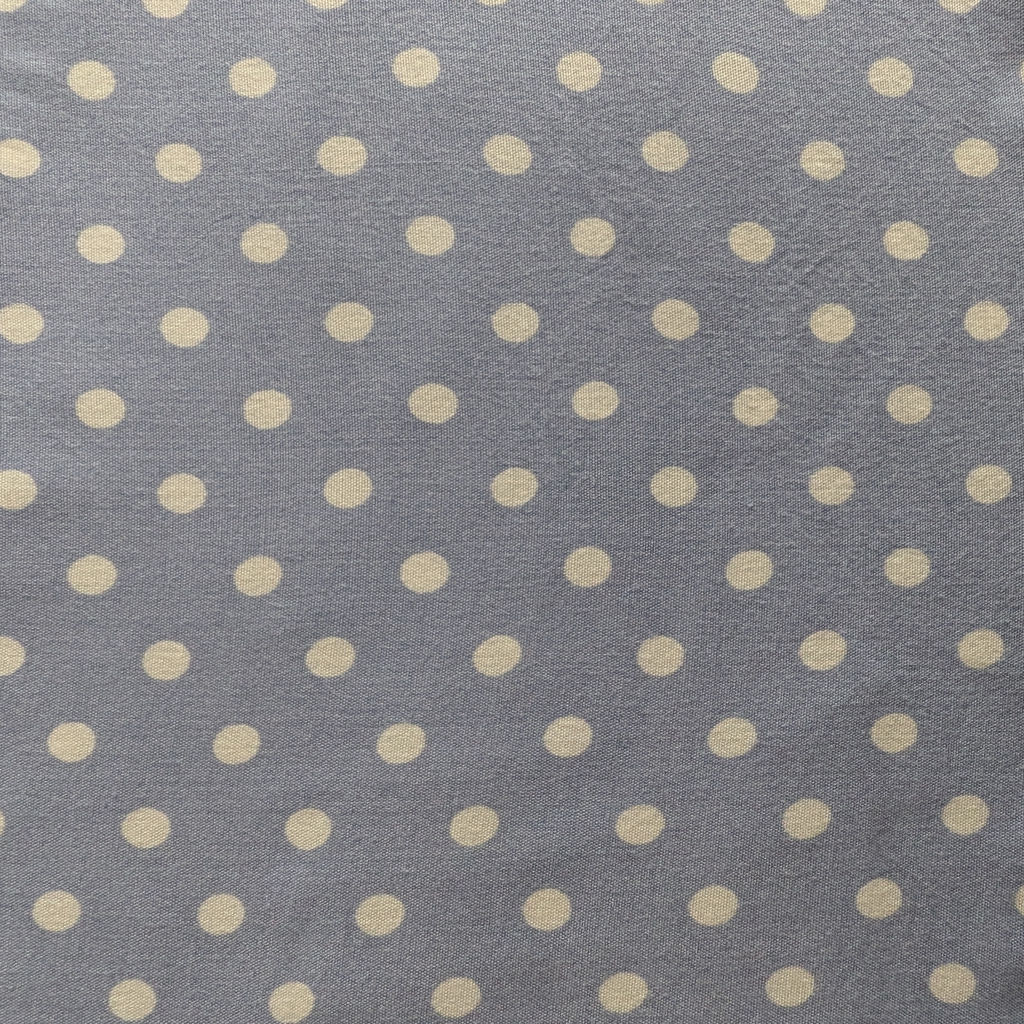} &
    \includegraphics[trim=160 109 145 290, clip, width=\cw]{images/ablation_noisemix/edit_mask.jpg} &
    \includegraphics[trim=160 109 145 290, clip, width=\cw]{images/ablation_noisemix/inv_c.jpg} &
    \includegraphics[trim=160 109 145 290, clip, width=\cw]{images/ablation_noisemix/inv_force.jpg} &
    \includegraphics[trim=160 109 145 290, clip, width=\cw]{images/ablation_noisemix/inv_ours.jpg} \\[4pt]

    \fighlabel{Original} & \fighlabel{Mask} & \fighlabel{Inv} + $\vect{c}$ & \fighlabel{Forced + $\vect{c}$} & \fighlabel{Ours}
    
  \end{tabular}
  \caption{\label{fig:ablation_nm}%
    Ablation of noise-mixing. Note that our edit better preserves the structure of the original texture, while maintaining the realism of the feature.}
  \Description{The images show that our noise-mixing method better preserves the original structure of the texture.}
\end{figure}

\begin{table}%
  \caption{\label{tab:ablate_anomaly}%
    Ablation of our feature detection and clustering components. Metrics are computed under optimal matching to the ground-truth classes.}
  \centering\small\tablefontfamily%
  \setlength{\tabcolsep}{7.5pt}%
  \begin{tabular}{lccc}
    \toprule
    & Accuracy $\uparrow$ & IoU $\uparrow$ & F1-score $\uparrow$ \\ 
    \midrule
    Ours w/o equation~\eqref{eq:threshold} & 0.963 & 0.320 & 0.407 \\
    Ours w/o stratified negatives & 0.794 & 0.367 & 0.451 \\
    Ours & \textbf{0.978} & \textbf{0.473} & \textbf{0.566} \\
    \bottomrule
  \end{tabular}
\end{table}

We ablate our thresholding method, described in Eq.~\eqref{eq:threshold}, and our stratified selection of negative pairs for contrastive learning.
The results of this experiment are summarized in Table~\ref{tab:ablate_anomaly} and show that the introduced mechanisms are vital for an accurate segmentation.

Our noise-mixing technique is ablated in Fig.~\ref{fig:ablation_nm}, by comparing our method with two variants.
In the first case, we simply use the inverted noise as input for the diffusion model with the new conditioning map $\vect{c}$. 
In the second, we also constrain the latents outside the edit to perfectly match the original image.
Both variants unfavorably alter the texture beyond the desired edit.
Additionally, we include in the suppl. material (\ref{asec:diffusion}) a discussion of our choice of diffusion model used as backbone for the feature painting.

\section{Limitations}
Naturally, the realism of the generated results depends on the plausibility of the input mask. Fig.~\ref{fig:many_icons} in the supplementary shows an exhaustive combination of input masks and generated texture features, demonstrating the versatility, but also limitations of the method. Overall, we believe that it shows that our method still extrapolates commendably beyond the mask shapes observed during training.
A different limitation is that our pipeline is not end-to-end trainable, meaning that imperfect segmentations from our automatic feature clustering can cause different features to be mapped to the same label. 
In practice, this causes the model to \emph{choose} what feature is being painted based on the shape of the conditioning label.
Finally, while our method has fast inference times, training the diffusion model for a new texture class can take 6 to 12 hours.

\section{Conclusion}

Our framework learns and disentangles the stationary characteristics from the prominent features given only a small collection of textures.
After training, we are able to generate tileable textures of arbitrary size and to paint features on similar and dissimilar images with realistic transitions.
These functionalities are bundled in a single model that can be controlled interactively for iterative authoring.
While the main focus of our work is asset creation, the utility of the method extends to various areas, such as, image augmentations, long-tail image generation, and generating fake anomalies for self-supervised anomaly detection.
Moreover, our editing method and noise uniformization algorithm can be leveraged in more general contexts, as shown in~\ref{asec:uniform} (supplementary material).

\begin{acks}
This project has received funding from the European Union’s Horizon 2020 research and innovation programme under the Marie Skłodowska-Curie grant agreement No 956585.
\end{acks}

\newpage
\bibliographystyle{ACM-Reference-Format}
\bibliography{main}

\clearpage%
\appendix
\setcounter{page}{1}
\makeatletter
\renewcommand{\thesection}{S\arabic{section}}
\makeatother

\twocolumn[\sffamily%
  \leftline{\Huge Example-based feature painting on textures}
  \bigskip
  
  \leftline{\LARGE\uppercase{Supplementary Material}}
  \bigskip\medskip
  
  \leftline{\LARGE\uppercase{Andrei-Timotei Ardelean}, \large{Friedrich-Alexander-Universität Erlangen-Nürnberg, Germany}}
  \smallskip
  \leftline{\LARGE\uppercase{Tim Weyrich}, \large{Friedrich-Alexander-Universität Erlangen-Nürnberg, Germany}}
  
  \bigskip\bigskip%
]

\section{Dataset examples}
\label{asec:data_examples}
\begin{table}%
    \caption{Overview of each texture category used in our experiments. It includes the number of anomaly/feature types (K), the number of images without anomalies (Normal images), the number of images that contain prominent features, and the total. The first 5 textures are from MVTecAD, the other 10 are acquired by us.}
    \label{tab:data_details}
    \begin{minipage}{\columnwidth}
    \begin{center}
    \begin{tabular}{lcccc}
        Texture name & K & Normal images & Feature images & Total \\
    \midrule
        Tile & 5 & 33 & 84 & 117 \\
        Grid & 5 & 21 & 57 & 78 \\
        Carpet & 5 & 28 & 89 & 117 \\
        Wood & 5 & 19 & 60 & 79 \\
        Leather & 5 & 32 & 92 & 124 \\
    \midrule
        Pavement & 2 & 0 & 9 & 9 \\
        Taboret & 2 & 0 & 1 & 1 \\
        Shirt & 1 & 0 & 1 & 1 \\
        Puzzle & 1 & 0 & 1 & 1 \\
        Wall & 2 & 0 & 5 & 5 \\
        Grass & 2 & 0 & 15 & 20 \\
        Chair & 1 & 0 & 3 & 3 \\
        Blueberries & 2 & 3 & 12 & 15 \\
        Dots & 3 & 1 & 6 & 7 \\
        SVBRDF & 2 & 0 & 1 & 1 \\
    \bottomrule
    \end{tabular}
    \end{center}
    \end{minipage}
\end{table}

\makeatletter
\begin{figure*}
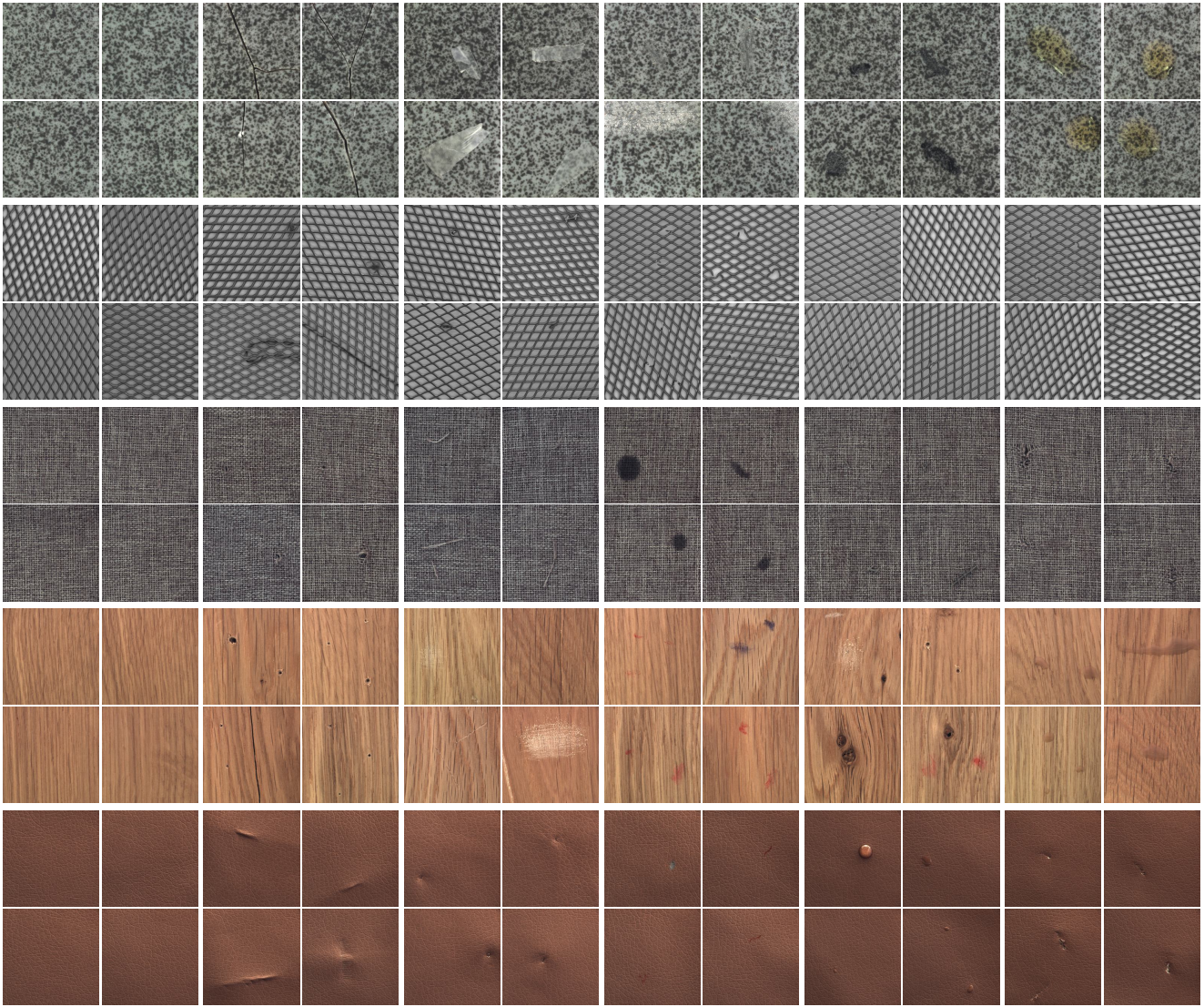

    \centering

    \newcommand{\objects}{{tile},{grid},{carpet},{wood},{leather}}
    \newcommand{\fnum}{6}
    \setlength{\tabcolsep}{+0.002\linewidth}
    \setlength{\cw}{0.078\linewidth}

    \def\tabledata{}%
    \foreach \c [count=\x from 1] in \objects { 
        \ifnum \x>1
            \protected@xappto\tabledata{\\[1.4cm]} 
        \fi
        \foreach \feat in {1,...,\fnum} {
            \ifnum \feat>1
                \protected@xappto\tabledata{&} 
            \fi
            \protected@xappto\tabledata{

                \begin{tabular}{cc}
                    \includegraphics[width=\cw]{images/dataview/\c/\feat_1.jpg} & \includegraphics[width=\cw]{images/dataview/\c/\feat_2.jpg} \\
                    \includegraphics[width=\cw]{images/dataview/\c/\feat_3.jpg} & \includegraphics[width=\cw]{images/dataview/\c/\feat_4.jpg}
                \end{tabular}
            }
        }

    }%

    \setlength{\tabcolsep}{+0.0012\linewidth}
    \renewcommand{\arraystretch}{0.3}
    \begin{tabular}{cccccc}
        \tabledata
    \end{tabular}
    
    \caption{\label{fig:dataview}%
           Overview of the image data in the MVTec AD textures.}
    \Description{
        The Figure includes 4 images for each of the 6 feature type for the 5 textures of MVTec AD.
    }
\end{figure*}
\makeatother

\begin{figure*}
    \centering
    \setlength{\cw}{0.135\linewidth}
    \setlength{\tabcolsep}{+0.002\linewidth}
    \begin{tabular}{ccccccc}
        \includegraphics[width=\cw]{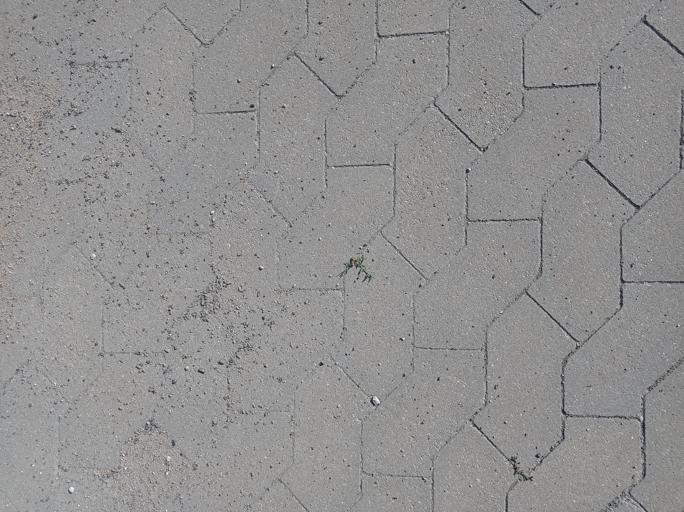} &
        \includegraphics[width=\cw]{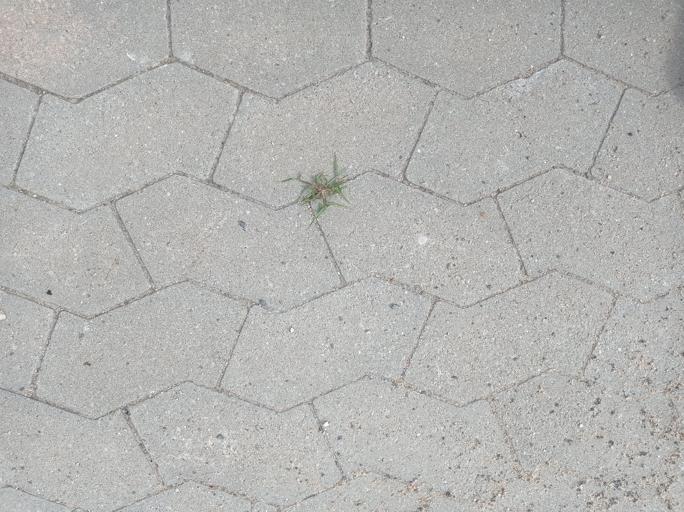} &
        \includegraphics[width=\cw]{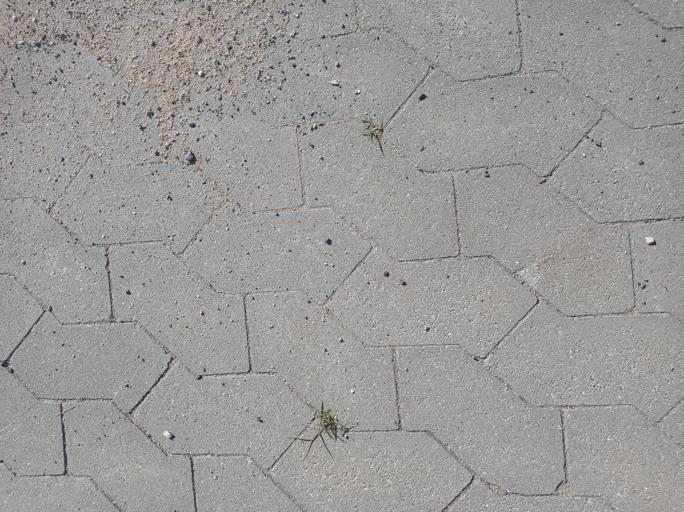} &
        \includegraphics[width=\cw]{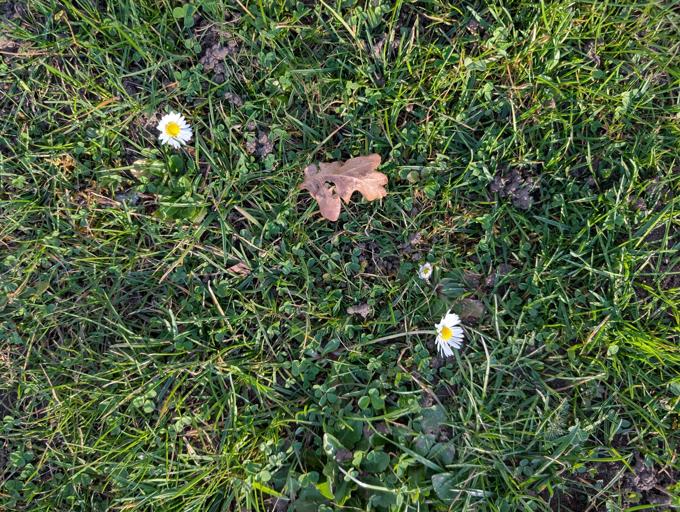} &
        \includegraphics[width=\cw]{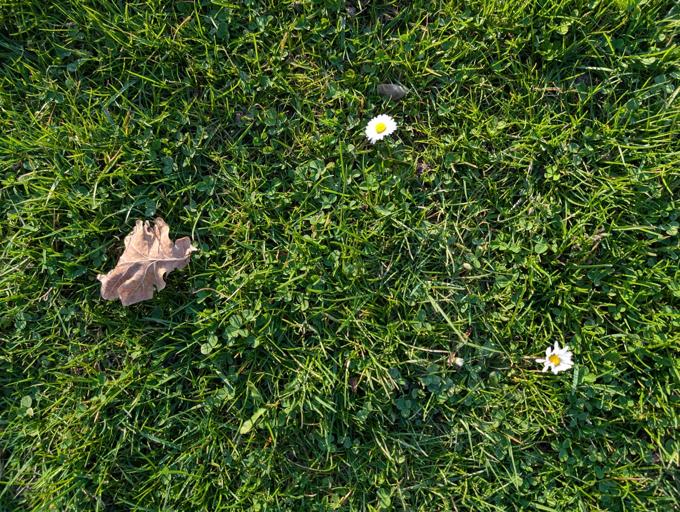} &
        \includegraphics[width=\cw]{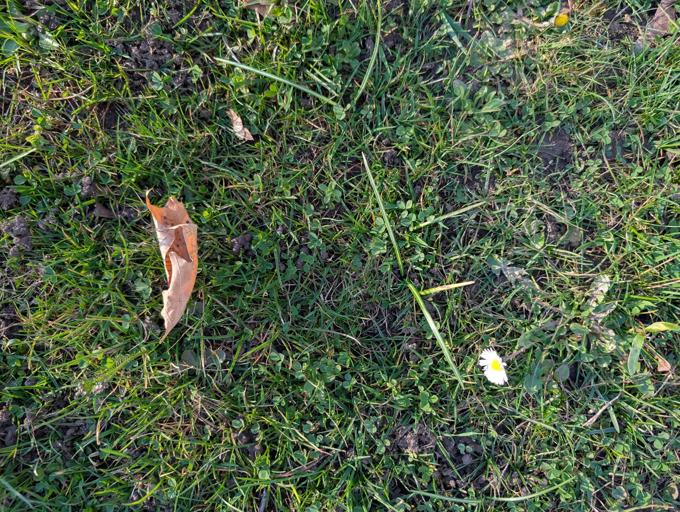} &
        \includegraphics[width=\cw]{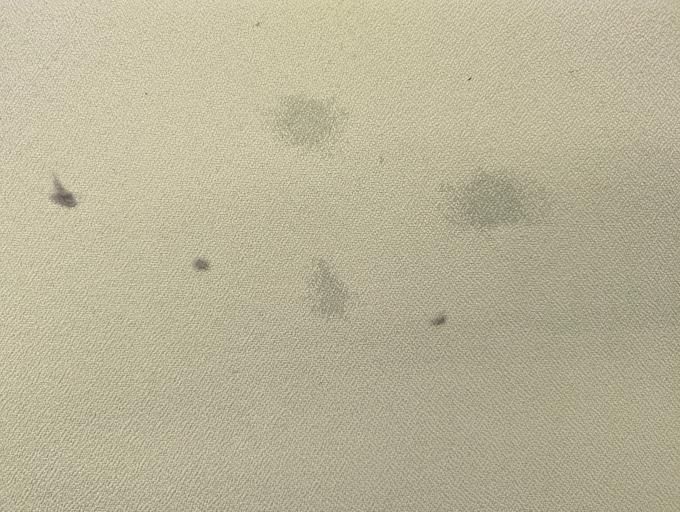}
        \\
        \includegraphics[width=\cw]{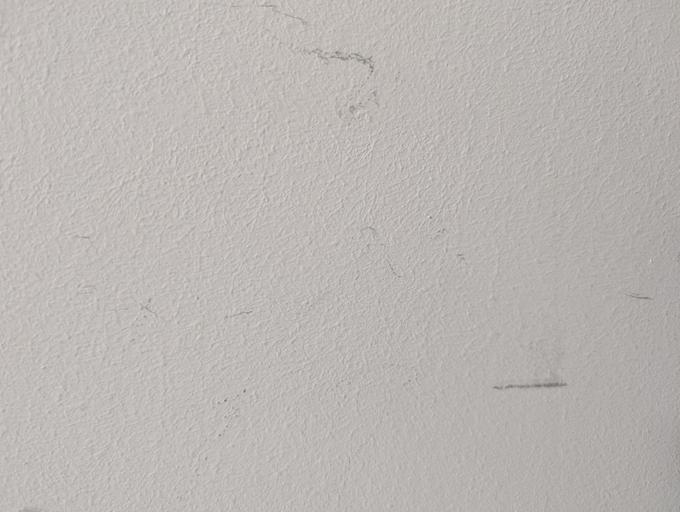} &
        \includegraphics[width=\cw]{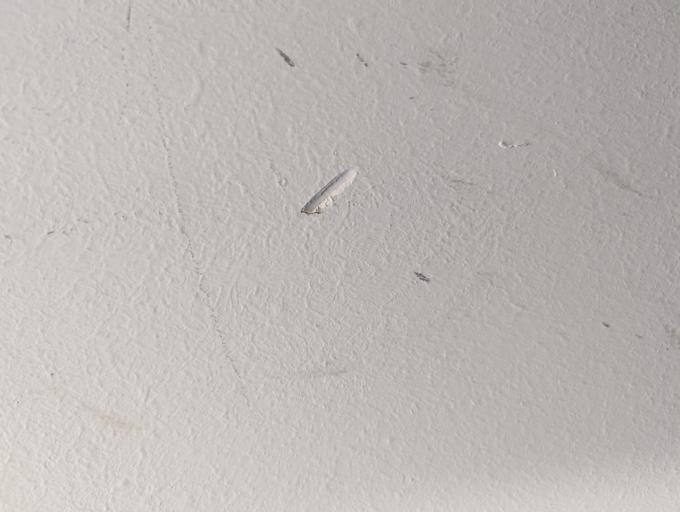} &
        \includegraphics[width=\cw]{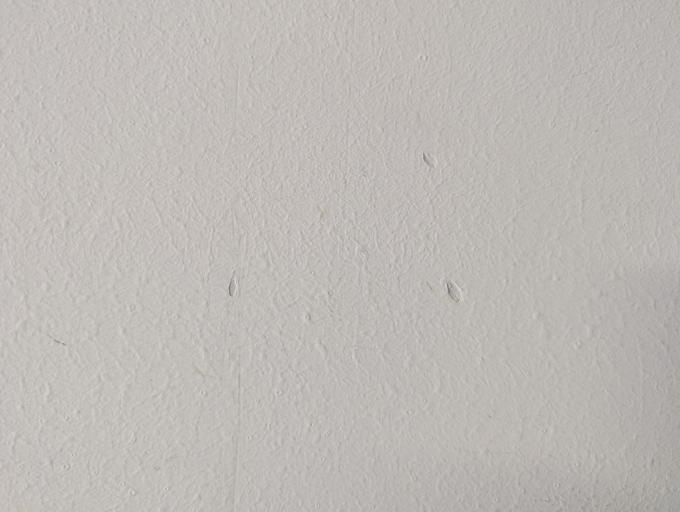} &
        \includegraphics[width=\cw]{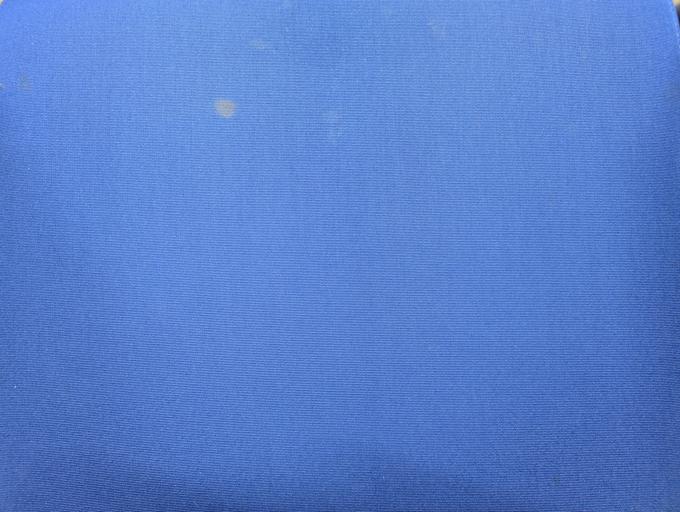} &
        \includegraphics[width=\cw]{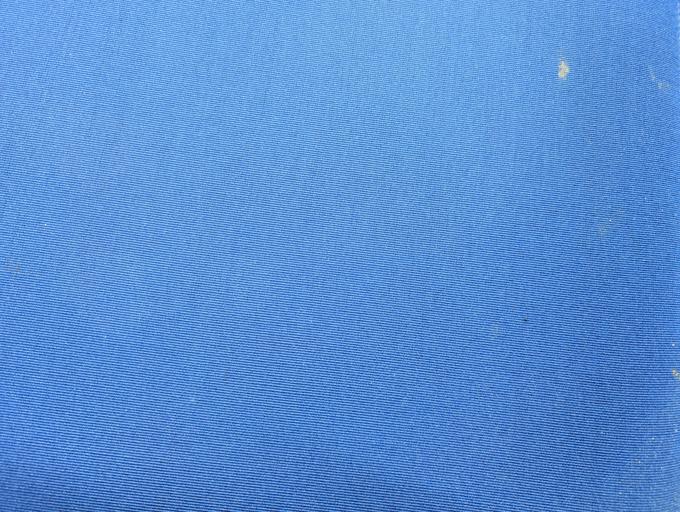} &
        \includegraphics[width=\cw]{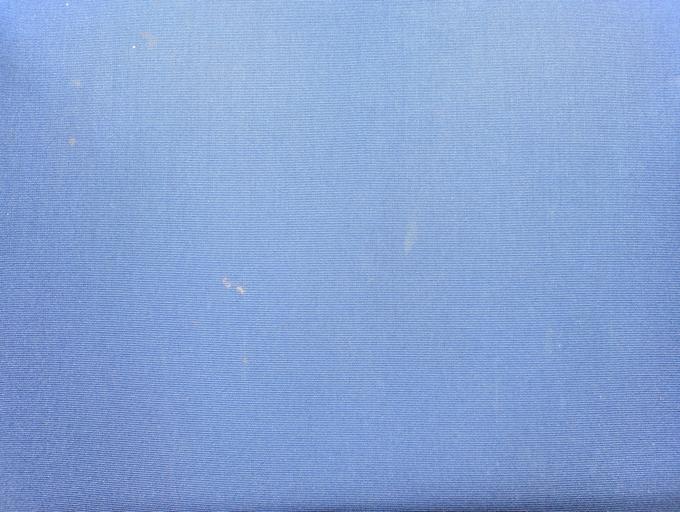} &
        \includegraphics[width=\cw]{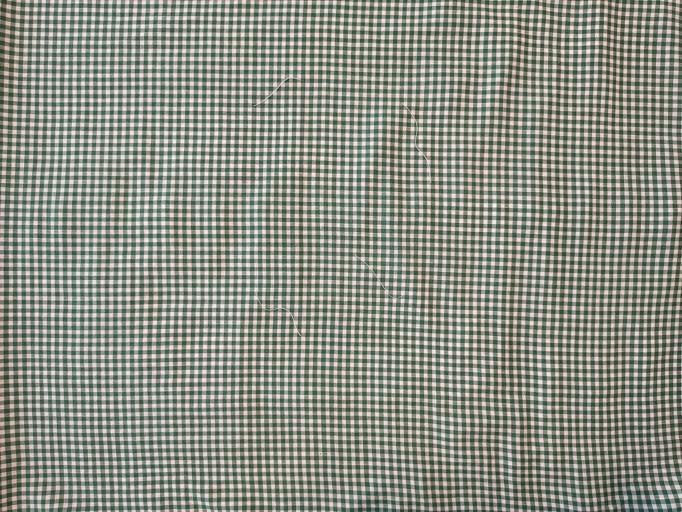} 
        \\
        \includegraphics[width=\cw]{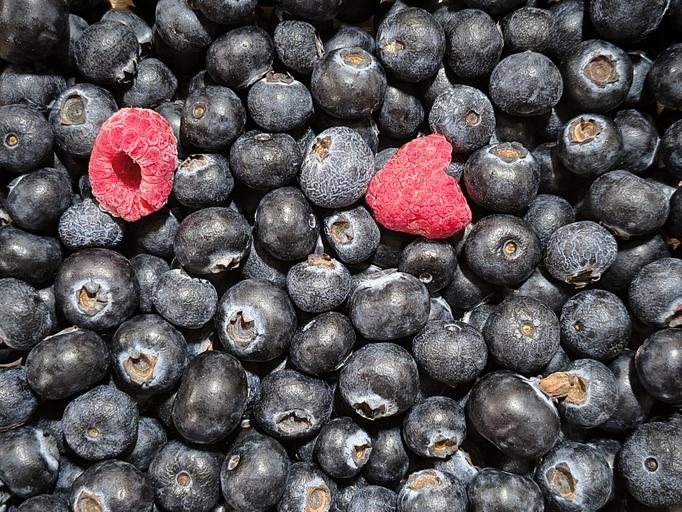} &
        \includegraphics[width=\cw]{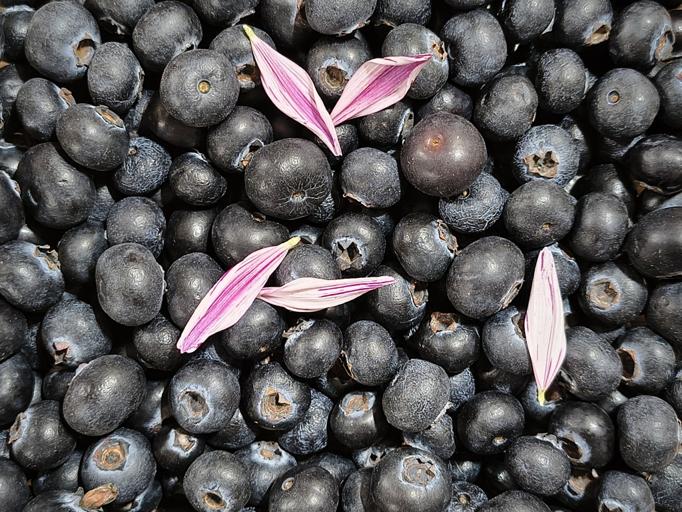} &
        \includegraphics[width=\cw]{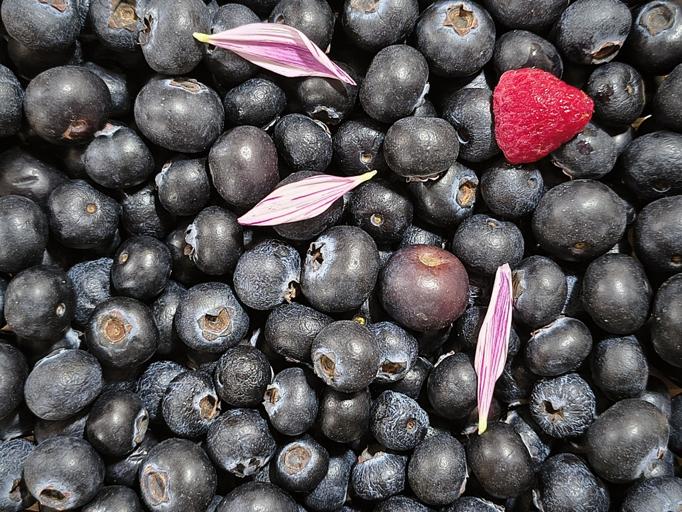} &
        \includegraphics[width=\cw]{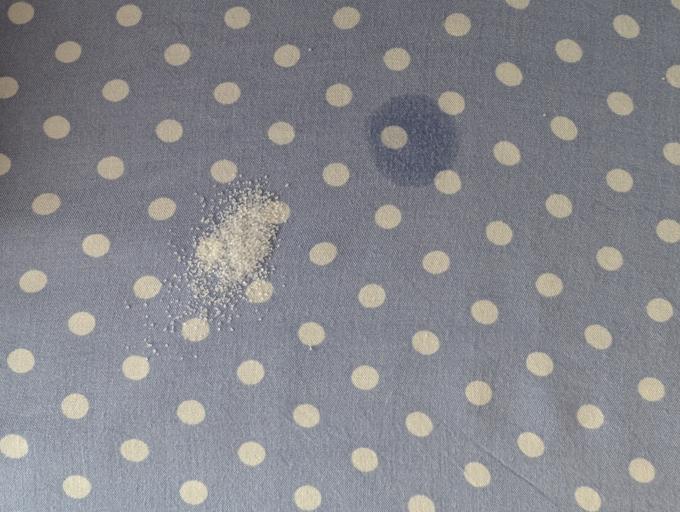} &
        \includegraphics[width=\cw]{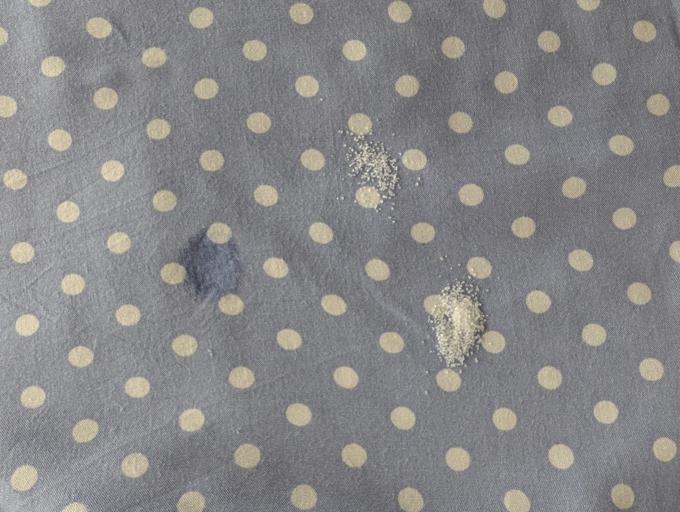} &
        \includegraphics[width=\cw]{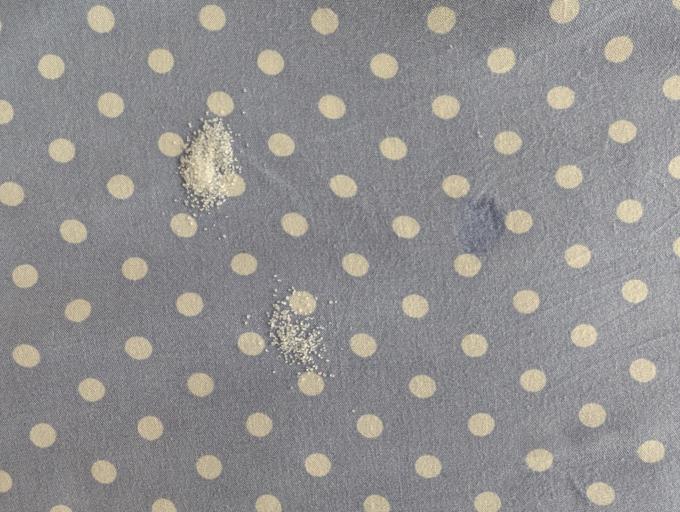} &
        \includegraphics[width=\cw]{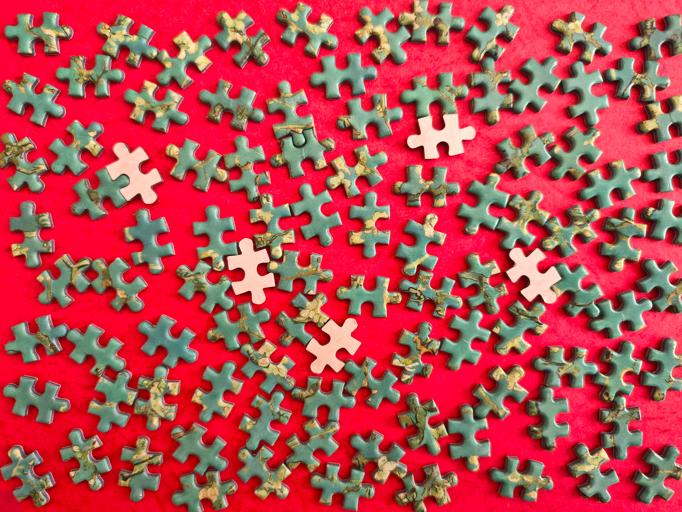}
        \\
    \end{tabular}
    \caption{\label{fig:dataview_our}Overview of our 9 textures, captured using a handheld phone camera. }
    \Description{3 rows with 7 images from our 9 textures.}
\end{figure*}

It is difficult to gauge the quality of the generated textures and the painted features without having a fair understanding of how the training data looks like.
Therefore, we include in Fig.~\ref{fig:dataview} and Fig.~\ref{fig:dataview_our} a set of images from each texture class, presenting all existing features.
Table~\ref{tab:data_details} describes how many images and anomalies, or prominent feature types, are contained in each dataset.

\section{Additional experiment on latent noise uniformization}
\label{asec:uniform}

\begin{figure*}
  \centering
  \setlength{\cw}{0.97\linewidth}
  \setlength{\tabcolsep}{+0.0\linewidth}
  \renewcommand{\arraystretch}{0.3}

  \begin{tabular}{c}        
        
        \includegraphics[width=\cw]{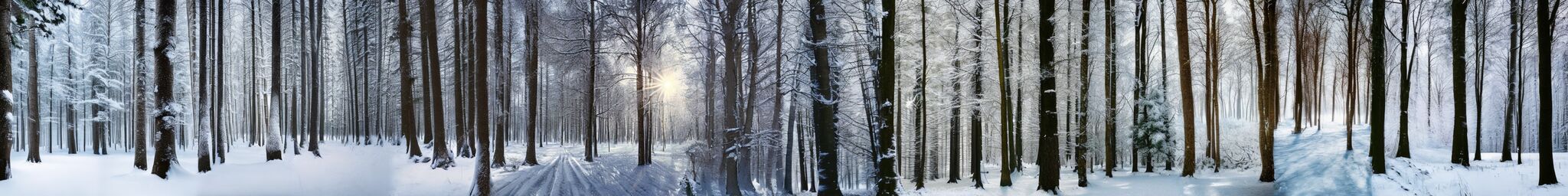} \\

        \includegraphics[width=\cw]{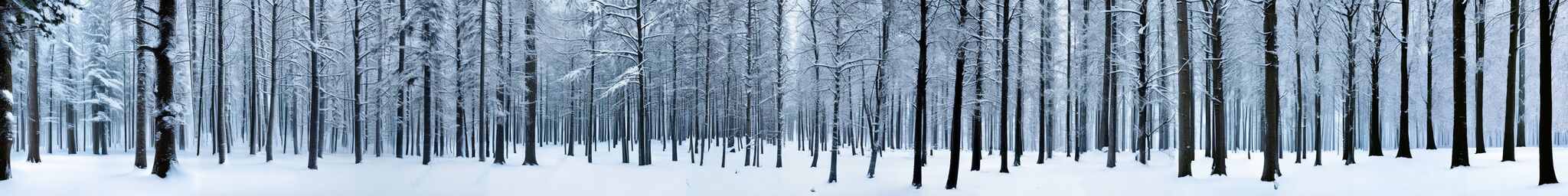} \\

        \includegraphics[width=\cw]{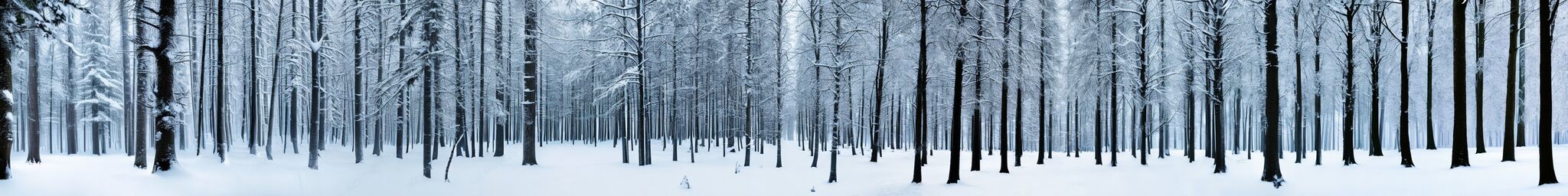} \\[0.3cm]

        \includegraphics[width=\cw]{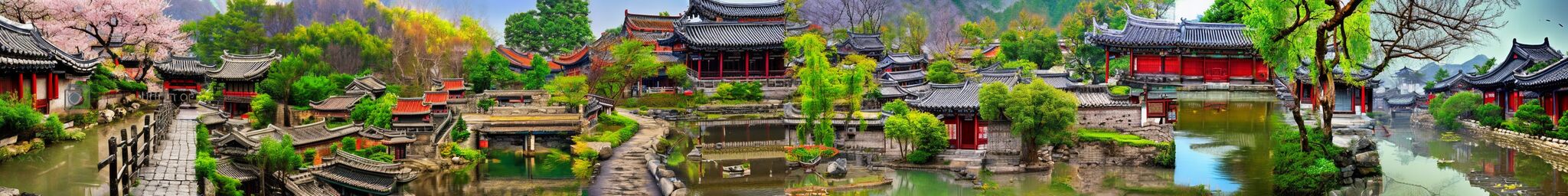} \\

        \includegraphics[width=\cw]{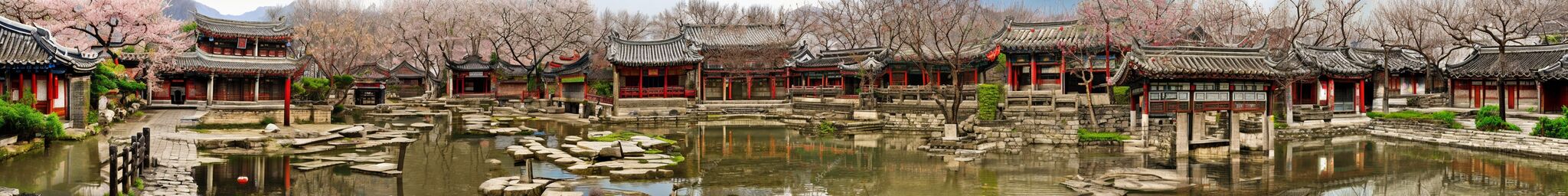} \\
        
        \includegraphics[width=\cw]{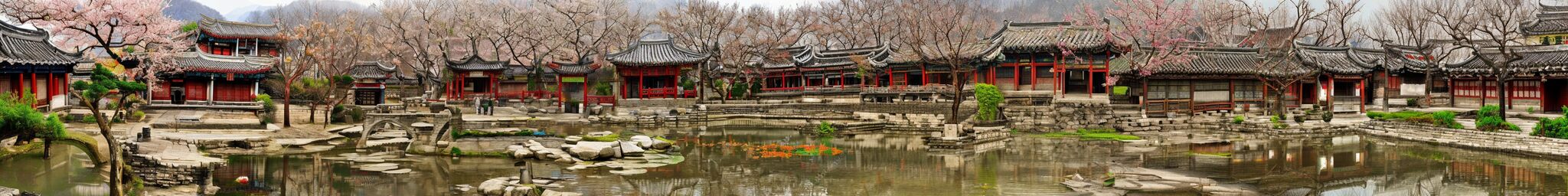}\\[0.3cm]

        \includegraphics[width=\cw]{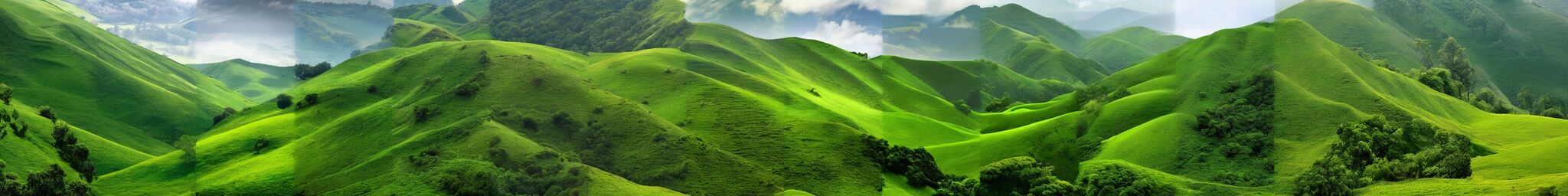} \\

        \includegraphics[width=\cw]{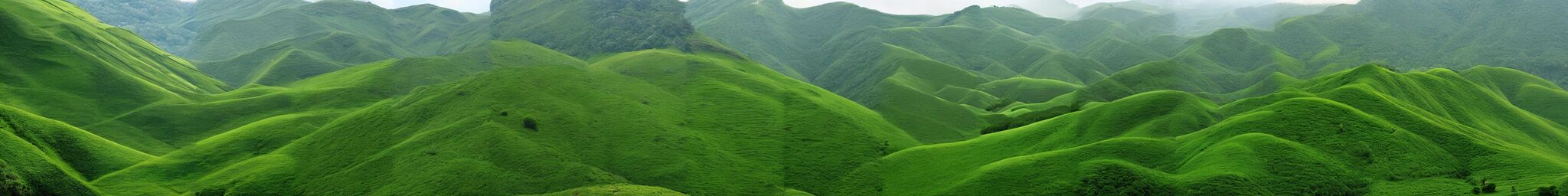} \\

        \includegraphics[width=\cw]{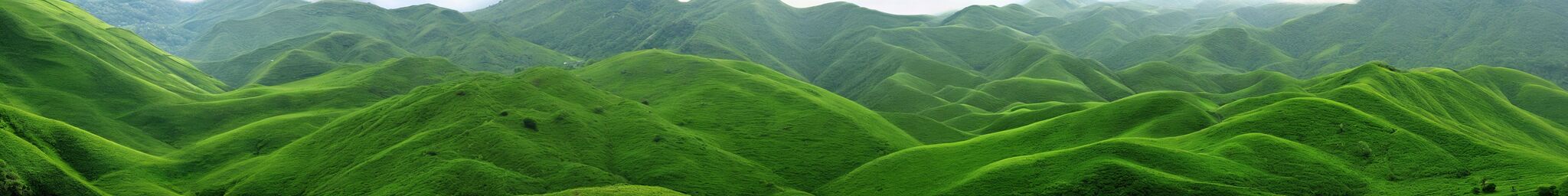} \\
            
    \end{tabular}
    \caption{\label{fig:sd_uniform}%
        Application of our noise uniformization for generic images synthesized using MultiDiffusion \cite{omer2023multidiffusion}. For each prompt the top image is obtained using pure white noise, the second image uses our harmonized noise, the third image additionally uses randomized sliding windows.
    }
    \Description{The figure show 3 groups of 3 images obtained with MultiDiffusion starting from differents types of noise.}
\end{figure*}

\begin{figure*}
  \centering
  \setlength{\cw}{0.97\linewidth}
  \setlength{\tabcolsep}{+0.0\linewidth}
  \renewcommand{\arraystretch}{0.3}

  \begin{tabular}{c}        
        
        \includegraphics[width=\cw]{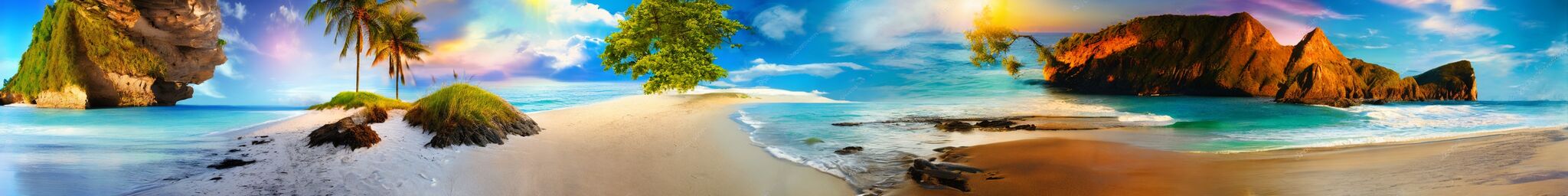} \\

        \includegraphics[width=\cw]{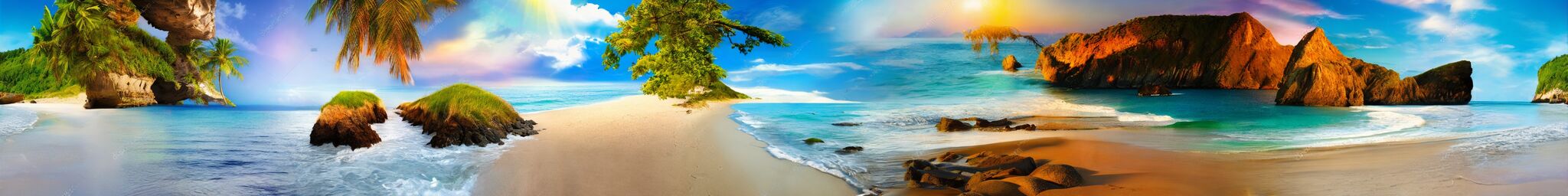} \\

        \includegraphics[width=\cw]{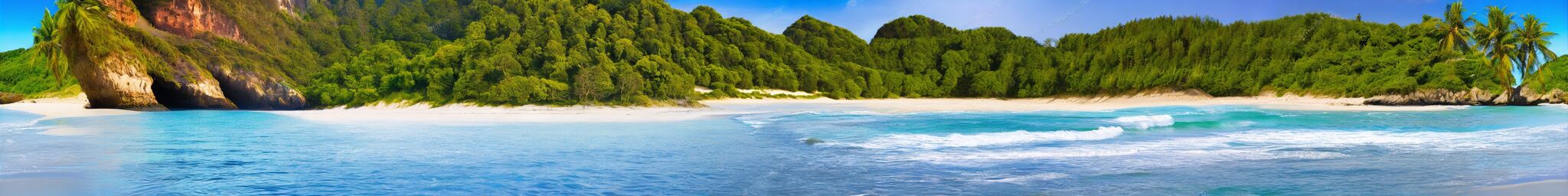} \\[0.3cm]

        \includegraphics[width=\cw]{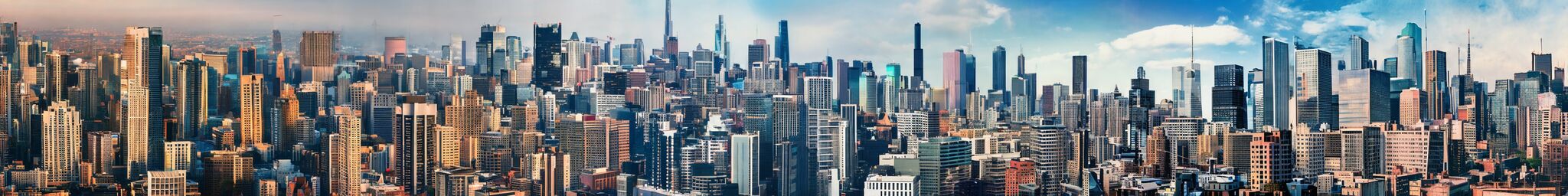} \\

        \includegraphics[width=\cw]{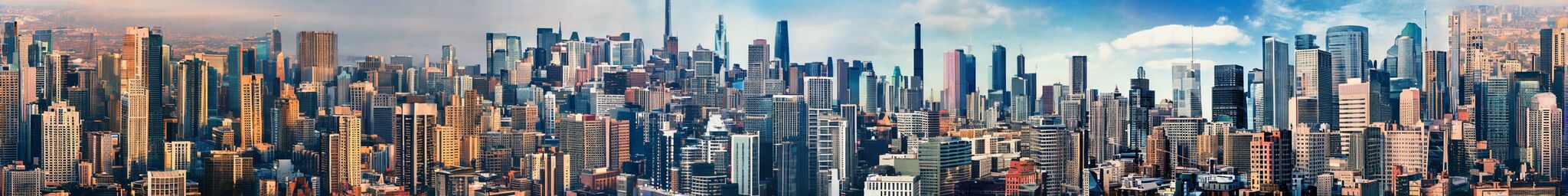} \\
        
        \includegraphics[width=\cw]{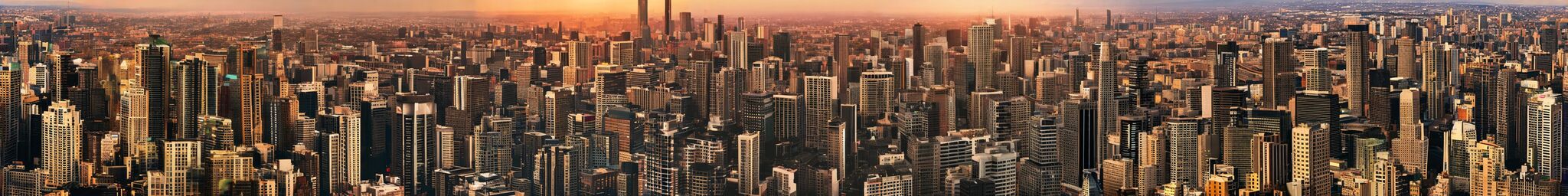}\\[0.3cm]

        \includegraphics[width=\cw]{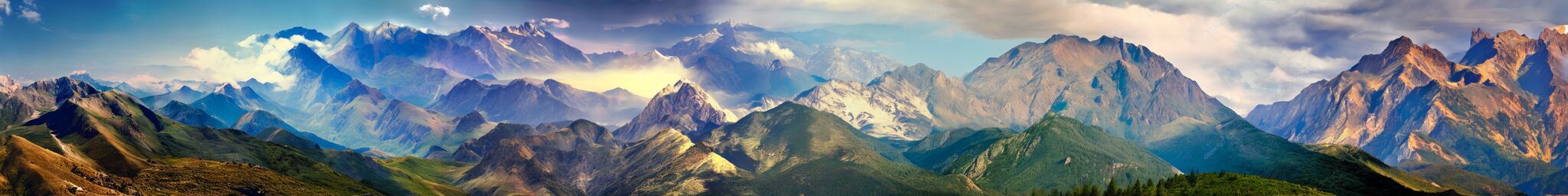} \\

        \includegraphics[width=\cw]{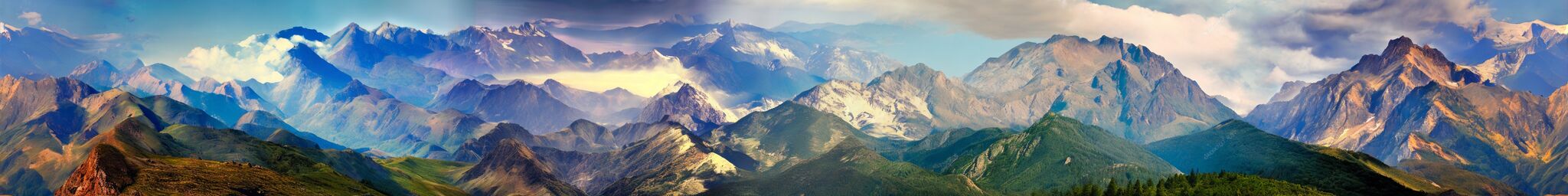} \\

        \includegraphics[width=\cw]{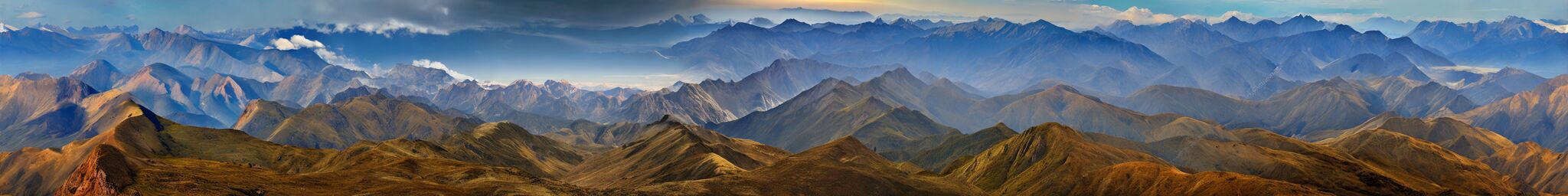} \\
            
    \end{tabular}
    \caption{\label{fig:control_uniform}%
        Application of our noise uniformization in conjunction with noise rolling~\cite{vecchio2024controlmat}. For each image triplet, the first row is obtained from \textbf{white noise using MultiDiffusion} with stride 1, 
        the second row represents image generation using \textbf{noise rolling from white noise}, the third row uses \textbf{noise rolling on top of our noise uniformization}.%
    }
    \Description{The figure show 3 groups of 3 images.}
\end{figure*}

While our noise-uniformization technique is designed for textures, the concept readily extends to more generic image types, such as panoramas.
We demonstrate this by incorporating our improvements in MultiDiffusion~\cite{omer2023multidiffusion} to generate large-scale images that are more realistic in terms of internal consistency.
The results of this experiment are presented in Fig.~\ref{fig:sd_uniform}, comparing our results with a vanilla MultiDiffusion based on Stable Diffusion $2.0$.
To highlight the benefits of the introduced improvements, we use a stride of 32 for the sliding window.
It can be easily seen that our results are more spatially consistent while still exhibiting a realistic amount of variation. This has the added advantage that seams are less noticeable, despite using a large stride.
The seams are almost completely eliminated when we additionally use our randomized sliding windows scheme. That is, instead of using a fixed stride, we offset each window by a random amount both horizontally and vertically. 
We limit the offset to be less than half of the stride to ensure full coverage of the latent map.
The seams seen in the first row of each group could also be resolved by running MultiDiffusion with a stride of 1; however, this would incur a significantly longer running time (18 minutes per image). Moreover, as seen in Fig.~\ref{fig:control_uniform}, the high-level inconsistencies remain even in this case.

We want to stress that our latent noise uniformization is complementary to the strategy used to generate large textures. 
For example, a competitive approach to the noise averaging employed by MultiDiffusion is the noise rolling algorithm introduced in ControlMat~\cite{vecchio2024controlmat}. 
In Fig.~\ref{fig:control_uniform}, we combine our noise uniformization with the noise rolling method for generating arbitrarily large images and show that the results are significantly improved.
Our uniformization technique is most useful when there is a high diversity of different images that would fit a given prompt, as it makes it difficult to reconcile adjacent patches (e.g. beach landscape, Fig~\ref{fig:control_uniform}).
The noise rolling algorithm has the advantage that each pixel is used just once for each timestep in the diffusion process (equivalent to using a stride of 64). The algorithm is thus significantly faster compared to the full MultiDiffusion (shown in the first row of each group in Fig.~\ref{fig:control_uniform}), with virtually no loss in quality. 
\citet{wang2024infinite} also proposed a similar way to improve upon MultiDiffusion in regards to the sampling of patches to be denoised. We, however, do not combine our noise uniformization with this method as their mechanism for sampling the random patches is not described with sufficient detail.

\section{Additional details on the tileable texture synthesis of arbitrary size.}
\label{asec:tileable}

\begin{figure*}
  \centering
  \setlength{\cw}{0.24\linewidth}
  \setlength{\tabcolsep}{+0.002\linewidth}
  \begin{tabular}{cc@{\quad}cc}
        \includegraphics[width=\cw]{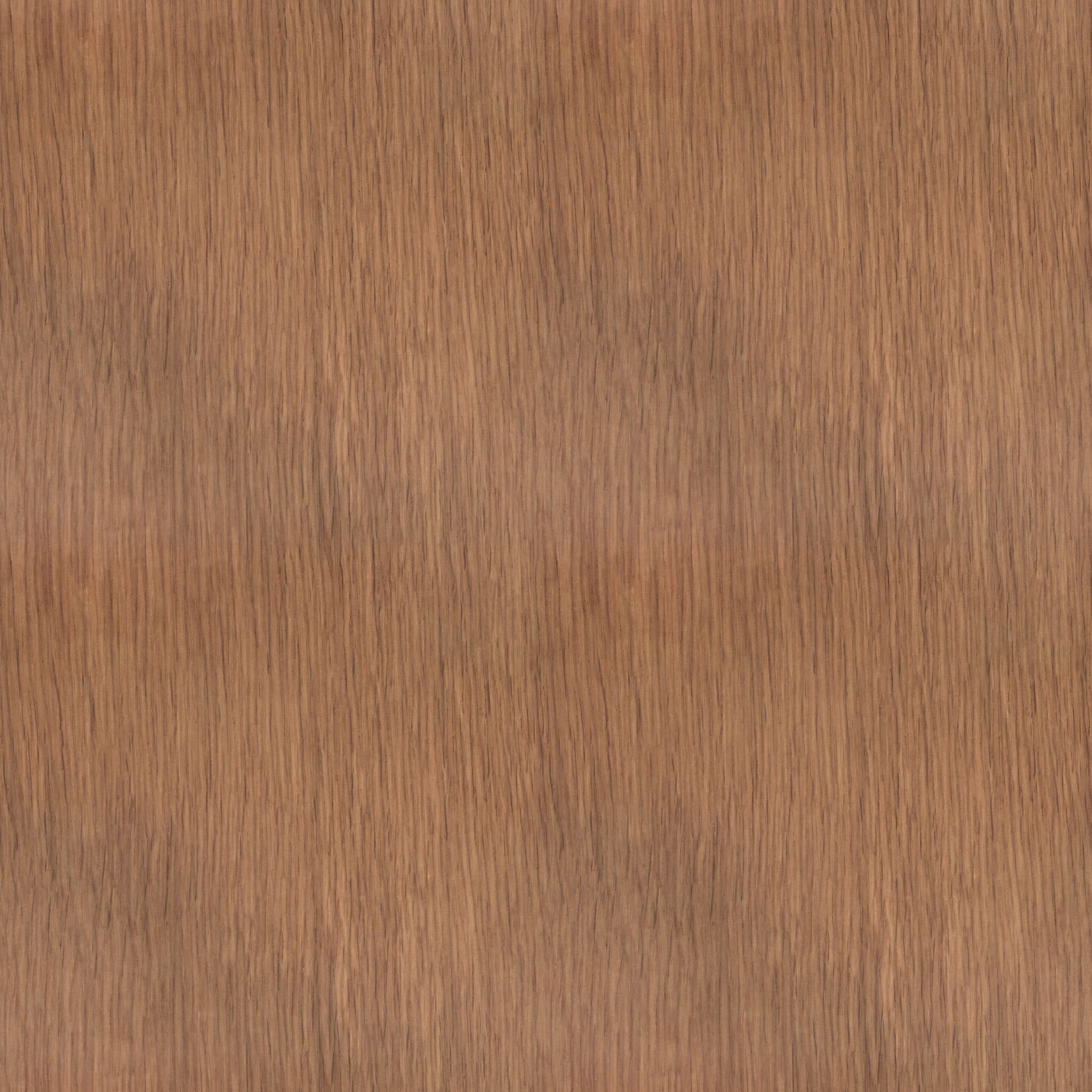} &
        \includegraphics[width=\cw]{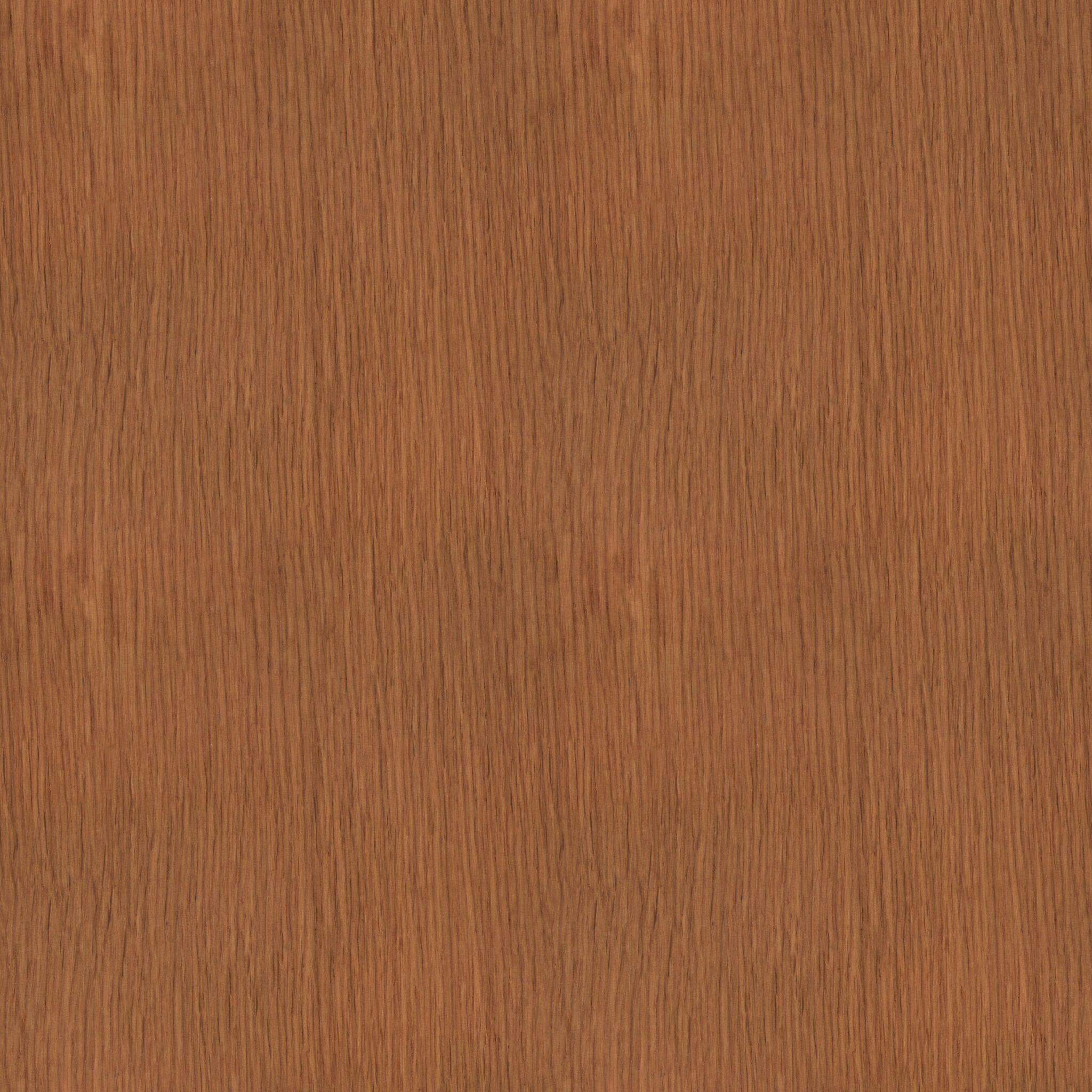} &
        \includegraphics[width=\cw]{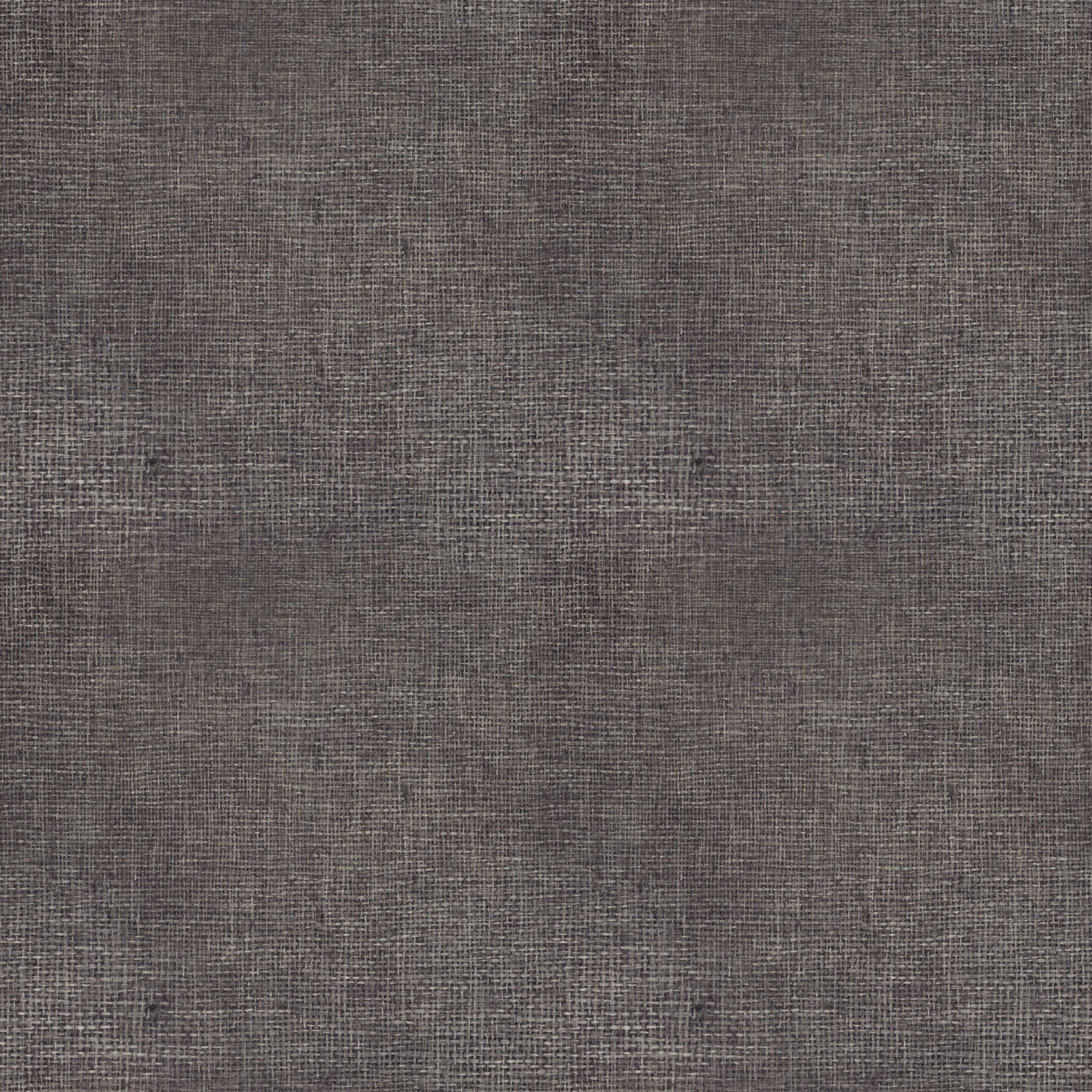} &
        \includegraphics[width=\cw]{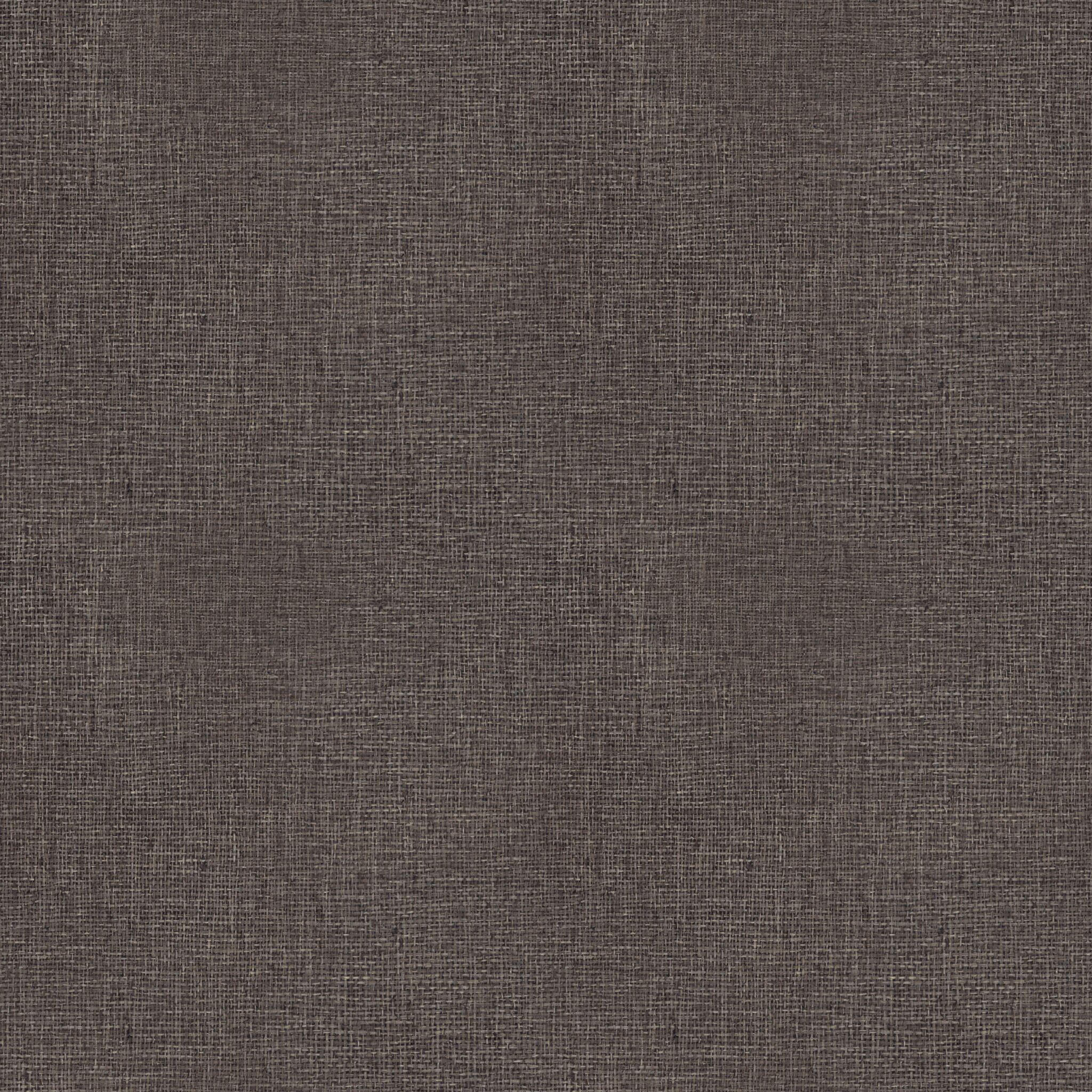}\\
        
        \includegraphics[width=\cw]{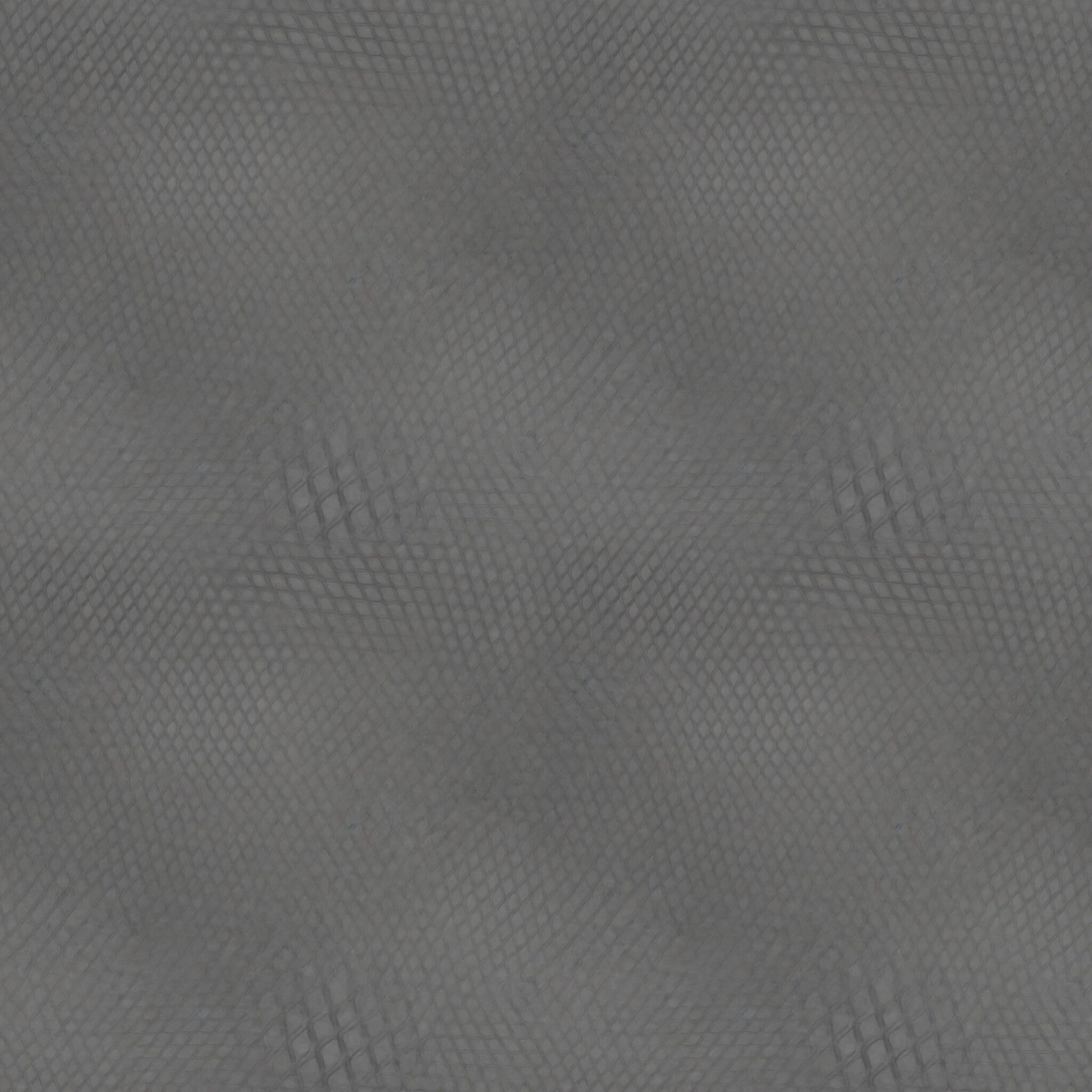} &
        \includegraphics[width=\cw]{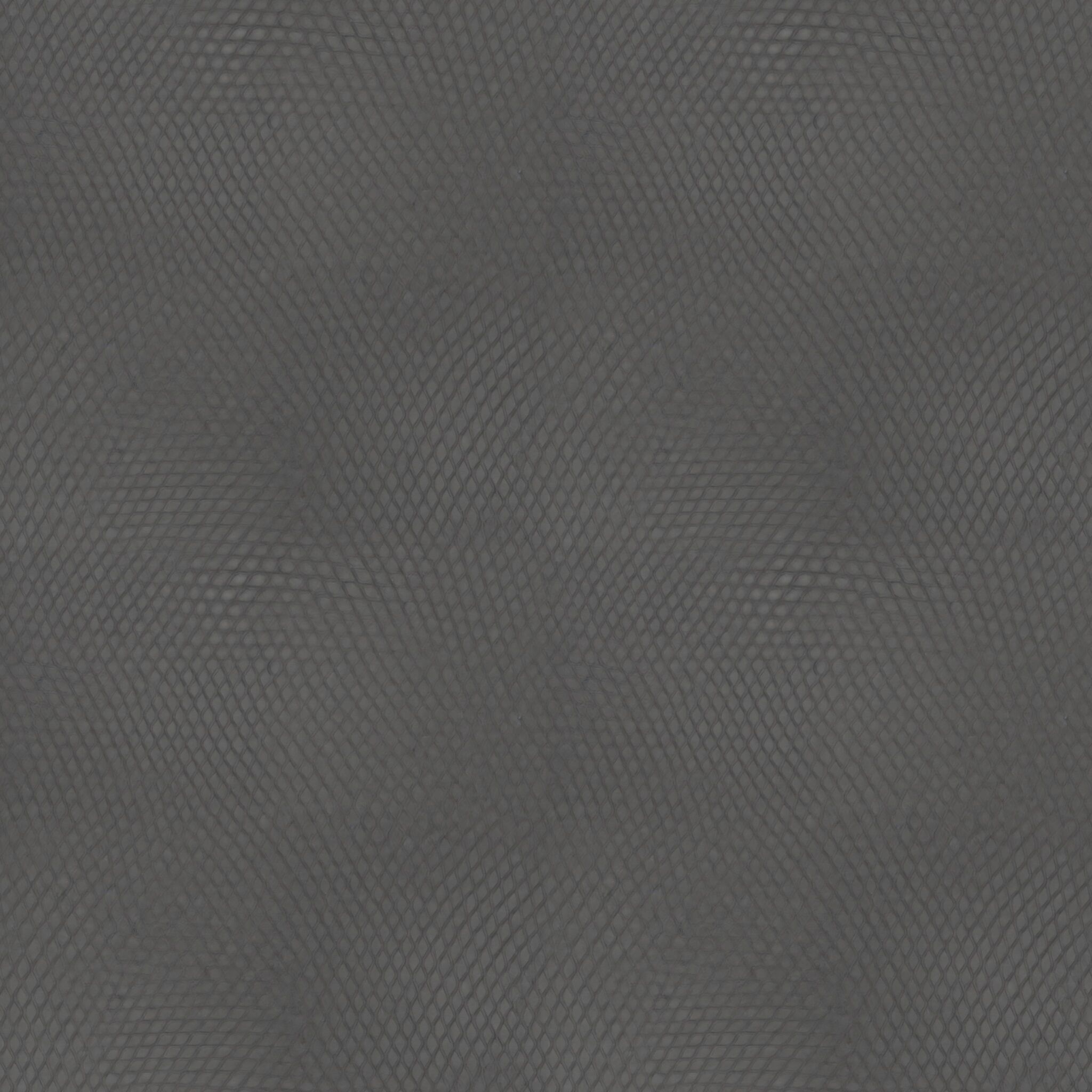} &
        \includegraphics[width=\cw]{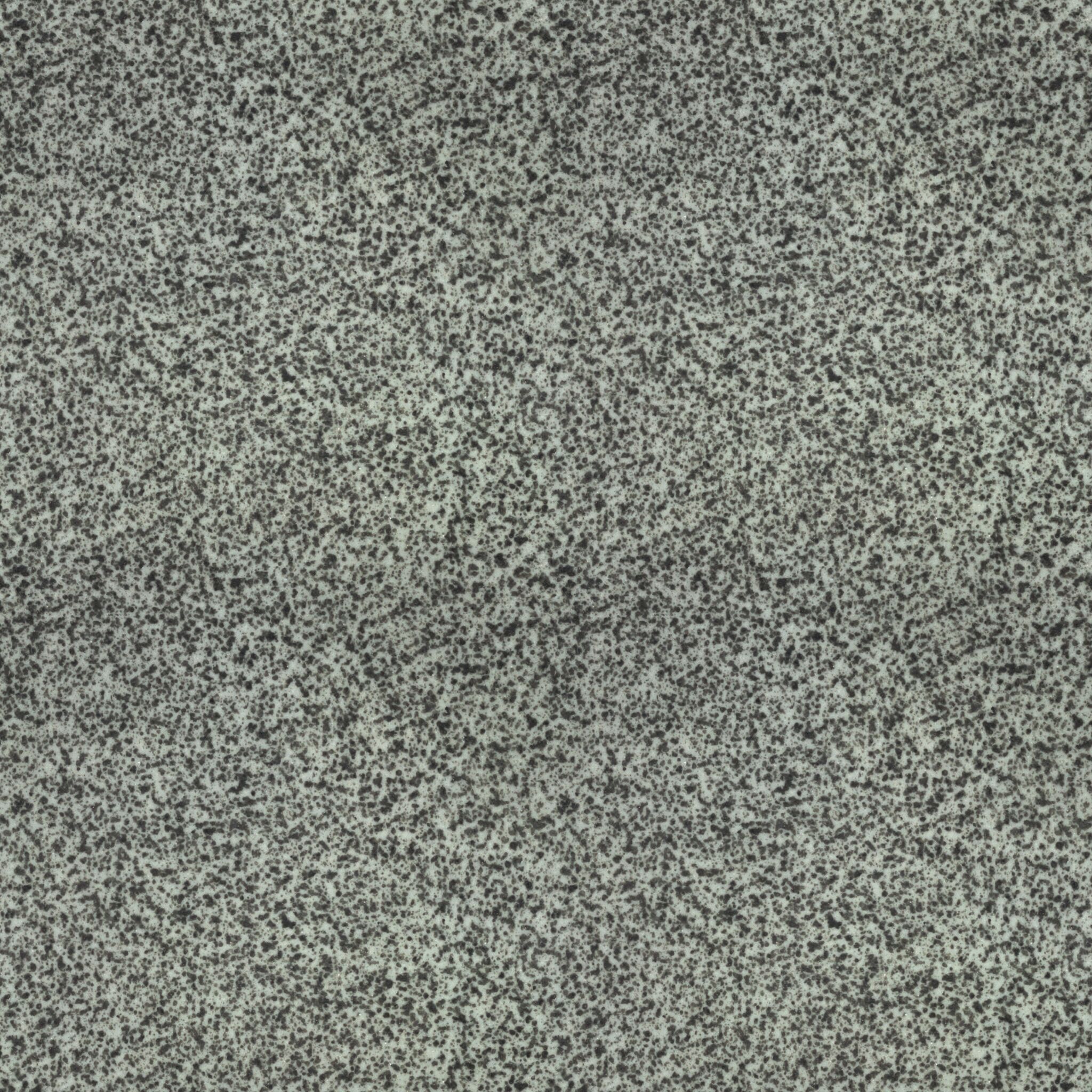} &
        \includegraphics[width=\cw]{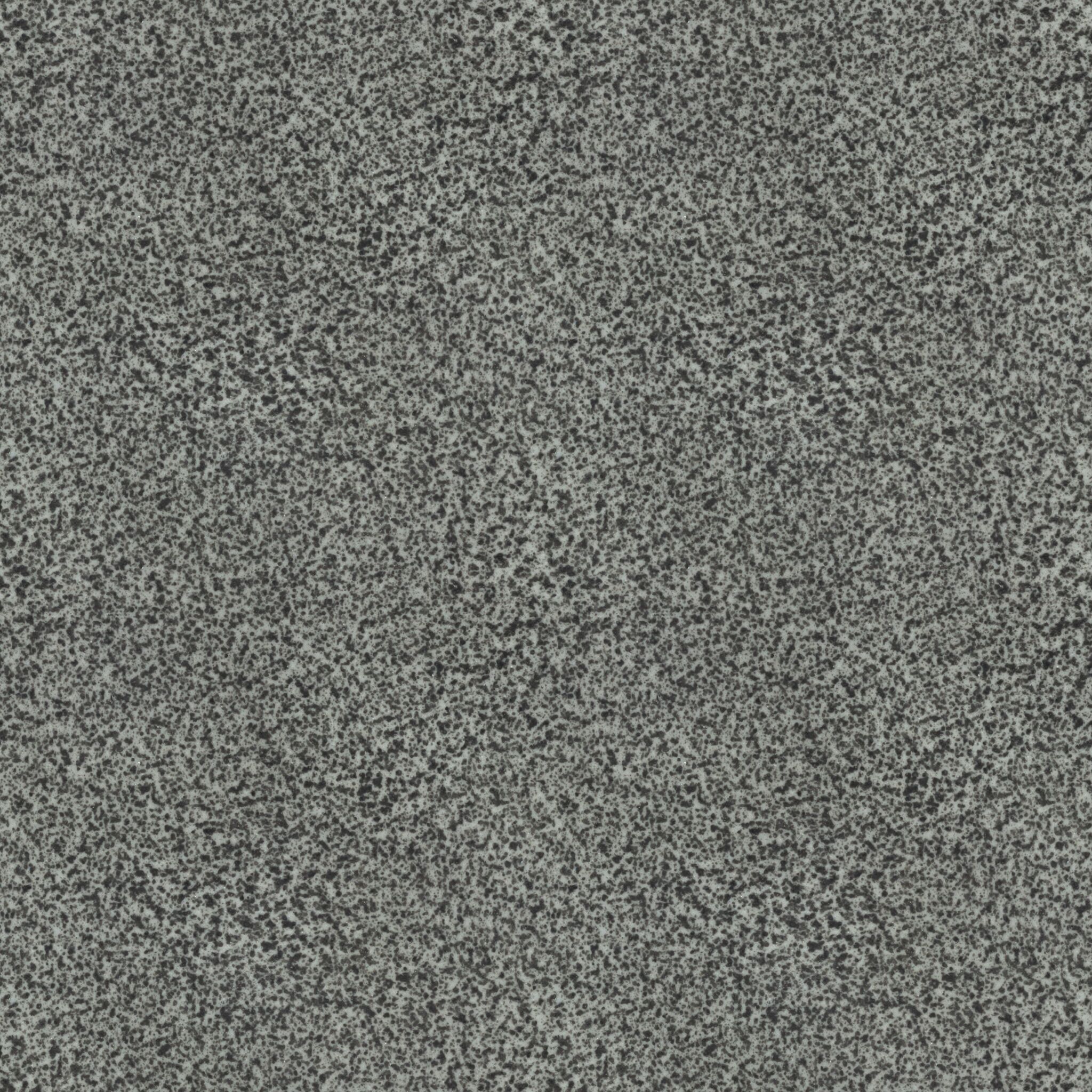} \\
          
        \includegraphics[width=\cw]{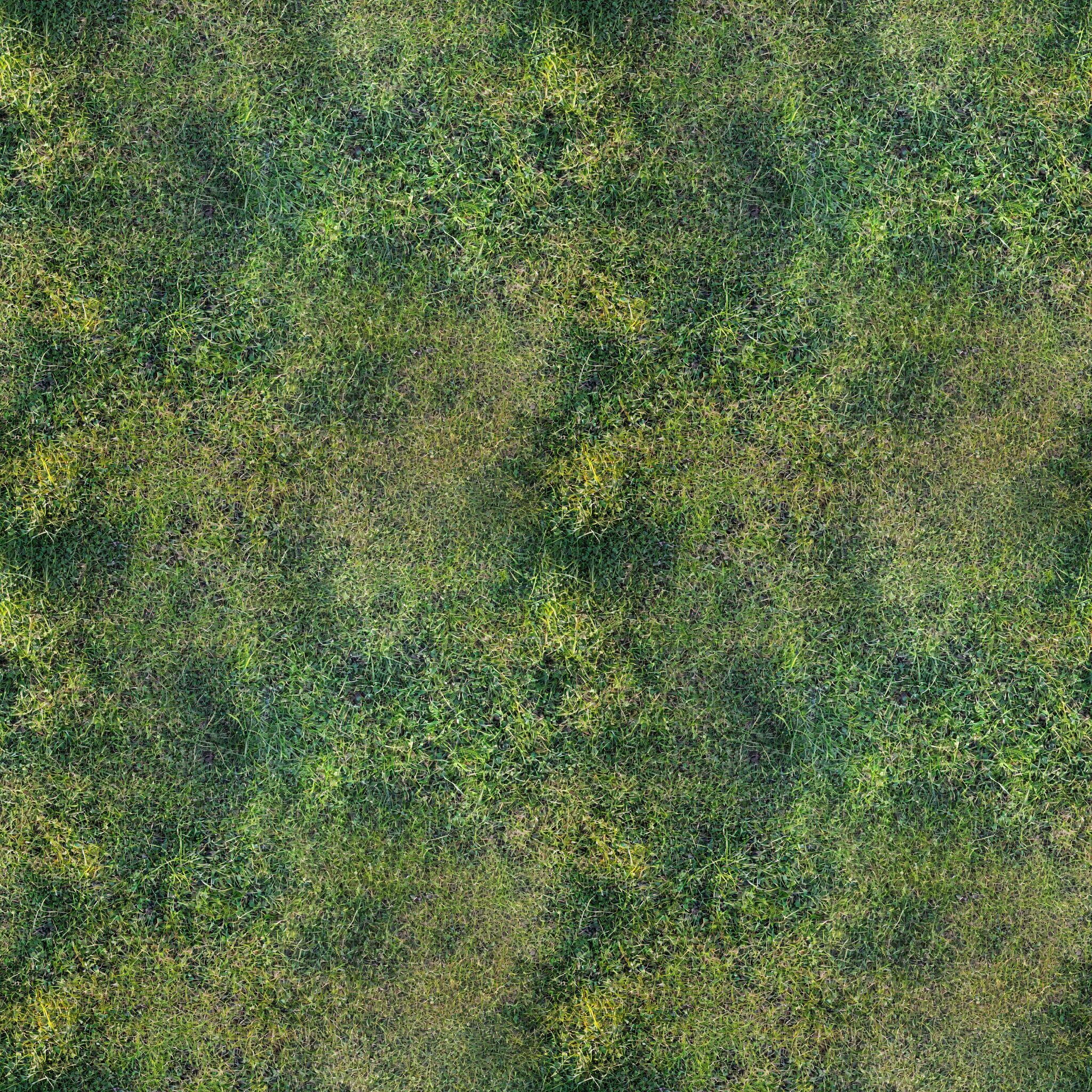} &
        \includegraphics[width=\cw]{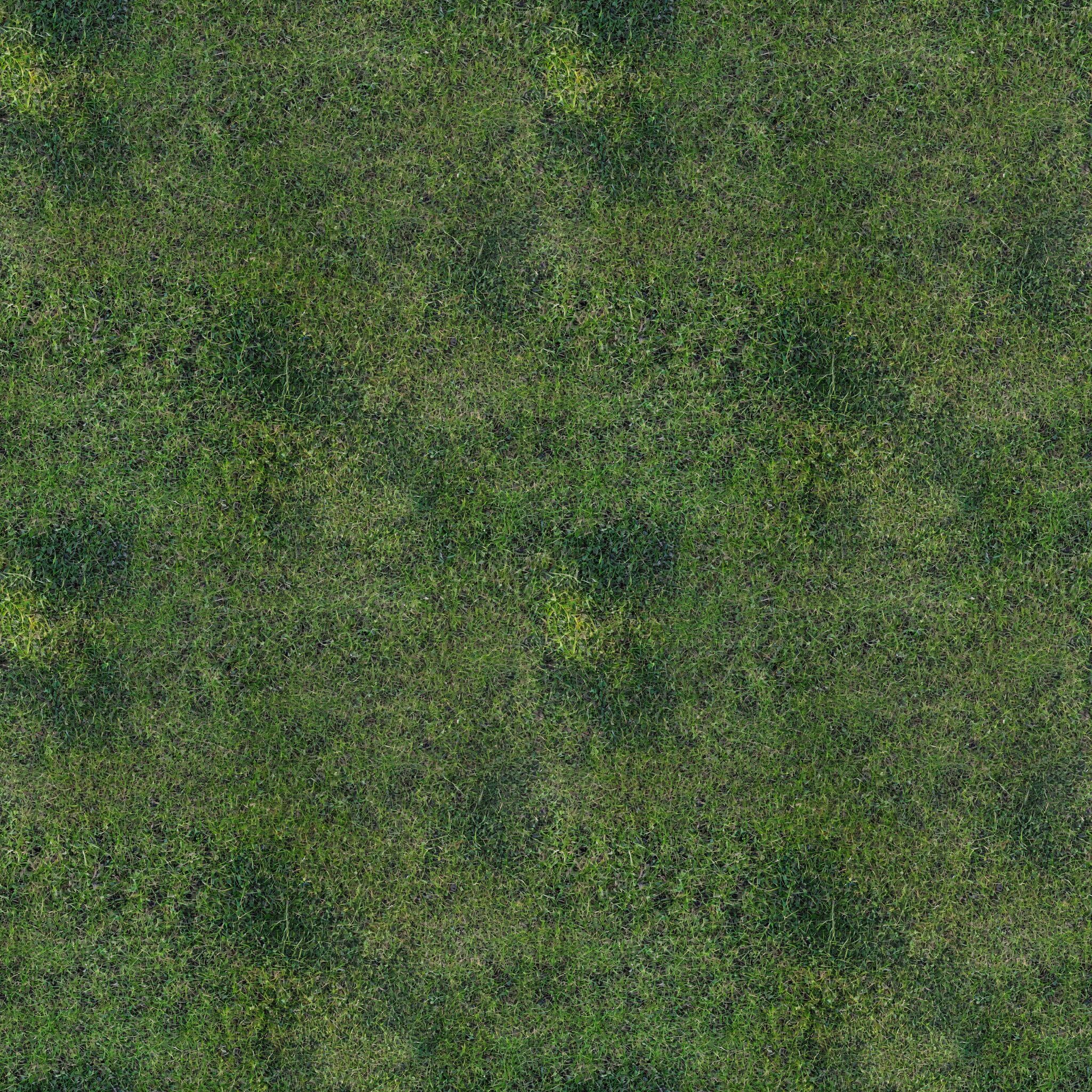} &
        \includegraphics[width=\cw]{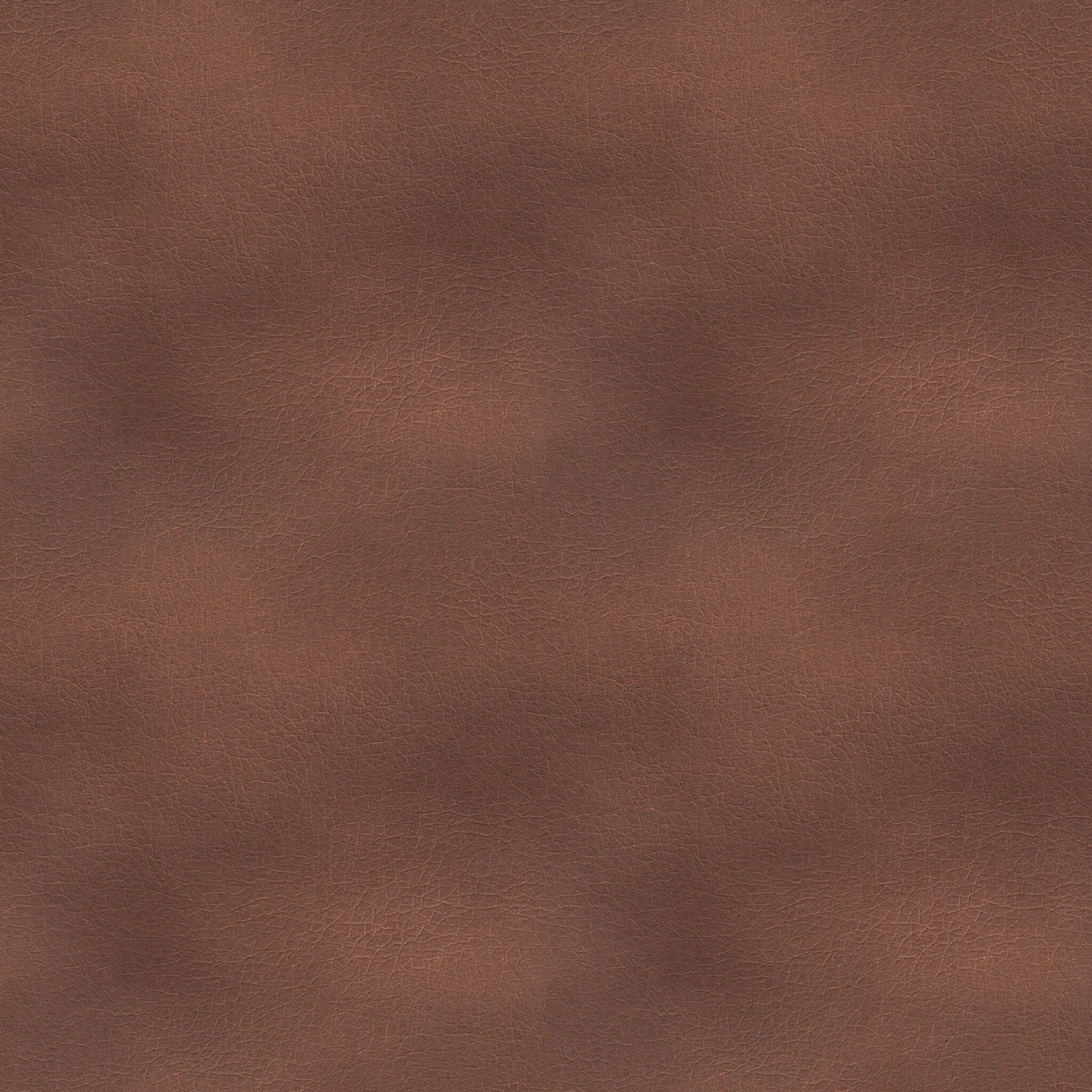} &
        \includegraphics[width=\cw]{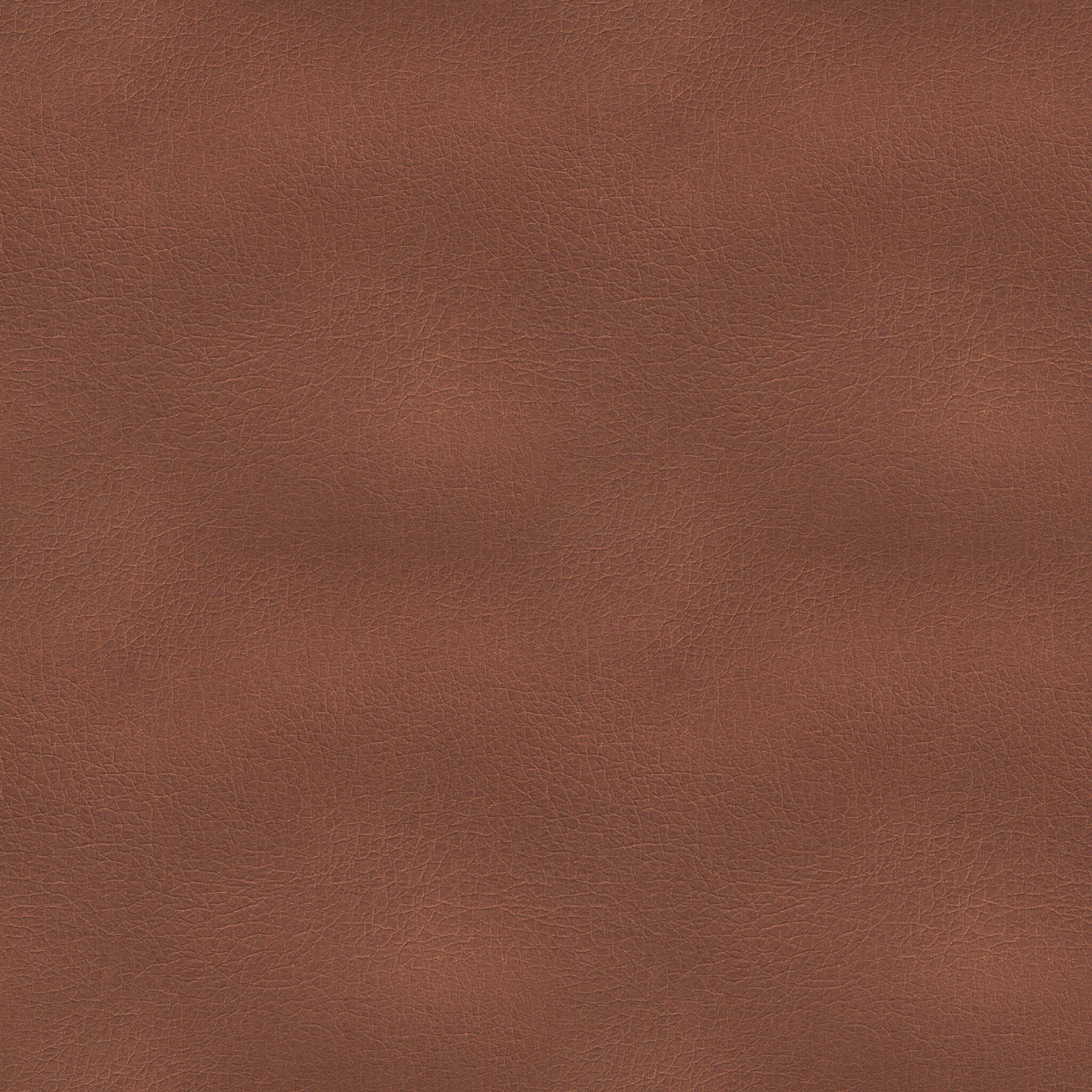}\\
    \end{tabular}

    \Description{The Figure tiled results generated by our method.}
    \caption{\label{fig:tileable}%
        Visualization of the generated tiled textures starting from white noise (left column) and our noise-uniformization method (right column). The multi-diffusion synthesizes tileable textures of 2048\texttimes{}2048, which are then tiled to form 4096\texttimes{}4096 images.
    }
\end{figure*}

Section~\ref{sec:infinity} of the main paper presents our approach to generating textures of arbitrary size using constant GPU memory. Crucially, our noise-uniformization preprocessing ensures that the generated textures are stationary.
To avoid seams caused by the patch-based denoising, we set the pixel overlap range to 32 (\ie, half the window size). This means the number of evaluations of the diffusion model is 4 times larger compared to an independent generation of the same number of pixels.
This constant factor aside, the complexity of texture synthesis is linear with respect to the images resolution, meaning that generating large textures in this manner is time consuming (\eg, 170 seconds for a 4096\texttimes{}4096 image). 
We approach this issue by making our textures tileable; this is done by wrapping around the sliding windows circularly during denoising.
This strategy directly ensures tileability in the latent space in which the diffusion model operates.
In order to obtain tileable texture in RGB space, we first circularly pad the output of the multi-diffusion.
Then, the enlarged map is decoded using the Stable Diffusion VAE~\cite{rombach2022high}, followed by a crop to the original (unpadded size). The decoded output can then be easily tiled.
This allows trading quality for efficiency by synthesizing textures up to a certain resolution, after which tiling is used to achieve the final size.

In Fig.~\ref{fig:tileable} we present 6 such textures, that were generated at a 2048\texttimes{}2048 resolution and then tiled into 2\texttimes{}2 blocks, resulting in seamless 4096\texttimes{}4096 images.
Our noise-uniformization technique reduces the impression of repeatability by making each tile more stationary. This improvement especially visible for the grid texture (second row of Fig.~\ref{fig:tileable}).

One can also note a difference in overall color and contrast between the images in the two columns. 
This appears as a consequence of the noise averaging, specific to MultiDiffusion~\cite{omer2023multidiffusion}.
Averaging different diffusion paths reduces variance and changes the final result for a specific window compared to what would have been obtained using the noise in that window in isolation.
Thanks to our noise uniformization, the first moment of the noise varies much less over different windows, which keeps the denoising trajectory from diverting too much from the original path.

\section{Conditional generation on MVTec Textures}
\begin{figure*}
  \centering%
  \setlength{\cw}{0.1884\linewidth}
  \setlength{\tabcolsep}{+0.003\linewidth}
  \begin{tabular}{ccccc}
    \figvlabel{Input labels}%
    \includegraphics[width=\cw]{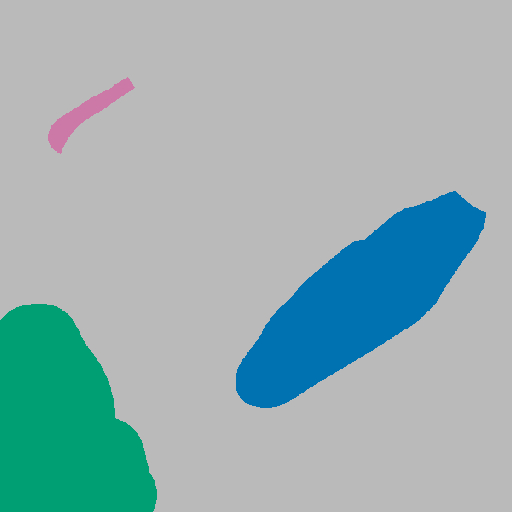} &
    \includegraphics[width=\cw]{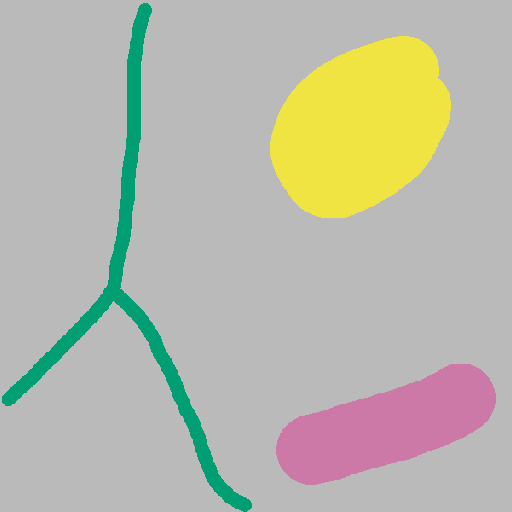} &
    \includegraphics[width=\cw]{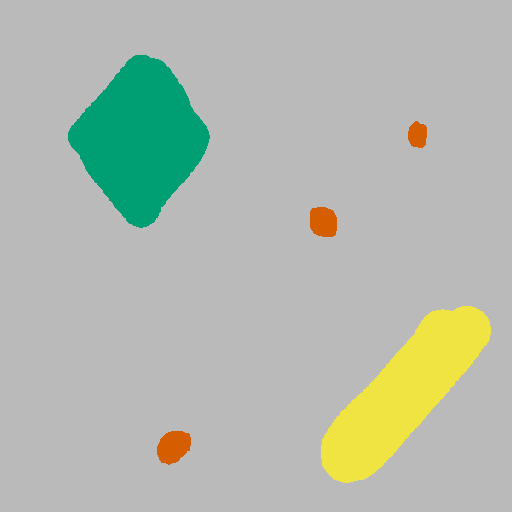} &
    \includegraphics[width=\cw]{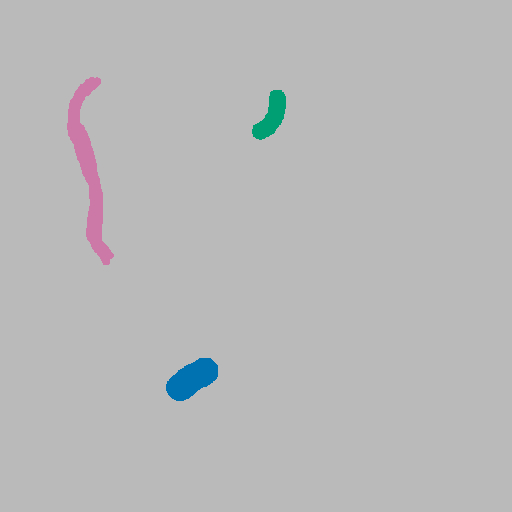} &
    \includegraphics[width=\cw]{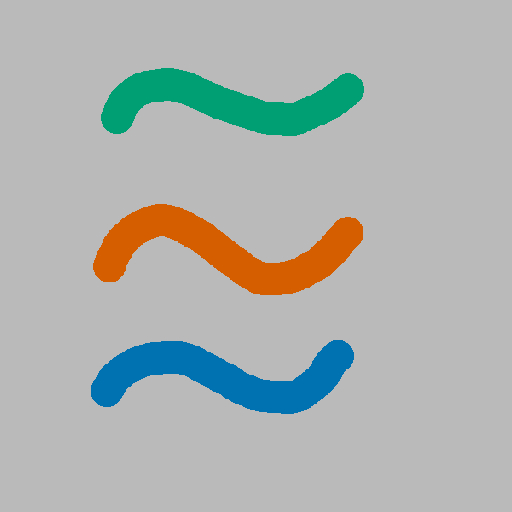} \\
    
    \figvlabel{Synthesis}%
    \includegraphics[width=\cw]{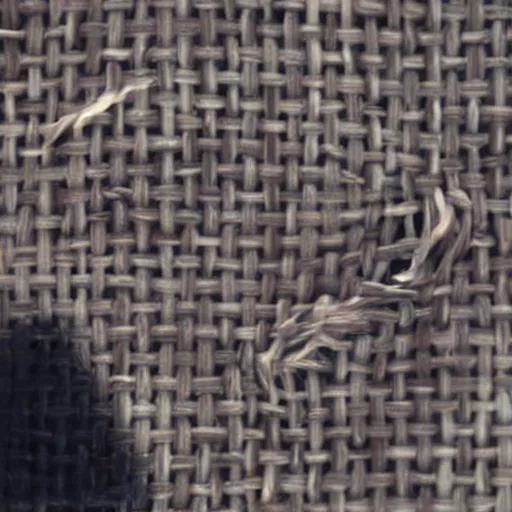} &
    \includegraphics[width=\cw]{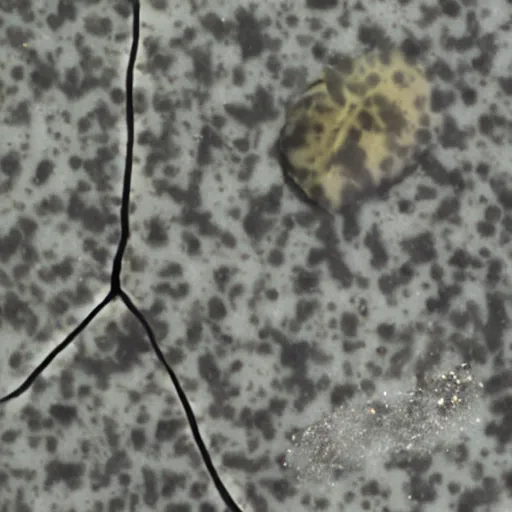} &
    \includegraphics[width=\cw]{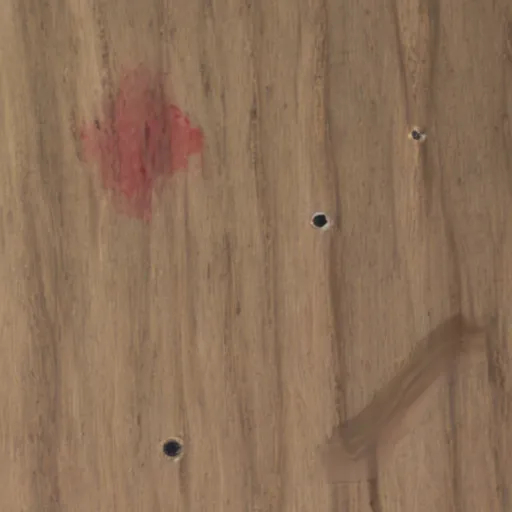} &
    \includegraphics[width=\cw]{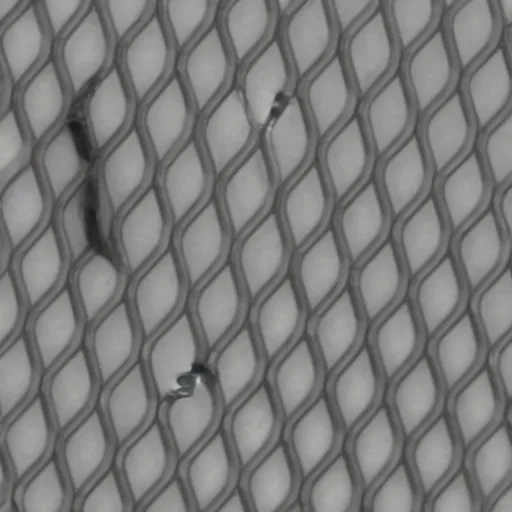} &
    \includegraphics[width=\cw]{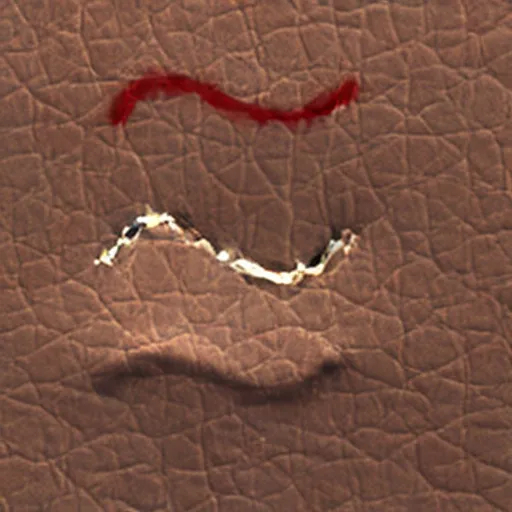} \\
    
  \end{tabular}
  \caption{\label{fig:ml_synthesis_mvtec}%
    Additional results on feature-conditioned texture generation by our model.%
  }
  \Description{The Figure shows a generated texture for each texture class in the MVTecDataset and our pavement texture class.}
\end{figure*}
In Fig.~\ref{fig:ml_synthesis_mvtec} we show additional results generated with our method. The results on MVTecAD textures shows that our method generalizes to diverse shapes and multiple anomaly types per image.

\section{Uncurated set of generated images}
\label{asec:many_icons}

\begin{figure*}
  \centering%
  \begin{tikzpicture}[scale=1.14, transform shape]
    \def\groupsep{0.006cm} %
    \def\imgsep{0.003cm} %
    \def\imgwidth{0.015cm} %

    \def\rows{7}
    \def\cols{5}

    \def\grouprows{5}
    \def\groupcols{5}

    \foreach \row in {1,...,\rows} {
      \foreach \col in {1,...,\cols} {
        \pgfmathsetmacro{\xstart}{(\col-1) * (\groupcols * \imgwidth + (\groupcols-1) * \imgsep) + (\col-1) * \groupsep}
        \pgfmathsetmacro{\ystart}{-(\row-1) * (\grouprows * \imgwidth + (\grouprows-1) * \imgsep) - (\row-1) * \groupsep}

        \pgfmathsetmacro{\iconid}{int((\row-1)*\cols + \col)}

        \foreach \grouprow in {1,...,\grouprows} {
          \foreach \groupcol in {1,...,\groupcols} {
            \pgfmathsetmacro{\x}{\xstart + (\groupcol-1) * (\imgwidth + \imgsep)}
            \pgfmathsetmacro{\y}{\ystart - (\grouprow-1) * (\imgwidth + \imgsep)}
            
            \node[anchor=north west, inner sep=0] at (\x, \y) {\includegraphics[width=0.5cm]{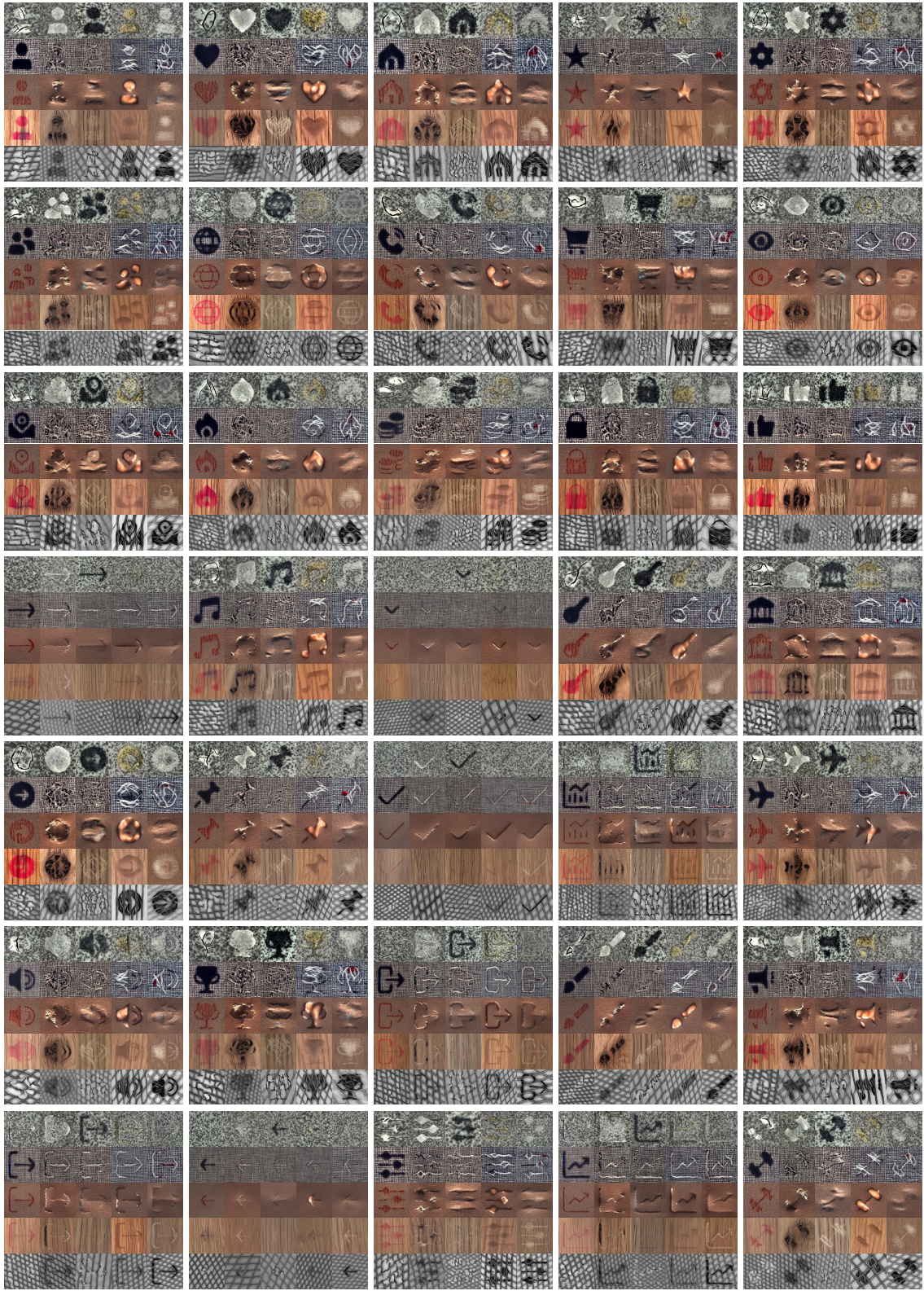}};
          }
        }
      }
    }
  \end{tikzpicture}
  \caption{\label{fig:many_icons}%
    Large set of uncurated images generated by our model based on icons from Flaticon.com. There are 7\texttimes{}5 different icons, for which we generate images with 5 anomalies for 5 different textures, for a total of 875 images.}
  \Description{
    A grid of 7x5 sets of 5x5 textures with painted features, showing different icons.
  }
\end{figure*}

In Fig.~\ref{fig:many_icons} we present a large set of images generated by our model, showing each texture-class feature-type combination for 35 different monochrome SVG icons from the internet (Flaticon.com).

\section{Additional high-resolution results}
\label{asec:high_res}

\begin{figure*}
    \centering
    \input{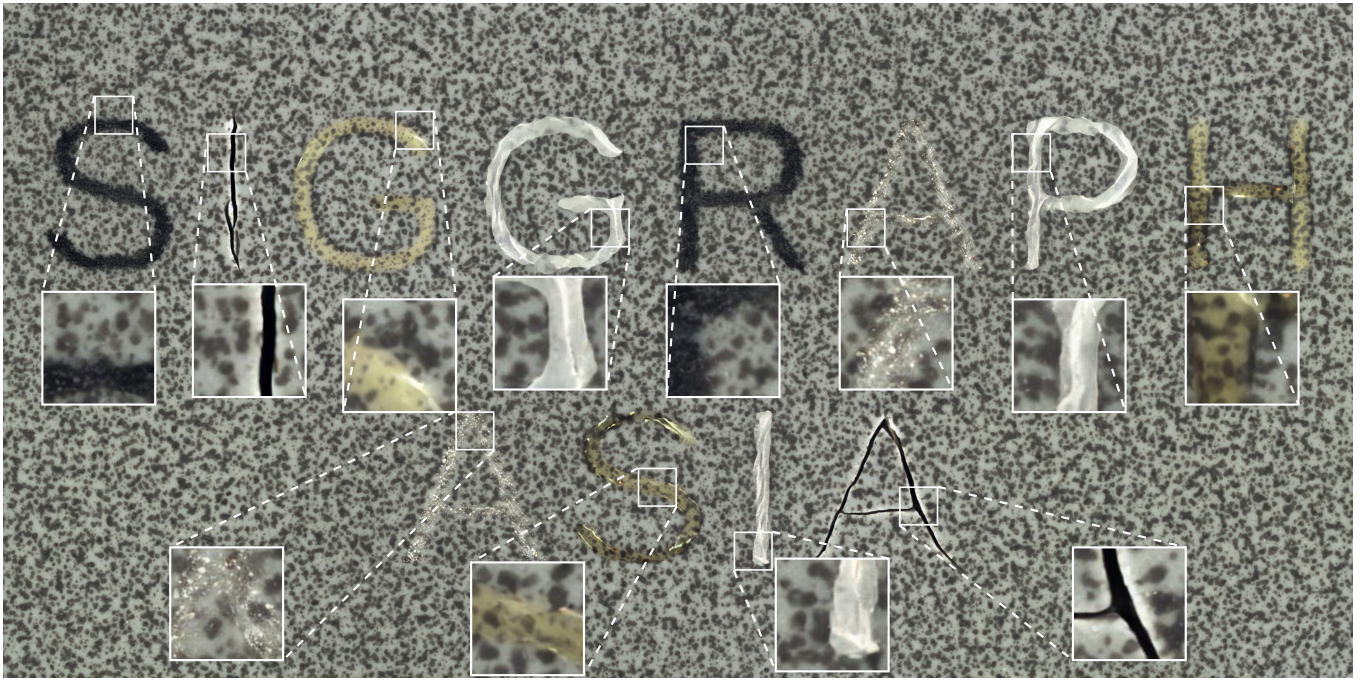}
    \caption{\label{fig:high_res_gen_1}%
      High-resolution (2048\texttimes{}4096) conditional generation example.}
    \Description{Large image reading SIGGRAPH}
\end{figure*}

\begin{figure*}
    \centering
    \includegraphics[width=0.49\textwidth]{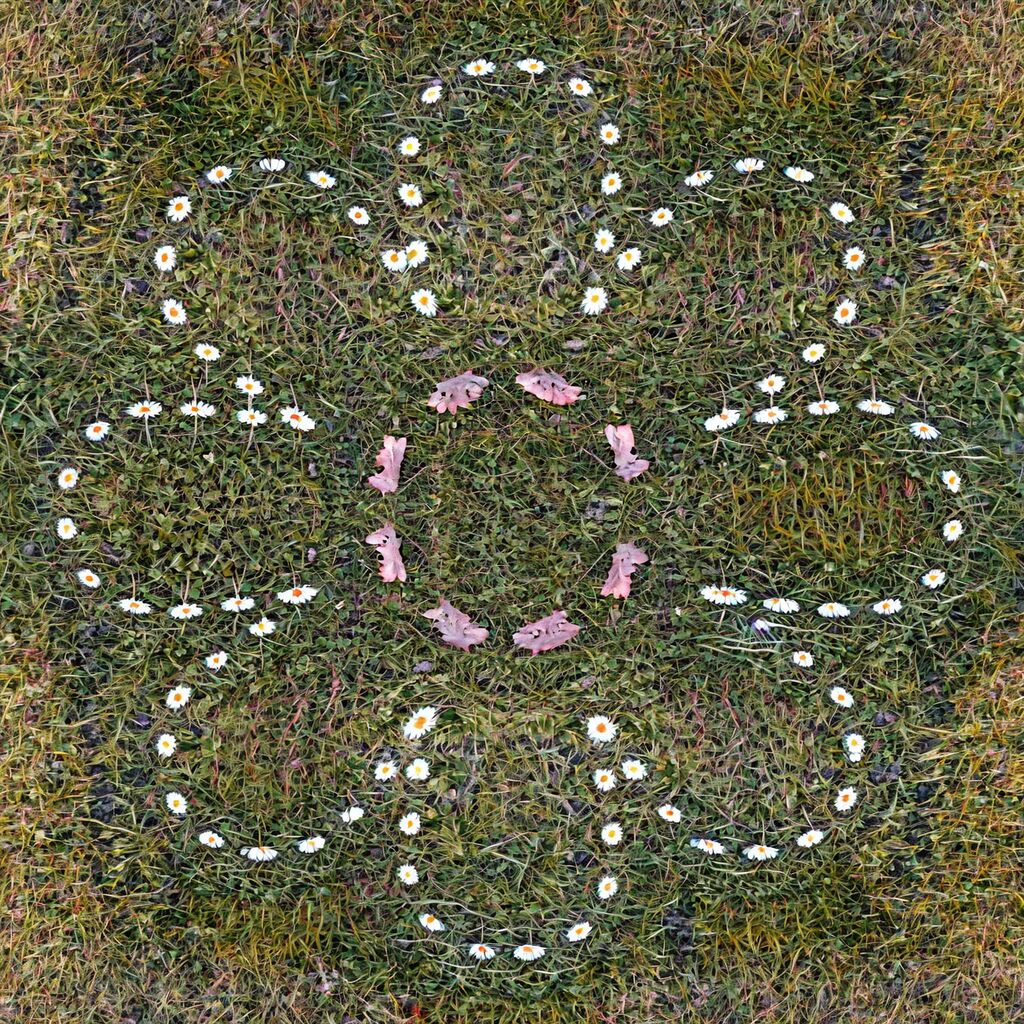}
    \includegraphics[width=0.49\textwidth]{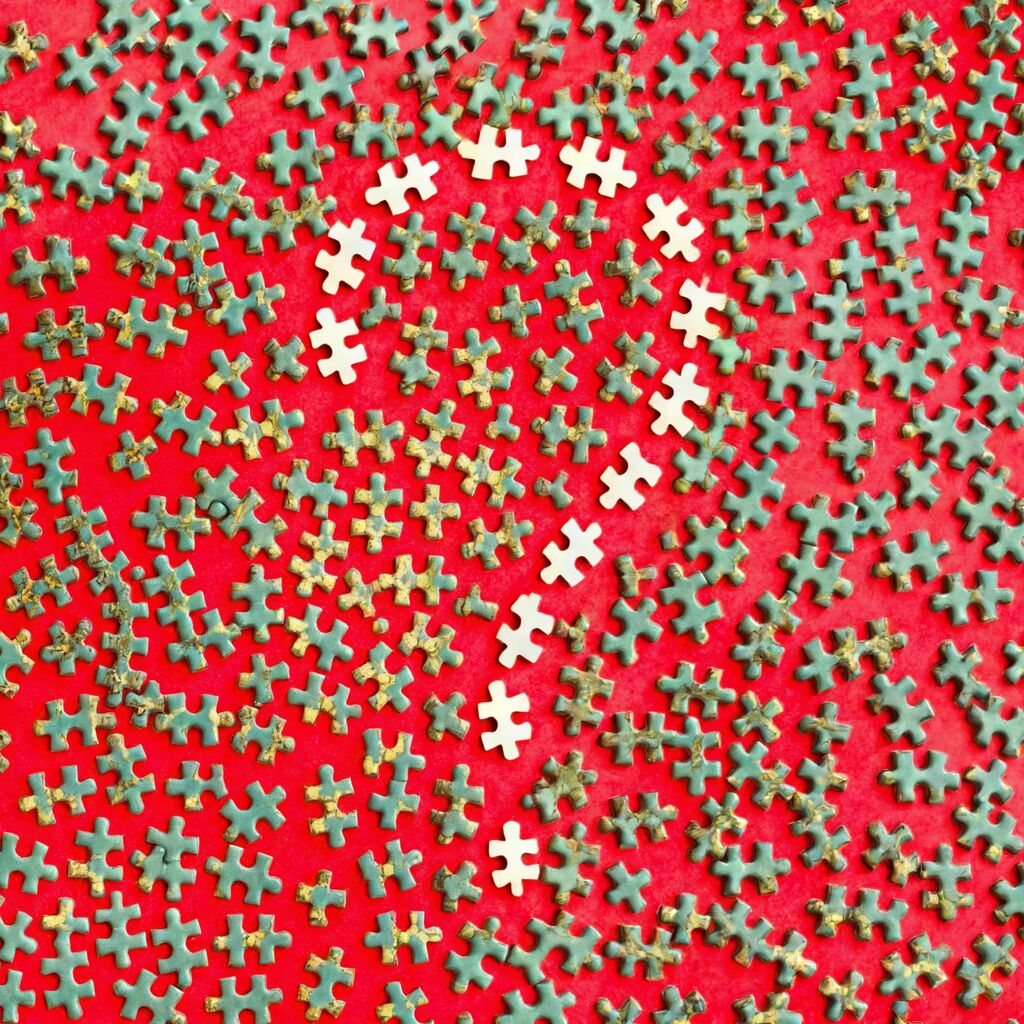}
    \caption{\label{fig:high_res_gen_2}%
      High-resolution conditional generation examples. \emph{Note that the puzzle model was trained from a single image (see Fig.~\ref{fig:dataview_our})}}
    \Description{Large images showing the outline of a flower and a question mark}
\end{figure*}

\begin{figure*}
    \centering
    \includegraphics[width=0.49\textwidth]{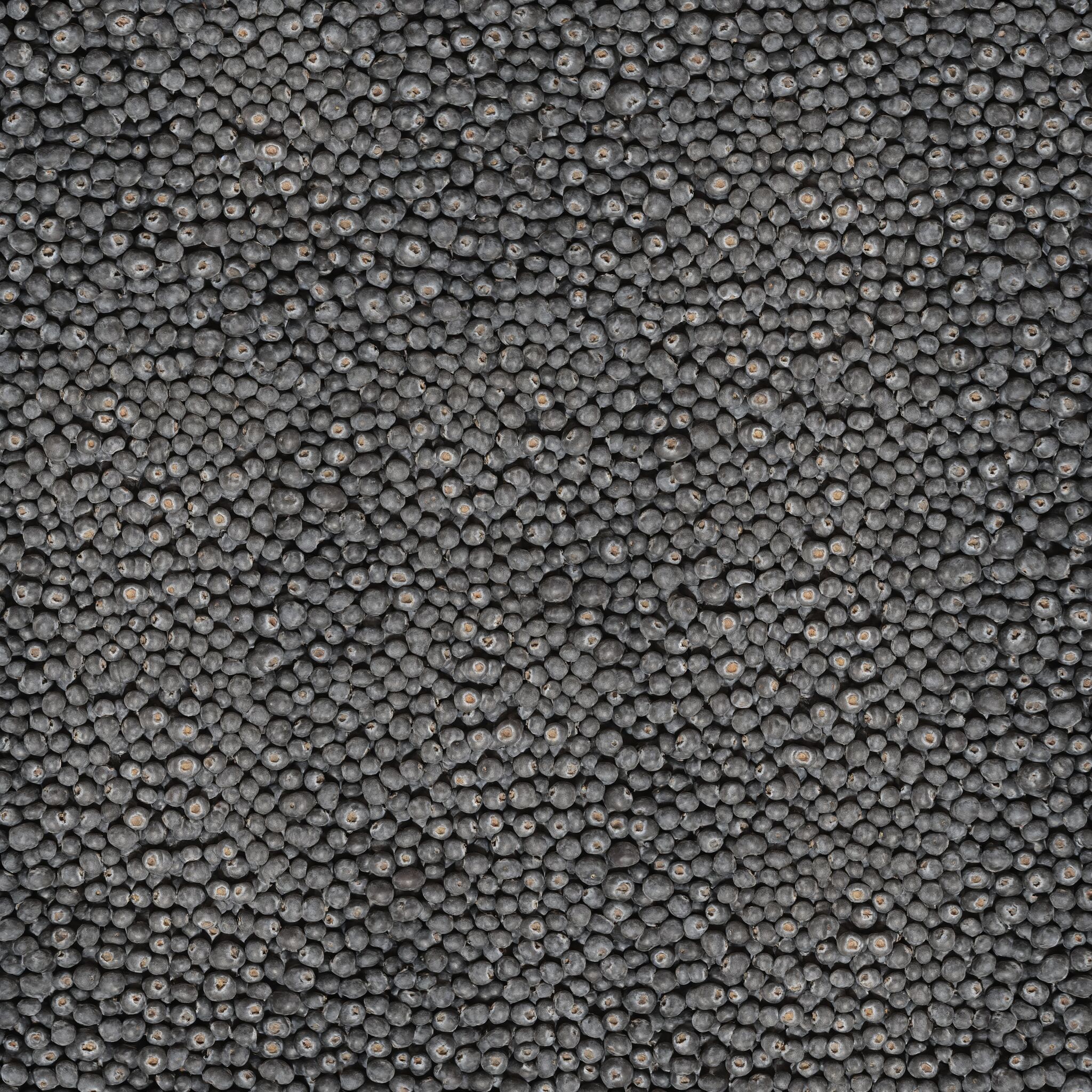}
    \includegraphics[width=0.49\textwidth]{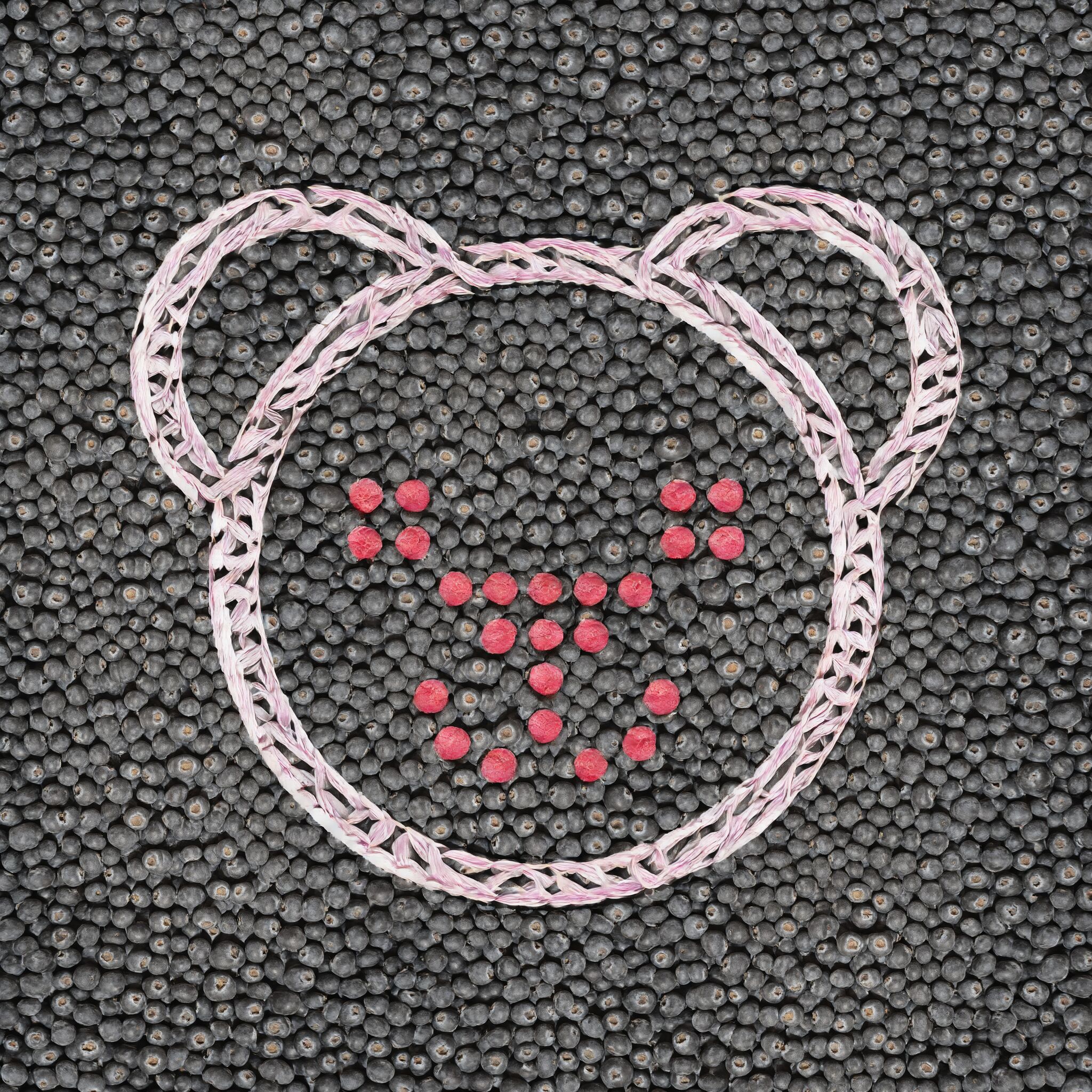}
    \includegraphics[width=0.49\textwidth]{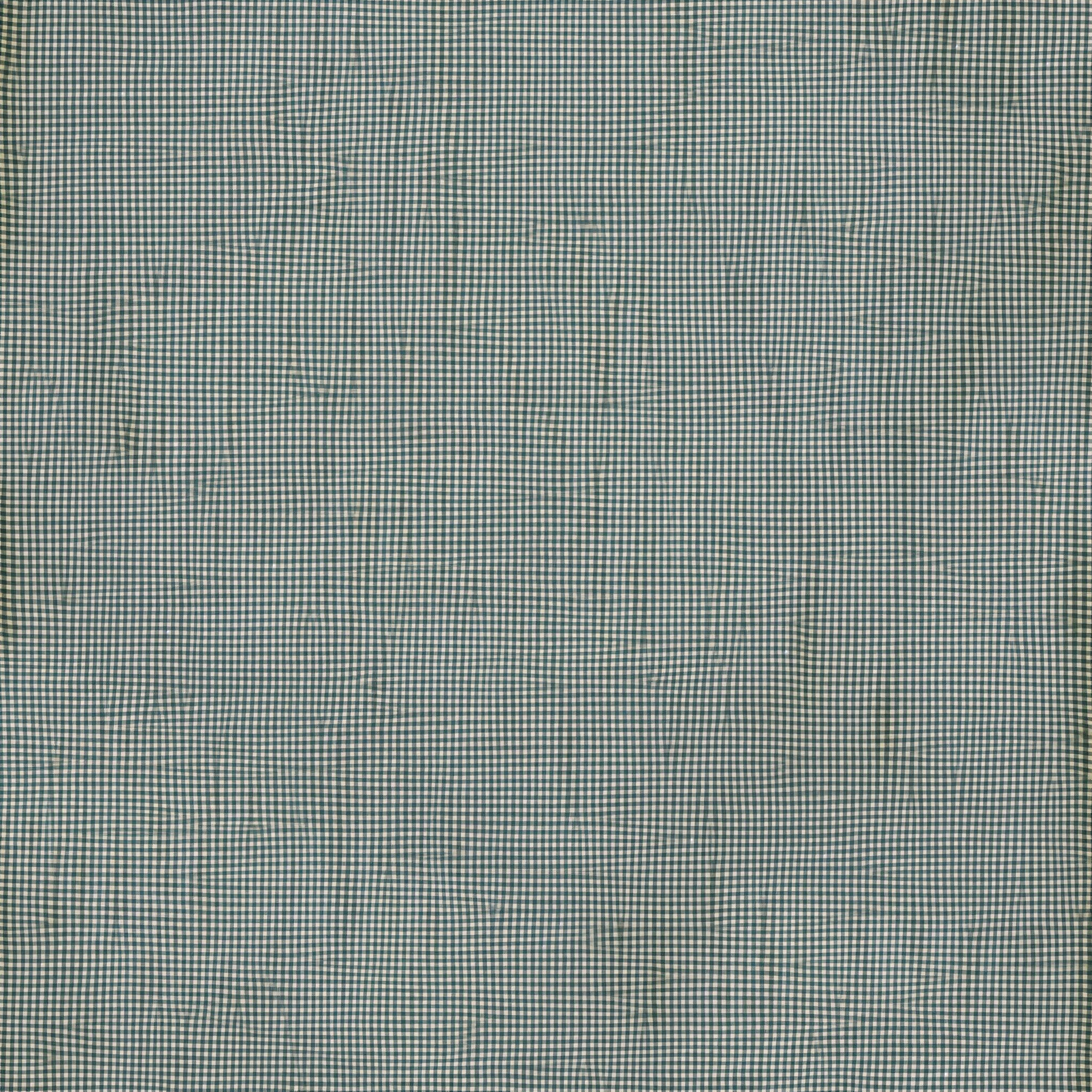}
    \includegraphics[width=0.49\textwidth]{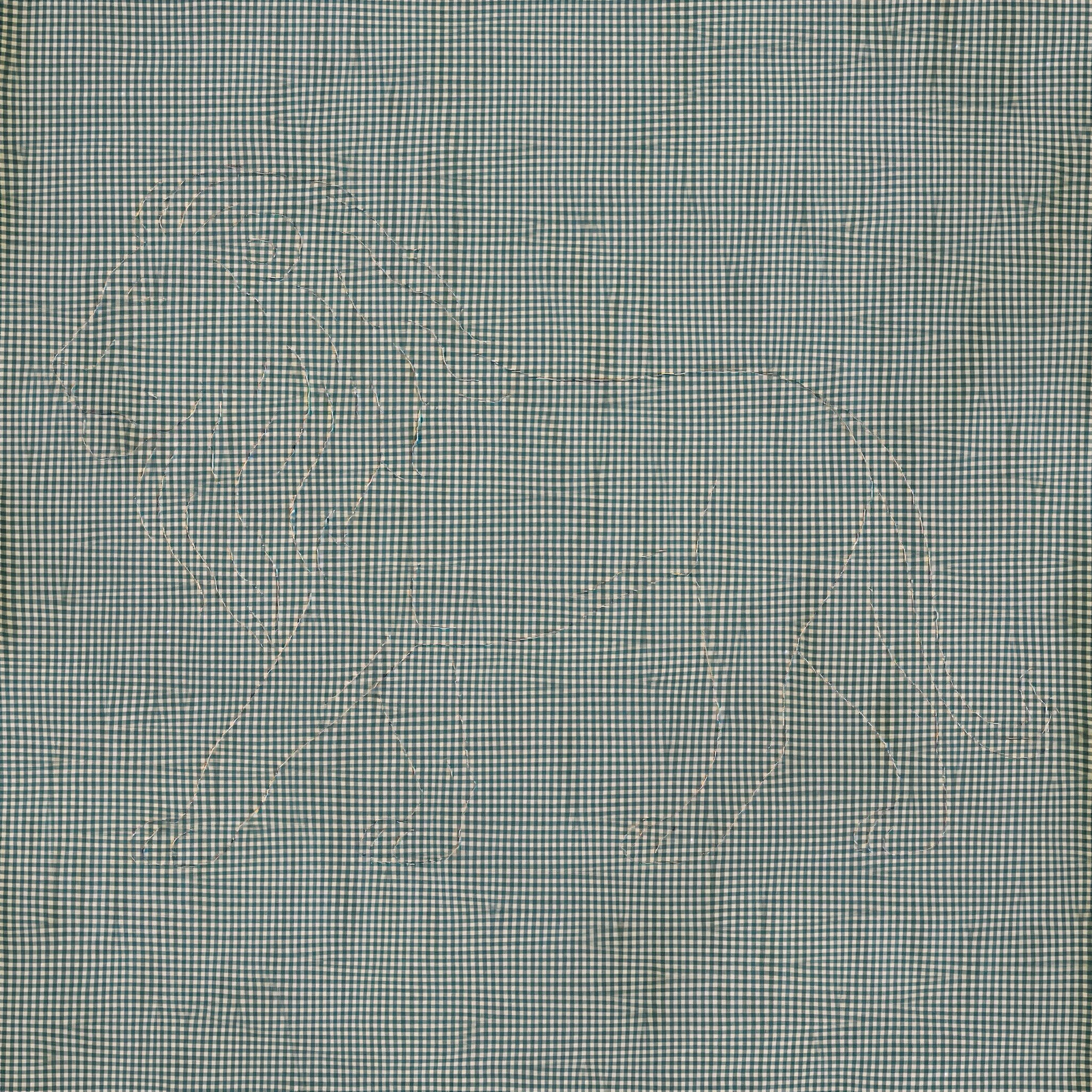}
    \caption{\label{fig:high_res_edit_1}%
      High-resolution (4096\texttimes{}4096) editing examples. \textbf{Please zoom in for full resolution.}}
    \Description{Large images showing editing results.}
\end{figure*}

Our method lends itself to high-resolution generation and editing. In Figures~\ref{fig:high_res_gen_1},~\ref{fig:high_res_gen_2}, we add results generated at high-resolution, presenting various painted features. 
Fig.~\ref{fig:high_res_edit_1} includes additional high-resolution editing results.

\section{Comparison Addition}
\label{asec:comparison}

\begin{figure}
  \centering
  \mbox{} \hfill
  \setlength{\cw}{0.165\linewidth}
  \setlength{\tabcolsep}{+0.001\linewidth}
  \begin{tabular}{cccccc}
        \includegraphics[width=\cw]{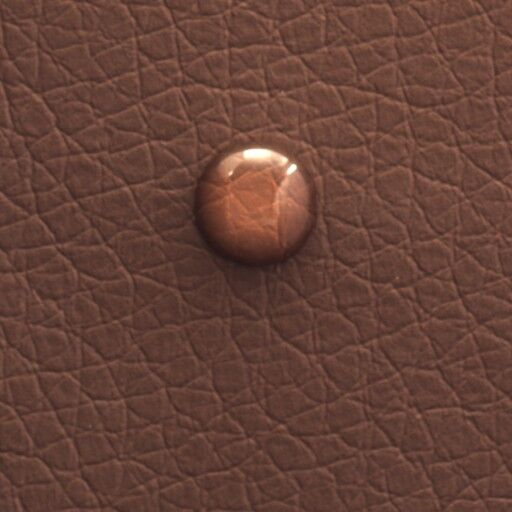} &
        \includegraphics[width=\cw]{images/qualitative_comparison/leather_mask_recol.jpeg} &
        \includegraphics[width=\cw]{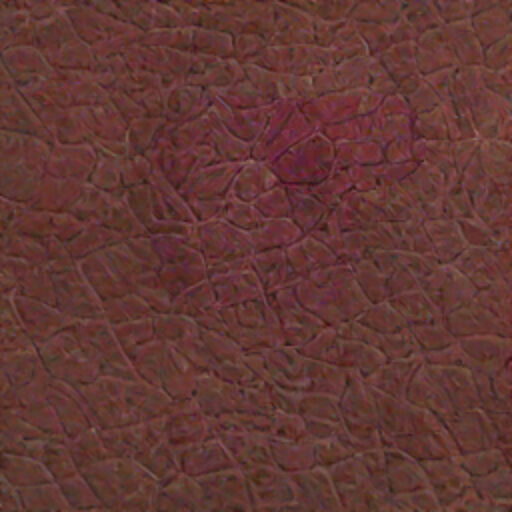} &
        \includegraphics[width=\cw]{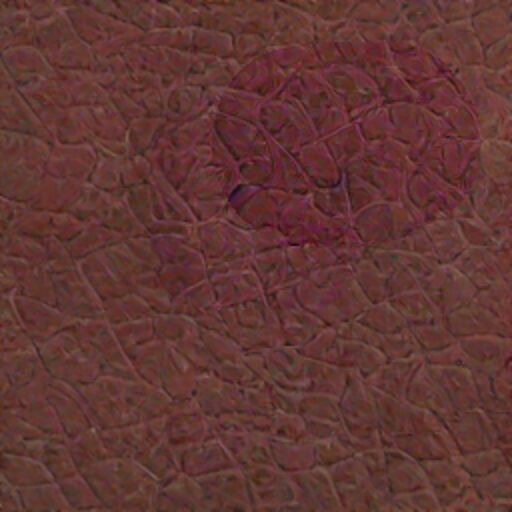} &
        \includegraphics[width=\cw]{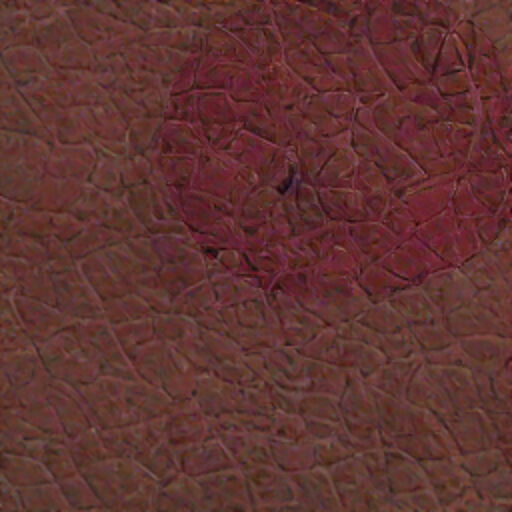} &
        \includegraphics[width=\cw]{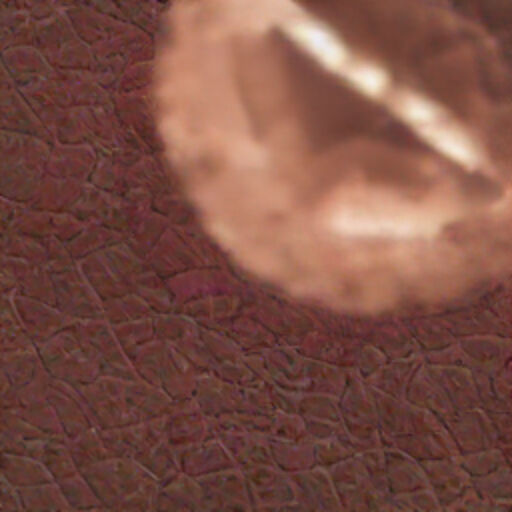} \\
        \fighlabel{Source} & \fighlabel{Mask} & \fighlabel{No dilation} & \fighlabel{$r_\text{dil}=27$} & \fighlabel{$r_\text{dil}=52$} & \fighlabel{$r_\text{dil}=77$}
            
    \end{tabular}
  \caption{\label{fig:painting_fail}%
           Failure case for Diffusion Texture Painting~\cite{hu2024diffusion}. The results are generated with a progressively dilated mask. Only at the largest size (dilated by 77 pixels), the feature is actually painted.}
    \Description{The Figure shows images generated by Diffusion Texture Painting for an increasing target mask.}
\end{figure}

We include in Fig.~\ref{fig:painting_fail} a representative failure case of Diffusion Texture Painting~\cite{hu2024diffusion}, as referenced in the main paper.

\section{Comparison of different thresholding functions}
\label{asec:compare_thr}
\begin{figure}
  \centering
  \setlength{\cw}{0.165\linewidth}
  \setlength{\tabcolsep}{+0.001\linewidth}
  \begin{tabular}{cccccc}
        \includegraphics[width=\cw]{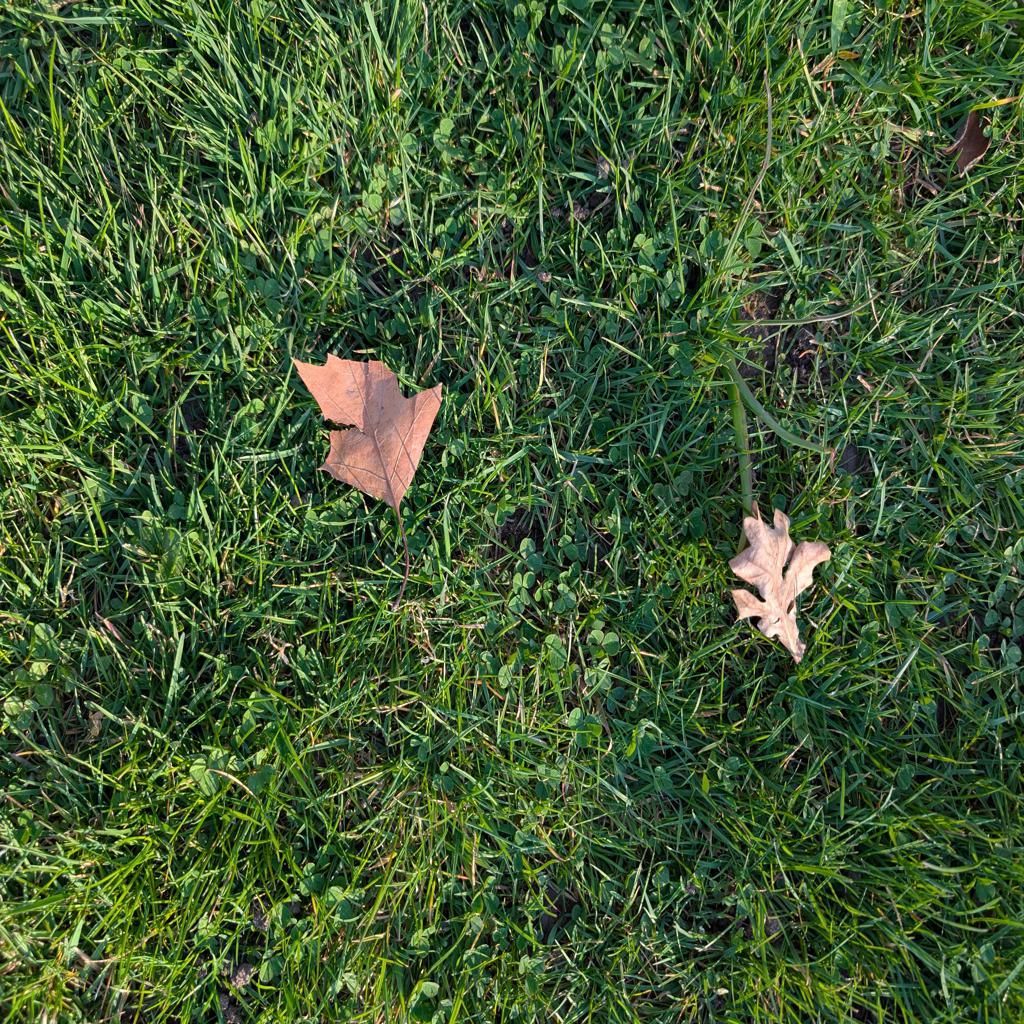} &
        \includegraphics[width=\cw]{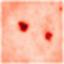} &
        \includegraphics[width=\cw]{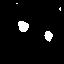} &
        \includegraphics[width=\cw]{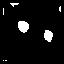} &
        \includegraphics[width=\cw]{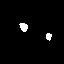} &
        \includegraphics[width=\cw]{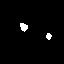} \\
        
        \includegraphics[width=\cw]{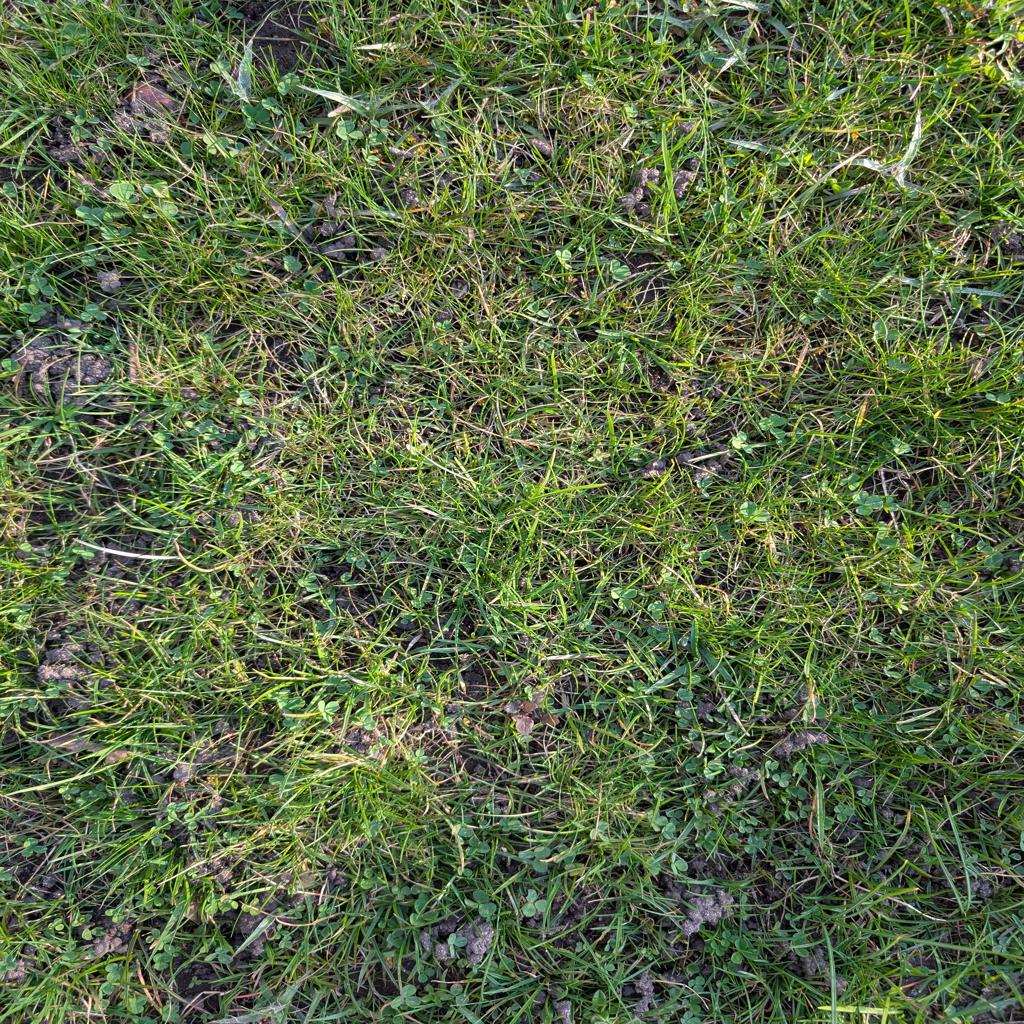} &
        \includegraphics[width=\cw]{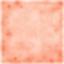} &
        \includegraphics[width=\cw]{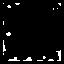} &
        \includegraphics[width=\cw]{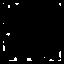} &
        \includegraphics[width=\cw]{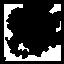} &
        \includegraphics[width=\cw]{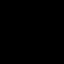} \\

        \includegraphics[width=\cw]{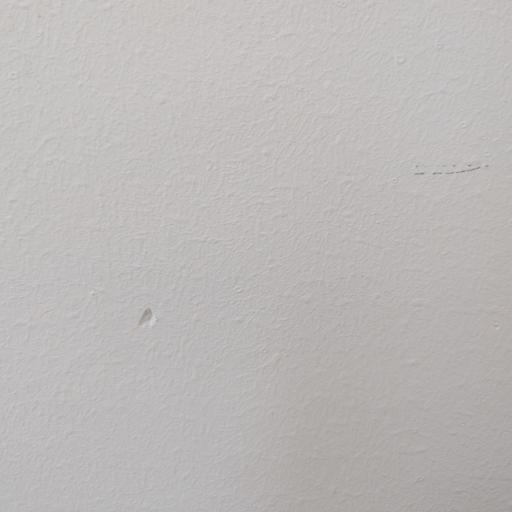} &
        \includegraphics[width=\cw]{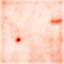} &
        \includegraphics[width=\cw]{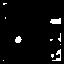} &
        \includegraphics[width=\cw]{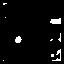} &
        \includegraphics[width=\cw]{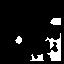} &
        \includegraphics[width=\cw]{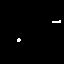} \\
        
        \includegraphics[width=\cw]{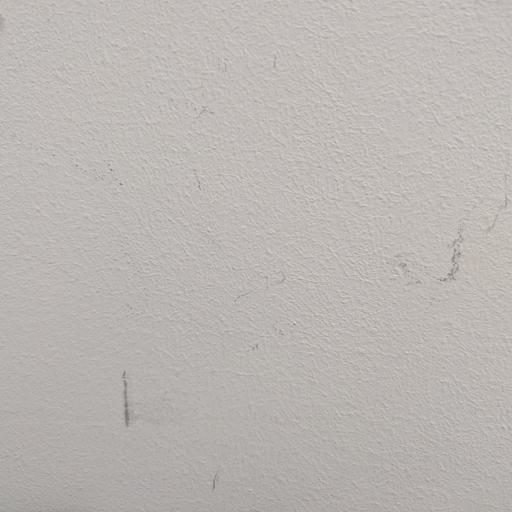} &
        \includegraphics[width=\cw]{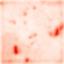} &
        \includegraphics[width=\cw]{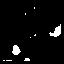} &
        \includegraphics[width=\cw]{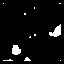} &
        \includegraphics[width=\cw]{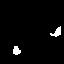} &
        \includegraphics[width=\cw]{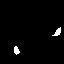} \\

        \includegraphics[width=\cw]{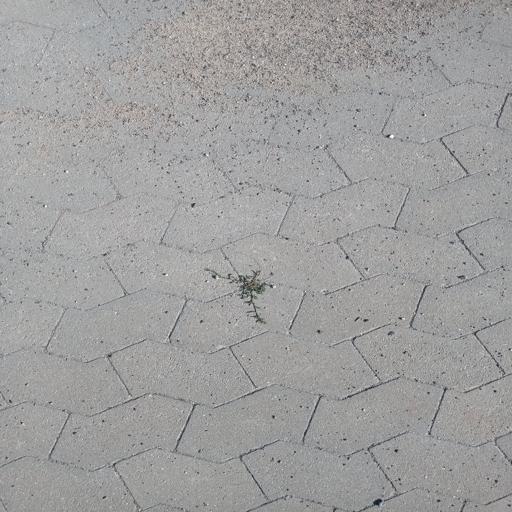} &
        \includegraphics[width=\cw]{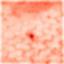} &
        \includegraphics[width=\cw]{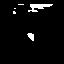} &
        \includegraphics[width=\cw]{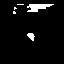} &
        \includegraphics[width=\cw]{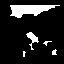} &
        \includegraphics[width=\cw]{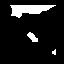} \\

        \includegraphics[width=\cw]{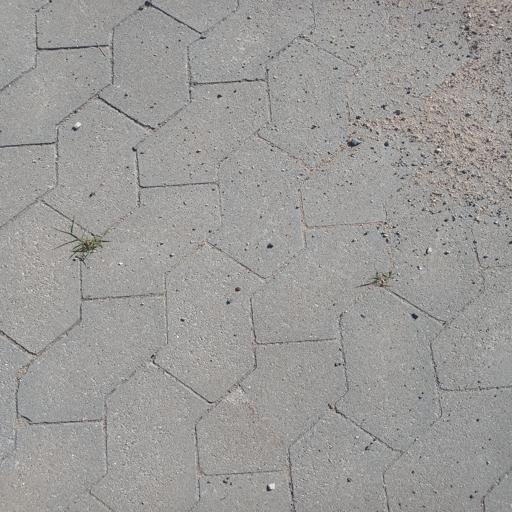} &
        \includegraphics[width=\cw]{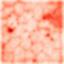} &
        \includegraphics[width=\cw]{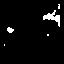} &
        \includegraphics[width=\cw]{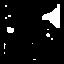} &
        \includegraphics[width=\cw]{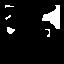} &
        \includegraphics[width=\cw]{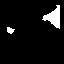} \\

        \fighlabel{Image} & \fighlabel{Ano. scores} & \fighlabel{Local 95\%} & \fighlabel{Dataset 95\%} & \fighlabel{Otsu} & \fighlabel{Our Eq.~\eqref{eq:threshold}} \\
            
    \end{tabular}
  \caption{\label{fig:compare_thr}%
    Comparing the binarization of anomaly scores using different thresholding functions.}
    \Description{The Figure shows binary images obtained with various methods.}
\end{figure}

In Section~\ref{sec:semantic}, we introduce our thresholding function, together with the theoretical justification for combining a local and global threshold.
In Fig.~\ref{fig:compare_thr} we present some representative results for alternative thresholding schemes that we considered. That is, we compare our approach to using local (per-image) quantiles, dataset-level quantiles, and Otsu's method.

\section{Timings}
\label{asec:time}

As presented in the main paper, our pipeline is trained in 3 stages: anomaly detection, feature clustering, and diffusion-based synthesis.
The first stage resembles the detection part of BlindLCA~\cite{ardelean2024blind}, which takes about 5 minutes.
The second stage performs the binarization, mines the positive and negative pairs of connected components, and performs the contrastive learning; the time is dominated by the last step, taking around 15 minutes.
The third stage is by far the most time-consuming: training the diffusion model to convergence takes between 6 and 12 hours on an A5000 NVIDIA GPU.
This time, however, could be greatly reduced by using a more fitting pre-training, with more textures compared to the DTD dataset used in our experiments.

\begin{table}%
    \caption{Comparison of inference time.}
    \label{tab:time_comparison}
    \begin{minipage}{\columnwidth}
    \begin{center}
    \begin{tabular}{lc}
      \toprule
      Method & Time (s) $\downarrow$ \\ 
      \midrule
      Image Analogies~\cite{hertzmann2001image} & 1200 \\
      Guided Correspondence~\cite{zhou2023neural} & \phantom{0}224 \\
      Neural Style Transfer~\cite{gatys2016image} & \phantom{00}75 \\
      Texture Reformer~\cite{wang2022texture} & 0.34 \\
      Ours & 0.98 \\
      \bottomrule
    \end{tabular}
    \end{center}
    \end{minipage}
\end{table}

\begin{figure}
    \centering
    \includegraphics[width=0.99\linewidth]{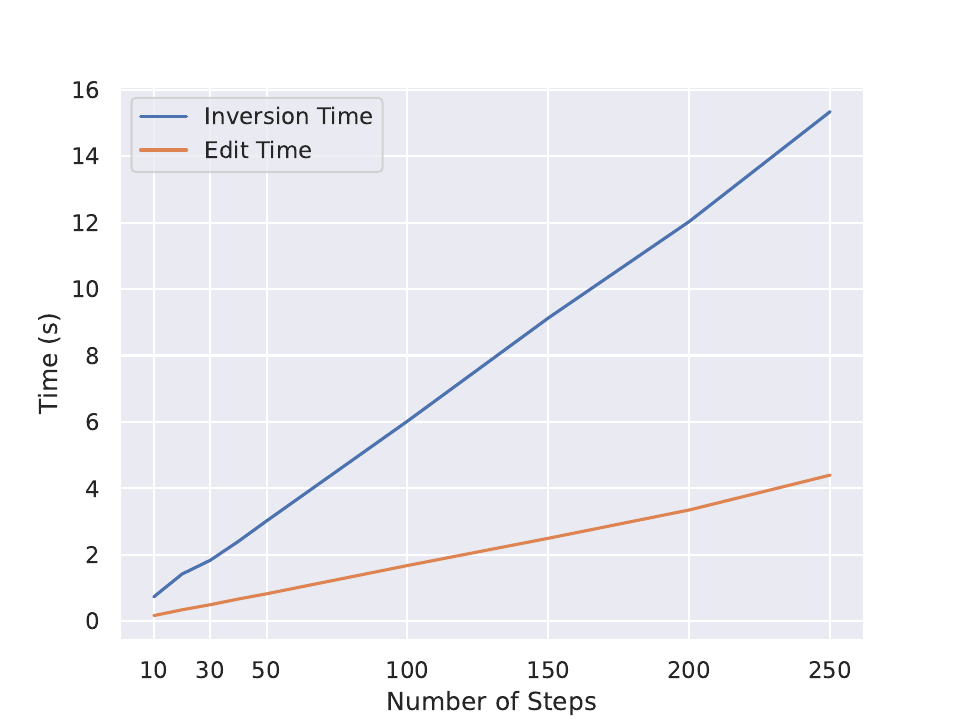} \\
    \includegraphics[width=0.99\linewidth]{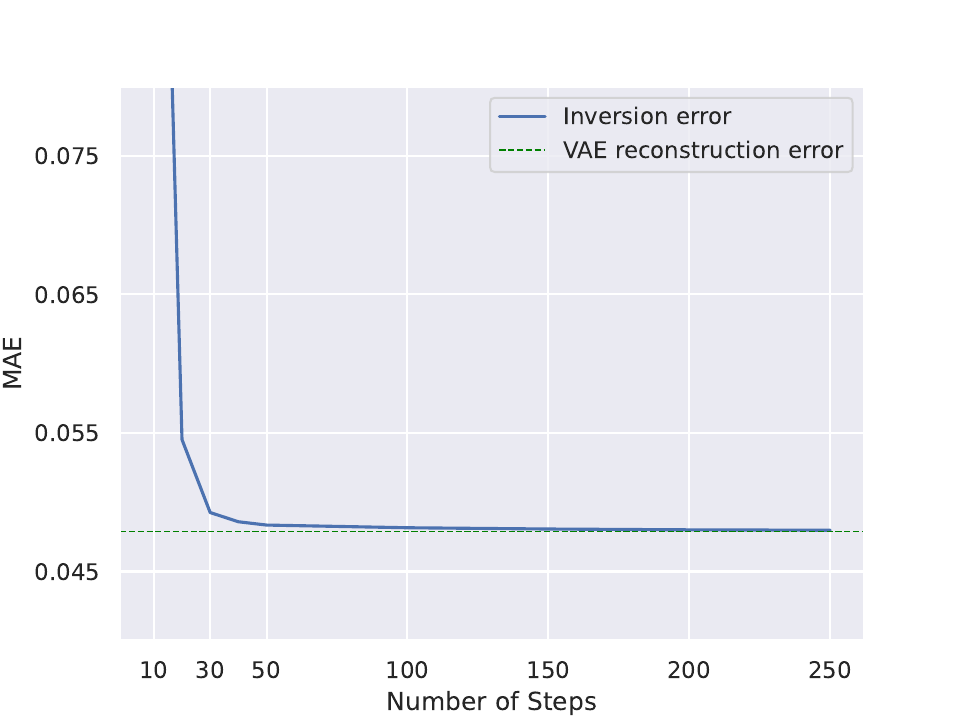}
    \caption{\label{fig:inversion_sup}Timing and errors for diffusion inversion.}
    \Description{Two plots showing the time and error development with the number of steps.}
\end{figure}

The training time is a drawback of our approach compared to single-exemplar texture synthesis methods (\eg, Image Analogies \cite{hertzmann2001image}, Guided Correspondence~\cite{zhou2023neural}); however, after training, our method can generate new images quickly, enabling interactive texture authoring.
During inference we perform 18 steps with the Heun solver, taking around 1 second for a 512\texttimes{}512 image, as shown in Table~\ref{tab:time_comparison}.
This Table, however, shows the timing for generating a new texture with a given mask. 
Editing an existing image requires an additional inversion step. 
Since we use an Euler solver with 4 fixed-point iterations, the number of diffusion model calls (NFE) is significantly larger. 
In the experiments performed in the main paper we use 250 steps (1000 NFE), taking about 15.5 seconds per image. After inversion, our noise-mixing editing does another 250 NFE, taking around 4.5 seconds.
The time spent on encoding and decoding using the Stable Diffusion VAE is negligible, which means the total time for editing is 20 seconds. 
That being said, our interactive editing framework benefits from the ability to dynamically choose the numbers of steps performed by the diffusion model. 
Namely, the user can use a small number of steps at the beginning of the editing sessions and only use the full number of steps to synthesize the final result.
Importantly, our noise-inversion on real images performs well even with significantly fewer steps, as seen in Fig.~\ref{fig:inversion_sup}. 
Note that after 40 steps the returns in quality are minimal and the inverted image converges to the VAE representation. The mean absolute error (MAE) is dominated by the loss of information during encoding-decoding.
For the interactive editing (as presented in the supplementary video) we only use 42 steps.

\section{Detailed numerical results}
\label{asec:numerical}

\begin{table}%
    \caption{Detailed quantitative results and comparison.}
    \label{tab:det_num}
    \begin{minipage}{\columnwidth}
    \begin{center}
    \begin{tabular}{@{}lccc@{\hspace{15pt}}ccc@{}}
      \toprule
        \textbf{Texture} & \multicolumn{3}{c}{\textbf{Ours}} & \multicolumn{3}{c}{\textbf{BlindLCA}} \\
        & Acc. $\uparrow$ & {IoU} $\uparrow$ & {F1} $\uparrow$ & {Acc.} $\uparrow$ & {IoU} $\uparrow$ & {F1} $\uparrow$ \\
        \midrule
        Tile    & \textbf{0.97}     & \textbf{0.70}     & \textbf{0.80}      & 0.96      & 0.54     & 0.67      \\
        Wood    & 0.96     & \textbf{0.48}     & \textbf{0.56}      & \textbf{0.97}      & 0.41     & 0.50      \\
        Leather & \textbf{0.99}     & \textbf{0.47}     & \textbf{0.57}      & 0.99      & 0.39     & 0.50      \\
        Carpet  & \textbf{0.99}     & \textbf{0.42}     & \textbf{0.53}      & 0.99      & 0.39     & 0.49      \\
        Grid    & \textbf{0.98}     & \textbf{0.30}     & \textbf{0.38}      & 0.53      & 0.17     & 0.25      \\
        \bottomrule
    \end{tabular}
    \end{center}
    \end{minipage}
\end{table}

We include in Table~\ref{tab:det_num} the detailed quantitative results of our method, and compare them with the metrics obtained using BlindLCA~\cite{ardelean2024blind}. We use macro averaging for the F1-score to emphasize the importance of detecting all feature types.
Note that this experiment uses a setting favorable to BlindLCA; that is, we excluded images that contain more than a single anomaly type per image.
Nonetheless, our approach consistently yields better metrics.

\section{Discussion on the choice of generative model}
\label{asec:diffusion}

\begin{table}%
\caption{Quantitative comparison of texture synthesis with different diffusion models. ``+ DTD'' denotes that the model was pretrained using the DTD~\cite{cimpoi14describing} textures dataset.}
\label{tab:diff_fid}
\begin{minipage}{\columnwidth}
\begin{center}
\begin{tabular}{lccccc}
  \toprule
  Model (FID $\downarrow$) & Tile & Carpet & Grid & Wood & Leather \\
  \midrule
  SD + ControlNet & 95.76 & 166.52 & 162.45 & 81.14 & 102.86 \\
  EDM2 & 41.60 & 132.12 & 125.70 & 38.67 & 126.05 \\
  EDM2 + DTD & 46.23 & \phantom{0}88.31 & \phantom{0}92.69 & 34.24 & 103.33 \\
  EDM + DTD & 50.04 & \phantom{0}57.14 & \phantom{0}81.82 & 68.29 & 105.88 \\
  \bottomrule
\end{tabular}
\end{center}
\end{minipage}
\end{table}

\begin{table}%
\caption{Quantitative comparison of texture feature painting using different diffusion models; EDM and EDM2 were pretrained on DTD~\cite{cimpoi14describing}. The models are evaluated by the accuracy of a image-to-class segmentation network trained with ground-truth labels.}
\label{tab:diff_seg}
\begin{minipage}{\columnwidth}
\begin{center}
\begin{tabular}{lccccc}
  \toprule
  Model (Acc $\uparrow$) & Tile & Carpet & Grid & Wood & Leather \\
  \midrule
  SD + ControlNet & 92.34 & 48.99 & 41.22 & 95.72 & \textbf{98.27} \\
  EDM2 & 94.40 & 93.86 & 90.58 & 95.49 & 86.82 \\
  EDM & \textbf{96.63} & \textbf{97.94} & \textbf{91.66} & \textbf{96.17} & 97.84 \\
  \bottomrule
\end{tabular}
\end{center}
\end{minipage}
\end{table}

\begin{table}%
\caption{Timing of different diffusion models}
\label{tab:diff_latency}
\begin{minipage}{\columnwidth}
\begin{center}
\begin{tabular}{lcc}
  \toprule
  Method & Throughput (img/s) $\uparrow$ & Latency (ms) $\downarrow$ \\
  \midrule
  SD + ControlNet & 0.8 & 2720 \\
  EDM2 & \textbf{1.4} & 2430 \\
  EDM & 1.3 & \phantom{0}\textbf{982} \\
  \bottomrule
\end{tabular}
\end{center}
\end{minipage}
\end{table}

Our feature painting framework is composed of several components that work together to enable the authoring and editing of textures with prominent features, which are learned from a small number of images.
The generative model is the component that links the anomaly segmentation to the various desired capabilities of the system (see points 2-4 in the introduction).
We choose to pose the generation as an image-to-image translation task (spatial labels to texture) using a diffusion model, and combine it with our noise-mixing and noise-uniformization to facilitate editing, feature transfer, and large texture generation. 
Alternatively, the generative task could be formulated as inpainting, to naturally support editing.
While arbitrary-size texture generation could in this case be formulated as out-painting, it is not clear how the other capabilities could be obtained. For example, it is not trivial to enable feature transfer, or to ensure that the painted feature is consistent with the initial texture (see Fig.~\ref{fig:ablation_nm} and Fig.~\ref{fig:edit_transfer}).

Another possible approach is to use an autoencoder or a VAE and encode the different types of features in latent space. A texture synthesis method that leverages this idea is TextureMixer~\cite{yu2019texture}. This approach could potentially be used within our framework by using the pixel-level anomaly segmentation masks to extract the rare features in a format compatible with TextureMixer's training process.
Nevertheless, it is unclear to what extent such method can adapt to thin structures (such as cracks) or how it can be extended to support feature transfer.

Other generative approaches, such as autoregressive models or flow-matching could be similarly considered. In general, however, we consider it out of the scope of this paper to incorporate all these methods in our framework to evaluate their performance on the various capabilities.
That being said, our choice of diffusion model is made without loss of generality within that category (spatially-conditioned diffusion models).
Our system uses EDM~\cite{karras2022elucidating} as the backbone for texture synthesis and editing; nonetheless, the proposed noise-mixing and noise-uniformization algorithms can be applied using virtually any diffusion model.
In our experiments, we use EDM because we find it to strike a good balance between image quality and speed.
In the following, we provide a quantitative comparison between EDM and two alternative diffusion models, namely Stable Diffusion (SD) \cite{rombach2022high} and EDM2~\cite{karras2024analyzing}.
In order to evaluate the image quality for a certain texture, we generate 100 images of the normal class and take 64 patches of size 128\texttimes{}128 from every image.
In the same manner, we also extract patches from the real images without anomalies, which were not seen during training.
Finally, the distribution of the generated and real patches are compared using the Fréchet Inception Distance (FID) \cite{heusel2017gans}.
We evaluate the three different models and in Table~\ref{tab:diff_fid}.
It can be seen that pretraining the models on the DTD~\cite{cimpoi14describing} dataset improves synthesis quality; however, rather surprisingly, Stable Diffusion performs worse despite its large scale pretraining.
Note that we have experimented with both full-model fine-tuning and ControlNet~\cite{zhang2023adding}, and we observed superior performance with the later.
We further evaluate the models ability to generate realistic prominent features on the textures in Table~\ref{tab:diff_seg}.
As the number of anomalous patches in MVTec is very small, and all anomalous images have been used for training the diffusion models, it is unfeasible to use FID in this case.
Therefore, we evaluate the models based on the ability of a segmentation network to correctly classify the generated features.
We train a CNN on the MVTec ground-truth labels and generate a set of semantic masks as test data. 
The diffusion models are conditioned on these masks to generate a set of 256 images, which are then segmented with the CNN.
Table~\ref{tab:diff_seg} reports the accuracy of these segmentations; a higher accuracy indicates that the features better reflect the training data (real MVTec anomalies).

The throughput and latency of the three methods are compared in Table~\ref{tab:diff_latency}. 
SD has the highest computation time of the three methods. 
While the throughput of EDM2 is similar to EDM, the latency is significantly higher, despite using the smallest (XS) variant of EDM2.
It can be reduced to 1166 milliseconds by using the same model for guidance, which allows computing the conditional and non-conditional noise directions in parallel.
Overall, these experiments suggest that EDM is a good choice for our use-case, considering the high-quality synthesis with a low latency.
Finally, we emphasize that even though we show that our choice of diffusion model is sound, and that good results can be obtained with a relatively small model with very little pretraining, we do not claim to have found the best possible model for this part of the pipeline, as this is not the scope of our work.

\section{Implementation Details}
\label{asec:implement}

Several implementation details have been omitted for brevity in the main text. We expand here the explanation of various steps and the hyperparameters used.

\subsection{Anomaly detection}
In the first stage of the pipeline, we apply FCA to the residuals of the VAE reconstruction. 
We use a similar preprocessing to \citet{ardelean2024blind}: resize the images to 512\texttimes{}512, use a Wide ResNet-50~\cite{BMVC2016_87}, and train the VAE for 10k iterations. 
After subtracting the original features from the reconstruction we use FCA with a patch size of 7\texttimes{}7, $\sigma_p=3$, and $\sigma_s=1$.

\subsection{Semantic feature segmentation}
\begin{figure}
  \centering
  \mbox{} \hfill

  \includegraphics[width=\linewidth]{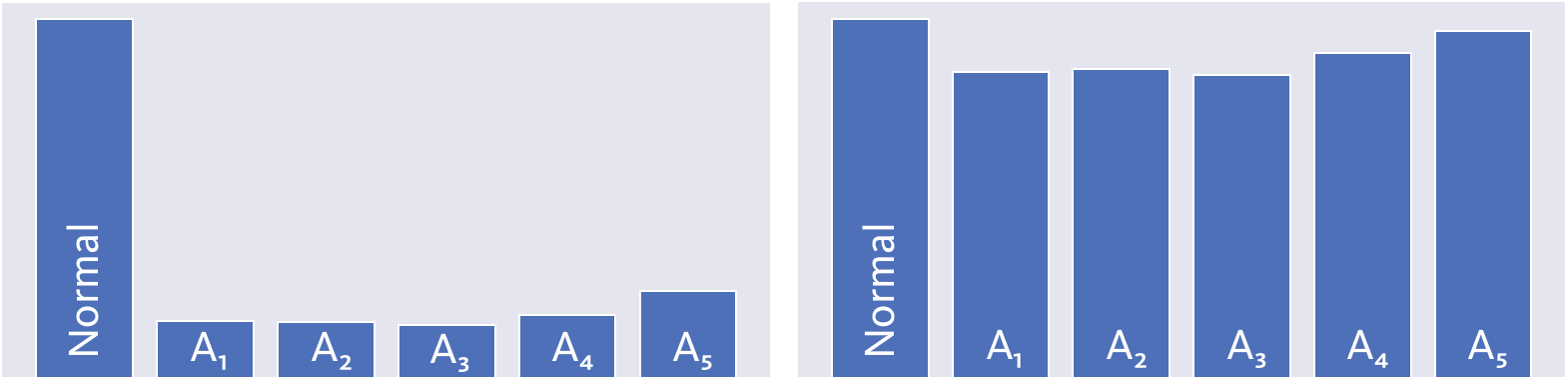}

  \caption{\label{fig:pair_sampling}%
    Histograms of sampled negatives from each anomaly class, used for contrastive learning: uniformly sampled (\emph{left}) and using our stratified sampling (\emph{right}).
    A disproportionate number of normal samples hinders the proper separation of the anomaly classes.%
  }
  \Description{Bar plot showing the relative number of samples from each class.}
\end{figure}

In order to enable an efficient training of the segmentation network through contrastive learning, we first build a database of positive and negative pairs.
As described in section~\ref{sec:semantic}, we first binarize the anomaly maps from the previous step using an adaptive threshold.
Afterward, we form groups of prominent features by finding the connected components.
To reduce some of the noise that arises from binarization, we perform a small erosion (2\texttimes{}2) and then eliminate objects smaller than 12 pixels.
For each region, we then compute a neural descriptor by averaging the ResNet features from the pixels inside the mask. We employ a weighted average using the softmax of the scores predicted by FCA, similarly to \citet{sohn2023anomaly}.
The descriptors are used to compute pair-wise distances between all feature groups.
To sample the positive pairs we simply take the closest $p=10$ feature groups in terms of distance.
Our stratified sampling of the negative pairs is more involved: the closest $50\%$ groups are first discarded as potential positives; then, the remaining descriptors are clustered using \kmeans to obtain coarse group categories.
We then select a number of $n$ negatives in a stratified manner from this pool, where $n$ is calculated as the expected number of groups. That is, the total number of groups minus the largest cluster, which contains normal features.
Since $n$ is generally smaller than $50\%$ of the original number of groups, the negatives are distributed more uniformly across the different types of prominent features.

Our segmentation network takes as input the ResNet features and computes task-aligned descriptors through contrastive learning. The network consists of only 3 convolutional layers with kernel sizes: 3\texttimes{}3, 3\texttimes{}3, 1\texttimes{}1,  LayerNorm normalization, and GeLU activations. During training, we use an additional 2-layer MLP head that is not used for clustering, as it is customary in self-supervised learning~\cite{grill2020bootstrap, chen2021empirical}.
We train the network for 10K iterations and then obtain per-pixel descriptors for all images.
Finally, these descriptors are clustered independently using \kmeans; the number of classes (prominent feature types) is assumed to be known by the user.
Generally, we believe the user would have a reasonable understanding of the features present in the dataset and choose the granularity of the clustering according to their use-case.
Alternatively, the user could simply use a clustering method that automatically detects the number of classes, such as DBSCAN.

\subsection{Synthesis}
To support conditional synthesis, we adapt the diffusion architecture used by EDM to incorporate spatial label maps.
Firstly, we lift the noise embedding to a H\texttimes{}W\texttimes{}C tensor instead of a C-dimensional vector.
Then, we compute label embeddings using two convolutional layers, and we add them to the noise embeddings.
Finally, we use another 1\texttimes{}1 convolution before propagating this spatial embedding to all the U-Net blocks.
This minimal modification of the architecture effectively enables the control of the model through the label mask.

We pretrain the model on the DTD dataset for 750K iterations, taking around 12 hours.
Then, we separately fine-tune the model for each texture for another 750K iterations.
The diffusion is performed in the latent space of SD, making it more efficient to generate high-resolution images.
The only exception is the SVBRDF synthesis experiment, which performs the diffusion directly in the space of material maps.
All models are trained at a resolution of 64\texttimes{}64.
We use Pytorch's RandomResizedCrop augmentation, along with random horizontal and vertical flips, and a slight color jitter.

For interactive editing, we developed a Blender script that leverages the native tool for painting masks on textures. 
We process the masked image to extract the desired edit and take a bounding patch of at least 442\texttimes{}442 around the masked region. 
Our noise-mixing algorithm is then applied with the Euler solver for 42 steps. The average edit latency is 1.5 seconds, which enables interactive asset modifications (as seen in the video attached to this supplementary material).

\subsubsection{Noise uniformization}

Our noise uniformization algorithm specifies a way to make a noise map $\vect{w}$ more consistent with a different noise tensor \vect{z}, which dictates the \emph{style} of the instance (overall color, contrast, pattern density, etc).
We achieve this by making the noise map follow the same low frequency distribution, as described in the main paper: 

\noindent $\vect{w}' := \vect{w} - \texttt{blur}(\vect{w}) + \texttt{upscale}(\texttt{shuffle}(\texttt{downscale}(\texttt{blur}(\vect{z})))$.

\noindent For blurring, we use a Lanczos filter with a cutoff frequency $f_c=0.1$, and we downsample the filtered noise to a resolution of 32 \texttimes{} 32. 
For the supplementary experiments, based on StableDiffusion (Fig.~\ref{fig:sd_uniform}), we use the same parameters, except that we only perform the downsampling and shuffling along the columns. This is because the generated images resemble panoramas, which only have a stationary nature along the width of the image.
To create a large uniform noise tensor, we first generate white noise and then divide the map into equally-sized non-overlapping patches. The first patch conveys the style ($\vect{z}$). All other patches ($\vect{w}$) are modified using the algorithm described above to follow this prototype; finally, the patches are rearranged to form the large noise map, which is the input for the diffusion model.

\subsubsection{SVBRDF}

The main difference between the synthesis of material maps compared to RGB images is that we do not use a latent model for SVBRDFs.
We train the model with 64\texttimes{}64 patches from the input material and do not use color jitter for this experiment.
Since there is only one SVBRDF as input, we did not use contrastive learning for the semantic segmentation.
Instead, as the irregularities are easily noticeable in either the albedo or the roughness maps, we computed embeddings using a random of set of 5\texttimes{}5 filters applied on the pixels' features.
The resulting descriptors were directly clustered using \kmeans.

\subsection{Video}

The video attached to this supplementary material contains an interactive editing session in Blender.
\TAedit{Here, we make use of the brush tool from Blender, which allows the user to paint on a 3D mesh, while enabling programmatic access to the updated underlying UV-mapped 2D texture.
We use the mask-painted texture to extract the semantic conditioning and then apply our trained method on the texture patch which is being edited.
To improve the latency, we preload into GPU memory the weights of the model and the diffusion trajectory needed for our noise-mixing.
This is done in a background thread when a certain object is selected for editing, so that a single set of weights must be stored in memory at a specific time.}
All textures in the scene have a resolution of 2048\texttimes{}2048, and they have been generated using our method except for the teapot, for which we use an arbitrary image to showcase our feature transfer capabilities. The blemishes are transferred from the MVTec tile texture.
Note that we only apply our method to the base color of the materials, leaving the other material maps unchanged.

\subsection{Code Release}

\TAedit{The code is available at: \href{https://github.com/TArdelean/FeaturePainting}{github.com/TArdelean/FeaturePainting}.}

\end{document}